\documentclass{article}

\usepackage{microtype}
\usepackage{graphicx}
\usepackage{subcaption}
\usepackage{booktabs} 

\usepackage{hyperref}

\usepackage[preprint]{icml2026}

\usepackage{amsmath}
\usepackage{amssymb}
\usepackage{mathtools}
\usepackage{amsthm}
\usepackage{braket} 

\usepackage[capitalize,noabbrev]{cleveref}

\theoremstyle{plain}

\theoremstyle{definition}

\theoremstyle{remark}

\usepackage[textsize=tiny]{todonotes}

\icmltitlerunning{\textsc{MolPaQ}: Modular Quantum–Classical Patch Learning for Interpretable Molecular Generation}

\begin{document}

\twocolumn[
  \icmltitle{\textsc{MolPaQ}: Modular Quantum–Classical \\ Patch Learning for Interpretable Molecular Generation}



  \icmlsetsymbol{equal}{*}

  \begin{icmlauthorlist}
    \icmlauthor{Syed Rameez Naqvi}{yyy}
    \icmlauthor{Lu Peng}{yyy}
  \end{icmlauthorlist}

  \icmlaffiliation{yyy}{Department of Computer Science, Tulane University, New Orleans, LA, USA}

  \icmlcorrespondingauthor{Syed Rameez Naqvi}{snaqvi@tulane.edu}
  \icmlcorrespondingauthor{Lu Peng}{lpeng3@tulane.edu}

  \icmlkeywords{molecular generation, quantum machine learning, hybrid generative models, graph synthesis}

  \vskip 0.3in
]



\printAffiliationsAndNotice{}  

\begin{abstract}
  Molecular generative models must jointly ensure validity, diversity, and property control, yet existing approaches typically trade off among these objectives. We present \textsc{MolPaQ}, a modular quantum–classical generator that assembles molecules from quantum-generated latent patches. A $\beta$-VAE pretrained on QM9 learns a chemically aligned latent manifold; a reduced conditioner maps molecular descriptors into this space; and a parameter-efficient quantum patch generator produces entangled node embeddings that a valence-aware aggregator reconstructs into valid molecular graphs. Adversarial fine-tuning with a latent critic and chemistry-shaped reward yields 100\% RDKit validity, 99.75\% novelty, and 0.905 diversity. Beyond aggregate metrics, the pretrained quantum generator, steered by the conditioner, improves mean QED by $\approx$2.3\% and increases aromatic motif incidence by $\approx$10–12\% relative to a parameter-matched classical generator, highlighting its role as a compact topology-shaping operator.
\end{abstract}

\section{Introduction}
\label{sect:intro}

Molecular generative models play a central role in modern drug discovery by enabling systematic exploration of chemical space to identify valid, novel, and synthesizable compounds~\cite{chen2023deep}. Existing approaches broadly fall into three categories: (1) sequence-based models that generate SMILES strings using recurrent or transformer architectures~\cite{andronov2025accelerating}, (2) graph-based methods built on variational autoencoders (VAEs), generative adversarial networks (GANs), or diffusion models~\cite{zhang2024quantitative}, and (3) 3D coordinate prediction frameworks that directly model atomic geometries~\cite{moon20233d}. While large-scale diffusion and flow-based models achieve strong likelihood-based performance, they typically rely on monolithic architectures and extensive training budgets, and are optimized for global distributional fidelity rather than modular control or interpretability.

Fragment- and motif-based generators~\cite{jin2020hierarchical,podda2020deep,luong2023fragment} improve plausibility by assembling reusable substructures, but often trade off flexibility and fine-grained property control~\cite{gao2022enhancing}. GAN-based models continue to face challenges such as mode collapse and unstable validity--uniqueness trade-offs~\cite{goni2025hingerlc}, while property conditioning is frequently heuristic and weakly integrated. These limitations motivate architectures that explicitly separate conditioning, generation, and assembly, enabling targeted control and systematic analysis under constrained generative capacity.

Quantum machine learning (QML) offers a compact and expressive representation for capturing correlated molecular structure via entangled latent variables~\cite{Falco2024quantum,torabian2025molecular,naguleswaran2024quantum}. However, most existing quantum generative models concentrate quantum computation within a single monolithic decoder, limiting modularity, interpretability, and controllability~\cite{cerezo2022challenges}. Recent work has begun to explore hybrid architectures in which quantum and classical components interact through well-defined interfaces, demonstrating improved control and bias mitigation~\cite{zou2025conquermodulararchitecturescontrol,tian}. These developments suggest that modular hybrid pipelines, rather than end-to-end quantum decoders, offer a promising path forward under realistic near-term constraints.

We introduce \textsc{MolPaQ}, a modular quantum–classical framework that generates molecules by assembling quantum‑generated latent patches. A parameterized quantum circuit produces entangled patch embeddings that capture local chemical structure, while classical modules handle descriptor conditioning and chemically constrained assembly. In contrast to prior quantum VAEs or GANs that confine quantum computation to a monolithic decoder, \textsc{MolPaQ} inserts quantum patch synthesis into a classical conditioning–aggregation pipeline, enabling hierarchical control and clean attribution of quantum contributions.

\textsc{MolPaQ} rests on three components. (1) A small conditioner maps descriptors (QED, logP, SA) into a property‑aligned subspace of a pretrained $\beta$-VAE, providing direct descriptor‑to‑latent steering. (2) A parameter‑efficient quantum circuit transforms each conditioned latent vector into an entangled patch embedding. (3) A constraint‑aware aggregator assembles patches into full molecules using distance‑based connectivity, degree caps, six‑ring detection, and aromatic upgrades, without relying on fixed fragment vocabularies. Decoupling conditioning, quantum patch synthesis, and assembly yields a controllable, interpretable generator under explicit capacity constraints. Our main contributions are as follows:

\textit{Modular quantum patch synthesis:} A hybrid generative framework that integrates quantum circuits as patch generators within a descriptor-conditioned molecular pipeline, isolating the contribution of quantum latent synthesis.

\textit{Constraint-aware molecular assembly:} A connectivity-first, valence-aware aggregation procedure that enforces chemical realism through explicit structural constraints.

\textit{Hybrid architecture and systematic analysis:} An end-to-end quantum--classical pipeline and quantitative studies of scalability, interpretability, and the structural effects induced by quantum patch generation relative to classical ablations.



\section{Related Work and Research Gaps}
\label{Sect:background}

Prior work in molecular generation spans SMILES-based models~\cite{andronov2025accelerating}, graph-based VAEs~\cite{gomez2018automatic}, GANs~\cite{decao2018molgan, liu2023molfiltergan, shi2020graphaf, zhang2024quantitative}, motif-driven assembly~\cite{jin2020hierarchical, podda2020deep, luong2023fragment}, and emerging quantum generative models~\cite{li2021quantum, moussa2023application, wu2024qvae, torabian2025molecular}. While these approaches have advanced validity and diversity, many treat generation as a monolithic process, lack explicit symbolic constraint integration, and offer limited architectural support for property-guided synthesis~\cite{wu2024qvae}. In particular, existing quantum models typically embed logic only at the decoder stage, restricting modularity and interpretability~\cite{avramouli2022quantum, smith2025bridging}.

Recent diffusion- and flow-based models~\cite{lee2023exploring} achieve strong likelihood-based performance at scale, but operate under substantially different assumptions, including extensive training budgets and end-to-end monolithic architectures. These methods optimize global distributional fidelity rather than modular compositional control, and are therefore complementary to, rather than directly comparable with, constrained hybrid pipelines designed for interpretability and structured synthesis.


\textsc{MolPaQ} is designed to address these gaps by adopting a modular quantum--classical architecture that supports patch-level generation, constraint-aware assembly, and property-guided control under explicit capacity constraints. A detailed comparison with existing classical and quantum approaches is provided in Supplementary Sect.~\ref{Sect:suppl_background}.

\section{Proposed Framework}
\label{sect:proposed}


\textsc{MolPaQ} draws inspiration from hierarchical graph generation~\cite{jin2020hierarchical,bian2024hierarchical} and constraint-aware generative loops~\cite{gomez2018automatic}, while incorporating quantum circuits as patch-level latent generators rather than monolithic decoders~\cite{li2021quantum}. 

\begin{figure*}
    \centering
    \includegraphics[width=0.8\linewidth]{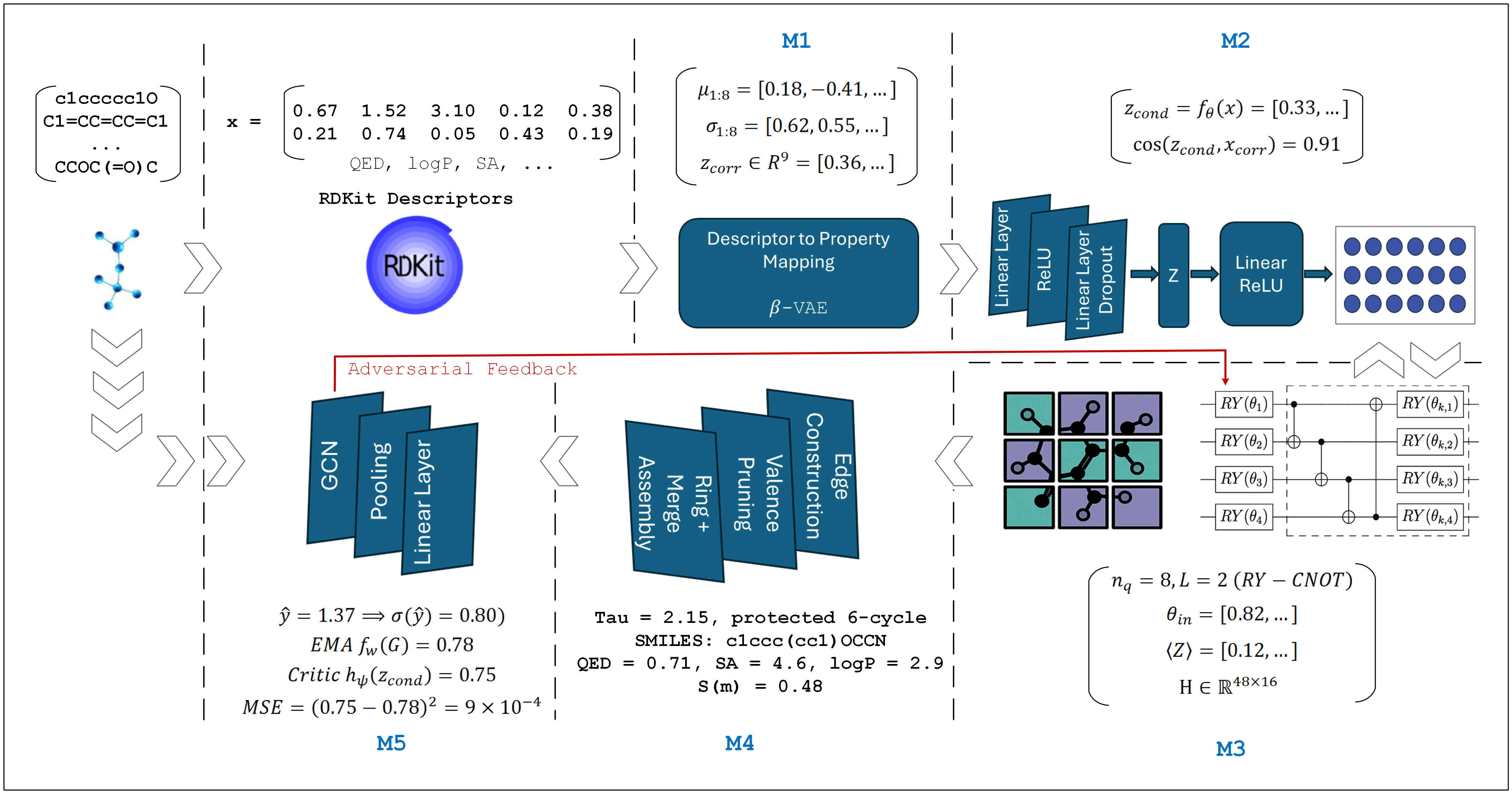}
    \caption{Overview of the \textsc{MolPaQ} Pipeline}
    \label{fig:blockdiagram}
\end{figure*}

\subsection{\textsc{MolPaQ} Pipeline}

Figure~\ref{fig:blockdiagram} summarizes the end-to-end flow from SMILES to molecules via five modules (M1–M5). The numeric examples correspond to real values from our experiments, and are provided for convenient dataflow visualization.

    



    \textit{\textbf{Latent Pretraining and Conditioning (M1--M2):}}
    To enable property-guided control, we first construct a property-aligned latent space using a $\beta$-VAE trained on QM9 (M1), where a GIN encoder maps molecular graphs to continuous latent representations. After training, latent dimensions are ranked by correlation with key molecular properties, and the top-$k$ axes define a reduced target subspace for conditioning.

    A lightweight 3-layer MLP (512-256, ReLU, dropout 0.1) maps descriptor vectors into the top-$k$ property-aligned latent axes of the pretrained $\beta$-VAE (M2), supplying a low‑dimensional conditioning signal for the quantum patch generator. These modules provide control but no generative capacity. Full objectives and architectural details for M1-M2 appear in Suppl. Sect.~\ref{sect:suppl_M1}-\ref{sect:suppl_M2}.

    \textit{\textbf{Patch Generator (M3):}} \textsc{MolPaQ} constructs each molecule from a single quantum-generated \emph{latent patch}, which encodes a coherent local motif. This patch is decoded into node-level embeddings and transformed into a chemically valid molecular graph by the aggregator (M4). While the generator operates at patch granularity, the aggregator enriches each patch with valence-aware connectivity, ring closure, and aromatic upgrades $-$ enabling full-molecule synthesis from a single expressive unit.

    The generator begins its operation upon receiving a conditioned latent vector $\mathbf{z}_{\text{cond}} \in \mathbb{R}^{d_z}$ from M2. This vector is first linearly projected to match the number of qubits, $n_q$, defining a set of input angles:
    \begin{equation}
        \boldsymbol{\theta}_{\text{in}} = W_{\text{in}} \mathbf{z}_{\text{cond}} + \mathbf{b}_{\text{in}}, 
        \qquad \boldsymbol{\theta}_{\text{in}} \in \mathbb{R}^{n_q}
        \label{eqn:qgen_input}
    \end{equation}
    These angles serve as rotation parameters for the quantum circuit. Each input element controls an $R_Y$ gate, producing a superposition that embeds the latent code into the amplitude space of a quantum register:
    \begin{equation}
        \ket{\psi_0} = \bigotimes_{i=1}^{n_q} R_Y(\theta_{\text{in},i}) \ket{0}^{\otimes n_q}
        \label{eqn:qgen_state}
    \end{equation}

    To enable non-local correlations between latent dimensions, we apply $L$ layers of trainable entangling operations. Each layer $\ell$ consists of single-qubit rotations parameterized by angles $\boldsymbol{\alpha}_{\ell,i}$ followed by a ring of CNOT gates:
    \begin{equation}
\begin{split}
    U_{\ell} &=
    \left(\prod_{i=1}^{n_q} 
        R_Z(\alpha_{\ell,i}^z)
        R_Y(\alpha_{\ell,i}^y)
        R_X(\alpha_{\ell,i}^x)
    \right) \\
    &\quad\times
    \left(\prod_{i=1}^{n_q}
        \text{CNOT}(i, (i{+}1) \bmod n_q)
    \right)
\end{split}
\label{eqn:qgen_layer}
\end{equation}

\begin{figure}
        \centering
        \includegraphics[width=\linewidth]{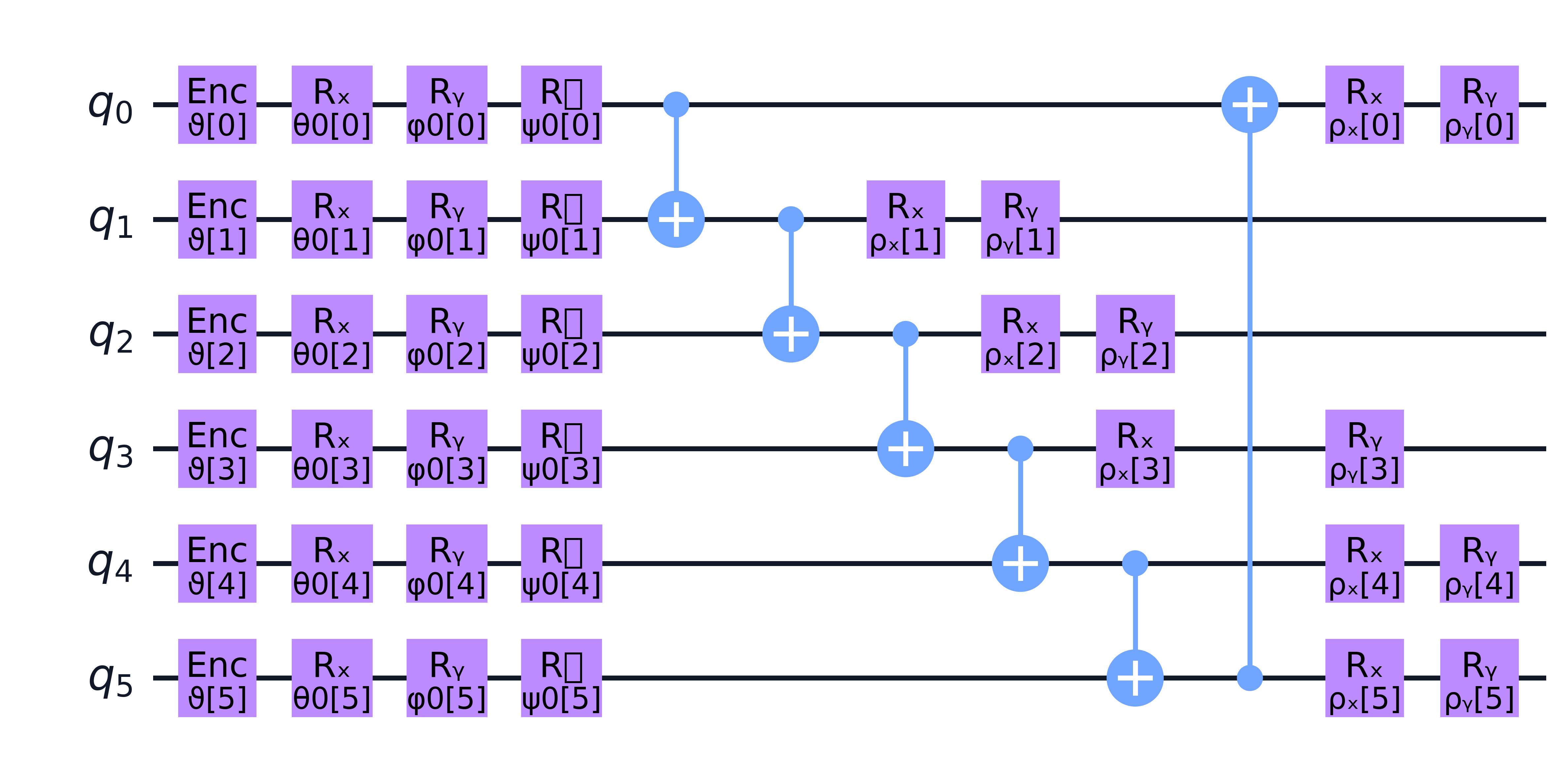}
        \caption{Parameterized RY–CNOT circuit used in M3. Latent angles control initial rotations, followed by two
        Strongly Entangling Layers (SEL) with ring-connectivity CNOTs. 
        The measured expectations $\langle Z_i \rangle$ form quantum-generated node embeddings.}
        \label{fig:qgen-circuit}
    \end{figure}

    Stacking such layers, as inspired by the Strongly Entangling Layers template, yields an expressive ansatz that can capture high-order dependencies with shallow circuit depth $-$ an essential feature for near-term quantum devices. After the final layer, each qubit is measured in the Pauli-$Z$ basis to extract its expectation value:
    \begin{equation}
        g_i = \langle \psi_L | Z_i | \psi_L \rangle, \quad i = 1,\dots,n_q
        \label{eqn:qgen_expval}
    \end{equation}
    producing a real-valued vector $\mathbf{g} \in [-1, 1]^{n_q}$. 
    These values are then normalized and post-processed by a lightweight classical head,
    \begin{equation}
        \mathbf{h} = \sigma\!\left(W_{\text{post}}\mathbf{g} + \mathbf{b}_{\text{post}}\right)
        \label{eqn:qgen_post}
    \end{equation}
    and reshaped into a tensor $\mathbf{H} \in \mathbb{R}^{N_{\text{nodes}} \times F_{\text{node}}}$, 
    where each row encodes the feature vector of a node within the generated patch.  
    In practice, $N_{\text{nodes}}{=}48$ and $F_{\text{node}}{=}16$ work well, offering a balance between local detail and computational efficiency. Figure~\ref{fig:qgen-circuit} visualizes the parameterized RY–CNOT circuit used in M3. Each input latent component $z_{\text{cond},i}$ is linearly mapped to a rotation angle $\theta_{in,i}$ controlling an
    RY gate on qubit~$i$. Two strongly entangling layers (SEL) follow, each comprising single-qubit rotations $\{R_X, R_Y, R_Z\}$ and a ring of CNOT gates. Expectation values $\langle Z_i \rangle$ are measured to form the node-feature vector $g \in [-1,1]^{n_q}$, subsequently post-processed by a classical head to yield the patch tensor $H \in \mathbb{R}^{N_{\text{nodes}}\times F_{\text{node}}}$. 

    The resulting patch embeddings can be viewed as quantum-generated node embeddings $-$ continuous, geometry-aware encodings that later form the building blocks of complete molecules.  
    Because every latent dimension is physically encoded as a rotation angle or an entangling parameter, the circuit inherently enforces smooth, differentiable transitions between structural motifs.  
    This property allows the model to interpolate between substructures in a chemically meaningful way, producing smoother latent-to-structure mappings than purely classical decoders.

    \textit{\textbf{Patch Aggregator (M4):}}
The aggregator is
\emph{connectivity-first and valence-aware} with explicit protection of aromatic cores. 
Given a quantum patch $\mathbf{H}\!\in\!\mathbb{R}^{N\times F}$ with nonzero rows as node embeddings $\{\mathbf{h}_u\}$, we propose undirected edges by proximity
\begin{equation}
E_0=\bigl\{(u,v):\ \|\mathbf{h}_u-\mathbf{h}_v\|_2\le \tau \bigr\}
\label{eq:agg_thresh_main}
\end{equation}
and construct $G_0=(V,E_0)$.
We then \emph{shape} $G_0$ into a chemically plausible scaffold before any RDKit call:

\smallskip
\noindent\emph{(1) Six-ring discovery and protection.} 
We detect 6-cycles using an auxiliary $k$NN graph built from $\{\mathbf{h}_u\}$; edges belonging to a discovered 6-cycle are added (within a slack on $\tau$), tagged as \textit{protected}, and preserved through later pruning. This biases assembly toward stable aromatic backbones without hard-coding ring vocabularies.

\noindent\emph{(2) Valence-aware degree pruning.}
Type-specific degree caps enforce headroom for $\mathrm{sp^2}$ chemistry (e.g., C:3 during build, O:2, N:3, F:1). 
We greedily remove surplus \emph{non-protected} edges until caps are met,  and then, if necessary, touch protected edges.

\noindent\emph{(3) Distance-ranked double-bond upgrades.}
From the sanitized single-bond scaffold we rank edges by latent-space distance and upgrade a small quota to \textsc{double}, with a mild preference for ring bonds; if sanitization fails, the edit is rolled back. 
This yields conjugation without pre-imposing alternating patterns.

\noindent\emph{(4) Two-pass aromatization without forced kekulization.}
We perform a light aromatization pass (C/N-only 6-cycles, degree $\le\!3$), fragment to the largest component, and run a second pass. 
Final sanitize omits forced kekulization to preserve aromatic flags (then re-perceives aromaticity).

\noindent\emph{(5) Dual decode \& selection.}
As in the baseline, we produce two candidates: 
(i) a graph-route from $G_0$ using atom/bond templates; 
(ii) a coordinate-route from pairwise distances. 
Both are sanitized and scored by

\begin{equation}
        \begin{split}
            S(m) &= \lambda_{\text{QED}}\!\cdot\!\text{QED}(m)
            - \lambda_{\text{SA}}\!\cdot\!\text{SA}(m) \\
            &\quad - \lambda_{\log P}\!\cdot\!\max\!\big(0,\log P(m)-\gamma\big)
        \end{split}
        \label{eq:agg_score}
    \end{equation}
    
\noindent and we return $\arg\max_m S(m)$. 
A small min-size fallback (atoms/bonds) ensures robustness. 

In Sect.~\ref{sect:results}, we report near-perfect RDKit validity, which we attribute to the three guardrails we have introduced in the proposed aggregator. As already mentioned, (i) connectivity-first assembly with type-aware degree caps, (ii) reversible double-bond upgrades with sanitize-rollback, and (iii) a dual-decode fallback for rare edge cases, prevent impossible valences and rescue borderline assemblies before RDKit sanitization. We therefore expect very low sanitize-failure rates independent of the quantum vs. classical choice of M3. See Alg.~\ref{alg:m4} for aggregator operation.

    \begin{algorithm}
        \small
        \caption{Connectivity-first, valence-aware aggregation}
        \label{alg:m4}
        \textbf{Input:}~Patch embeddings $\mathbf{H}$; threshold $\tau$; caps $\mathcal{C}$ \\
        \textbf{Output:}~Molecule $m$ \\
        $E_0 \!\leftarrow\! \{(u,v):\|\mathbf{h}_u-\mathbf{h}_v\|_2\le\tau\}$; $G_0\!\leftarrow\!(V,E_0)$ \\
        \textbf{Ring protection:} detect 6-cycles in $k$NN graph; add edges (within slack) and tag \textsc{protected} \\
        \textbf{Degree pruning:} greedily remove non-protected edges until type caps $\mathcal{C}$ satisfied \\
        \textbf{Double upgrades:} upgrade a small quota of shortest edges; rollback on sanitize failure \\
        \textbf{Aromatize:} two passes around fragmenting; final sanitize w/o forced kekulization \\
        $m_1\!\leftarrow$ graph-route decode($G_0$);\quad $m_2\!\leftarrow$ coord-route decode($\mathbf{H}$) \\
        \textbf{return} $\arg\max_{m\in\{m_1,m_2\}} S(m)$ per Eq.~\ref{eq:agg_score}
    \end{algorithm}

\textit{\textbf{Graph Discriminator (M5):}} We propose an edge-aware GINE discriminator with node features
(atomic number, total valence, degree, aromatic flag, formal charge, total H) and edge features (bond order, aromatic, conjugated, in-ring). 
Two GINEConv layers (trainable $\epsilon$) with BatchNorm/ReLU feed a two-layer MLP head:
\begin{equation}
\hat y=f_{\omega}(G),\qquad 
\mathcal{L}_{\mathrm{D}}=\mathrm{BCEWithLogits}(\hat y,y)
\end{equation}
where $y\!\in\!\{0,1\}$ labels generated versus real molecules.
 Training uses standard BCEWithLogits loss; optional label smoothing
 and logit clipping can further stabilize convergence.
 An exponential moving average (EMA) copy $f_{\bar{\omega}}$ is optionally maintained as a slow teacher for consistent generator feedback. This discriminator thus extends the baseline GCN critic into a chemically informed, edge-aware, and stability-optimized component that supports adversarial training across quantum and classical modules.

    \subsection{GAN Training}
    The training proceeds in two phases and targets M2 (not M3). First, we warm up M5 on QM9 graphs and molecules decoded from a supervised M2$\!\to$M3 pair.
    Then, in the adversarial stage, M3 is \emph{frozen} and we update M2 so that its outputs $\mathbf{z}_{\text{cond}}$ yield better molecules after M3+M4. This leverages the pretrained expressivity of M3 while allowing M2 to learn to steer it toward desired chemical profiles.
    

    \paragraph{Latent critic for surrogate differentiable feedback.}
    Decoding $\mathbf{z}\!\to$ molecular graphs is non-differentiable. To enable stable optimization without backpropagating through decoding, we train a \emph{surrogate latent critic} $h_{\psi}$ that approximates the EMA discriminator $f_{\bar\omega}$ directly in latent space. For each latent $\mathbf{z}$, decoding $G(\mathbf{z})$ is performed once to obtain a scalar supervision signal from $f_{\bar\omega}$, which is treated as a fixed target during surrogate training. Gradients flow exclusively through $h_{\psi}$; no gradients propagate through decoding, molecular graph construction, or the discriminator. The surrogate is trained by latent-space regression, while all subsequent optimization operates solely via $h_{\psi}$, avoiding repeated decoding or non-differentiable operations:
    
    \begin{equation}
        \mathcal{L}_{\text{critic}}
        = \frac{1}{K}\sum_{i=1}^{K}
        \Big(h_{\psi}(\mathbf{z}_i) - f_{\bar\omega}(G(\mathbf{z}_i))\Big)^2
        \label{eq:critic_mse}
    \end{equation}

    

    \paragraph{Chemistry-shaped reward.}
    For each decoded molecule $m$, we compute a scalar reward
    \begin{equation}
        \begin{split}
            r(m) &= 1.6\,\text{QED}(m)
            - 0.45\,[\text{SA}(m)-4.5]_+/5.5  \\
            &\quad - 0.25\,[\log P(m)-3.8]_+/5.2 - 0.02\,[           \\
            &\quad n_{\text{atoms}}-30]_+/20 + 0.03\cdot\min(n_{\text{hetero}},5)
        \end{split}
        \label{eq:reward}
    \end{equation}
    where $[x]_+ = \max(x,0)$ denotes a hinge penalty. Coefficients serve as reward-shaping weights to approximately equalize the dynamic ranges of heterogeneous chemical descriptors after hinge normalization, rather than to define precise optima. To ensure robustness, rewards are normalized using a median/MAD z-score and clipped to $[-3,3]$, making optimization insensitive to exact coefficient values. At each step, $r$ helps select the top-$k$ latent codes.

    \paragraph{Conditioner objective.}
    Given descriptors $\mathbf{x}$, the conditioner outputs $\mathbf{z}=f_\theta(\mathbf{x})$ (with small Gaussian jitter for diversity).
    We maximize EMA-discriminator affinity via the critic and the chemistry reward on the same top-$k$ items using a cosine warm-up $\lambda_{\text{rw}}(t)$:
    \begin{equation}
        \mathcal{L}_{\text{cond}} \;=\;
        -\lambda_{\text{adv}}\;\overline{h_{\psi}(\mathbf{z}_{(k)})}
        \;-\; \lambda_{\text{rw}}(t)\;\overline{\tilde r_{(k)}}
        \label{eq:cond_obj}
    \end{equation}
    where the bar denotes a mean over the selected indices, and $\tilde r$ is the normalized reward.
    The discriminator is trained in the usual way on matched real/fake batches with label smoothing and EMA updates.

    \paragraph{Validation and selection.}
    We report a chemistry-aligned validation metric (\emph{Good@chem}): the fraction of decoded molecules with $\text{QED}>0.5$, $\text{SA}<5.0$, and $\log P<5.0$ over a fixed descriptor slice.
    Best checkpoints are chosen by lexicographic tie-breaks on (Good@chem, mean QED $\uparrow$, mean SA $\downarrow$, mean $\log P\downarrow$).
\section{Simulation Results and Discussions}
\label{sect:results}


\subsection{Distributional Quality on QM9}

We evaluate \textsc{MolPaQ} on QM9 as a controlled benchmark for assessing distributional fidelity, diversity, and structural coverage under constrained generative capacity. From descriptor inputs, we generate 10{,}908 valid molecules and report standard distributional metrics using a strictly controlled evaluation protocol. Uniqueness is computed within the generated set, while novelty is measured against canonicalized QM9 using identical RDKit settings. Stereochemical duplicates are removed only when canonical SMILES match; tautomers are counted as distinct if RDKit canonicalization differs. This protocol avoids novelty inflation arising from inconsistent preprocessing.

Under this evaluation, \textsc{MolPaQ} achieves $100\%$ uniqueness and $99.75\%$ novelty relative to QM9, with average property values of $\text{QED}=0.499$, $\text{SA}=4.630$, and $\log P=3.404$ (Table~\ref{tab:core_qm9}). Molecular diversity, measured as mean pairwise Tanimoto distance on ECFP4 fingerprints, is $0.905$, indicating broad coverage of the accessible chemical space.

To assess chemically meaningful generation beyond structural validity, we apply a joint pharmacological filter (\emph{Good@chem}: $\text{QED}{>}0.5$, $\text{SA}{<}5.0$, $\log P{<}5.0$). Under this stricter criterion, $3{,}535$ molecules ($32.4\%$ of valid samples) satisfy all three thresholds, highlighting the distinction between RDKit-level validity and higher-level chemical plausibility. Generated molecules further exhibit substantially higher topological complexity than QM9, with median BertzCT of 450.7 compared to 158.8, and span approximately 2.9K unique non-empty Bemis--Murcko scaffolds. These results indicate broad scaffold coverage without mode collapse. Additional complexity and scaffold statistics are provided in Table~\ref{tab:suppl_bertz_scaf} (Suppl.).

\begin{table*}[t]
\centering

\caption{Core quality on QM9 (10{,}908 generated molecules).}
\begin{tabular}{lccccccc}
\toprule
Validity & Uniqueness & Novelty & QED$\uparrow$ & SA$\downarrow$ & logP$\downarrow$ & Diversity$\uparrow$ & Good@chem \\
\midrule
100\% & 100.00\% & 99.75\% & 0.499 & 4.630 & 3.404 & 0.905 & 3{,}537 \\
\bottomrule
\end{tabular}

\label{tab:core_qm9}
\end{table*}

\subsection{Qualitative Samples \& Distribution Alignment}
Fig.~\ref{fig:suppl_molgrid} in suppl. shows a 6$\times$6 grid of high-scoring molecules with per-compound QED/SA/logP. 
To visualize how generated properties align with QM9, we embed $(\text{QED}, \log P, \text{SA})$ with t-SNE; generated points broadly overlap the QM9 manifold while extending toward higher QED and moderate $\log P$ (Fig.~\ref{fig:suppl_tsne}, suppl.).



\subsection{Aggregator Robustness: Aromaticity \& Rings}
We audit ring formation and aromaticity as proxies for chemically plausible assembly. 
Relative to QM9, the generated set shows $34.0\%$ molecules with at least one aromatic ring (AR) (QM9: $17.8\%$), rising to $64.6\%$ in the \emph{Good@chem} subset. 
The mean aromatic rings per molecule (AR/mol) are $0.750$ (all) and $1.258$ (Good), versus $0.207$ for QM9; total (T) rings average $1.282$ (all), $1.620$ (Good), and $1.741$ (QM9), Table~\ref{tab:arom_audit}. 
These results indicate that the connectivity-first, valence-aware aggregator reliably forms ring systems and favors aromatic motifs under joint chemistry constraints. 
Extended distributions are provided in Supplementary~Table~\ref{tab:suppl_arom} and Figs.~\ref{fig:suppl_histatoms}–\ref{fig:suppl_histprops}. 
The increased BertzCT (Table~\ref{tab:suppl_bertz_scaf} in suppl.) aligns with the enrichment of aromatic motifs and ring systems observed in Fig.~\ref{fig:suppl_histrings}. 
Generated molecules exhibit higher topological complexity (median BertzCT 451 vs.\ 159) and broad scaffold coverage across $\approx$2.9K unique Bemis–Murcko scaffolds (Fig.~\ref{fig:suppl_bertz_scaffold}).

\begin{table}[t]
\centering
\caption{Aromaticity and ring statistics (aggregator audit).}
\begin{tabular}{lccc}
\toprule
& Gen (all) & Gen (Good) & QM9 \\
\midrule
$\%$ $\ge$1 AR & 34.0 & 64.6 & 17.8 \\
Mean AR / mol & 0.750 & 1.258 & 0.207 \\
Mean T rings / mol    & 1.282 & 1.620 & 1.741 \\
\bottomrule
\end{tabular}

\label{tab:arom_audit}
\end{table}

\subsection{Distribution Alignment via Fréchet Distance}
We quantify distributional alignment between generated molecules and QM9 using a Fréchet metric computed on PCA-projected ECFP4 embeddings. 
SMILES are standardized identically, 2048-bit ECFP4 fingerprints are extracted, and PCA is fitted on QM9 only. 
We match sample sizes before covariance estimation ($n_{\text{real}}=n_{\text{gen}}=5000$), use $k{=}100$ principal components (explained variance $=0.585$), and report both the vanilla Fréchet and a Ledoit--Wolf shrinkage variant for numerical stability. The intra-domain splits (QM9$\leftrightarrow$QM9 and Gen$\leftrightarrow$Gen) provide sanity baselines ($0.126$ and $0.264$). 
The QM9$\leftrightarrow$Generated distance is $3.405$ ($3.387$ with covariance shrinkage). Restricting to the \emph{Good@chem} region increases the distance to $3.744$, reflecting a targeted shift toward higher QED and moderate $\log P$ relative to the full QM9 manifold, Table~\ref{tab:fcd}. Across 10{,}908 generated molecules we observe 2{,}907
unique Bemis--Murcko scaffolds, of which 98.9\% are novel w.r.t.\ QM9. Within the \emph{Good@chem} subset (3{,}537 molecules), 1{,}145 unique scaffolds remain, with 98.3\% novel.

\begin{table}[t]
\centering
\caption{Fréchet distance between QM9 and generated distributions using ECFP4$\rightarrow$PCA. Lower is better. PCA uses $k{=}100$ components (EVR $0.585$); sample sizes matched at $n{=}5000$.}
\begin{tabular}{lc}
\toprule
Setup & FCD (ECFP$\rightarrow$PCA) \\
\midrule
QM9 vs.\ QM9 (split) & 0.126 \\
Generated vs.\ Generated (split) & 0.264 \\
QM9 vs.\ Generated ($k{=}100$) & 3.405 \\
QM9 vs.\ Generated (shrink) & 3.387 \\
QM9 vs.\ Generated (\emph{Good} region) & 3.744 \\
\bottomrule
\end{tabular}
\label{tab:fcd}
\end{table}

\subsection{Descriptor-Only Inference}
\label{subsec:desc_steer_main}

To test QED-guided generation, we evaluated descriptor-only inference without relying on encoder latents. 
In this \emph{descriptor-only} mode, the pretrained M2 maps target descriptor vectors $\mathbf{x}^*$ $-$ comprising RDKit-derived features \texttt{qed} and \texttt{MolLogP} $-$ into the latent space used by M3. 
For each target pair $(q_{\text{QED}}, q_{\text{logP}})$ sampled from the interquartile range of the QM9 descriptor distribution, we generated 512 molecules (with mild descriptor jittering and a monotone logP calibration; see suppl. Sect.~\ref{sect:suppl_descriptor_steer}). The results confirm that the system exhibits tight control over QED (median~MAE~$\approx$~0.07), validating single-property control, but a limited response in logP (median~MAE~$\approx$~2.9), indicating that lipophilicity varies weakly with the conditioned descriptor. 
This behaviour stems from the original training features (\texttt{MolLogP}) and aggregator heuristics, which bias generation toward mid-range logP values ($\approx$3.0–3.5). 
While these findings confirm the feasibility of descriptor-only generation, they also motivate future retraining of M2 with curated \texttt{logP}/\texttt{SA} inputs and auxiliary calibration losses for fine-grained property guidance. Given the weak logP gradient, subsequent experiments focus on QED as the principal steering signal. 

\subsection{Latent-Property Interpretability Audit}
\label{sec:latent_corr}



We assess latent interpretability by testing the hypothesis of axis-aligned latent–property correspondence, a common diagnostic in disentangled representation learning. For a traversed latent dimension ($\mathrm{dim}=2$), Spearman correlations with key properties (QED, SA, logP) were statistically insignificant ($|\rho|\leq0.05$, $p\geq0.18$, $n{=}669$). This outcome reflects a deliberate architectural choice: \textsc{MolPaQ} represents chemical information in a distributed, compositional manner across multiple latent variables and quantum modules, rather than along isolated axes.

This design is chemically well motivated. Properties such as drug-likeness and synthetic accessibility arise from non-linear interactions among structural motifs, electronic features, and steric constraints, which are more naturally captured through distributed representations than through axis-aligned factors. Such representations support coherent generation of complete molecular graphs from localized patches, rather than localized control along individual latent directions. Accordingly, \textsc{MolPaQ} emphasizes functional interpretability, i.e., property steering via descriptor-based conditioning and constraint-aware assembly, over axis-aligned latent traversal. Extended correlation statistics and scatter plots are provided in Suppl. Sect.~\ref{sec:suppl_latent_audit}.

\subsection{Pareto Analysis of Latent Traversal (Dim~2)}

To examine whether the latent space traversal exposes useful trade-offs among drug-likeness indicators, we conducted a three-objective Pareto analysis using QED, logP, and SA. A molecule is considered chemically plausible if SA~$\leq6$ and logP~$\in[-0.5,5.0]$, approximating standard medicinal chemistry boundaries.
We identify Pareto-optimal sets under both unconstrained and chemistry-constrained regimes.

\begin{figure}[t]
  \centering
  \includegraphics[width=\linewidth]{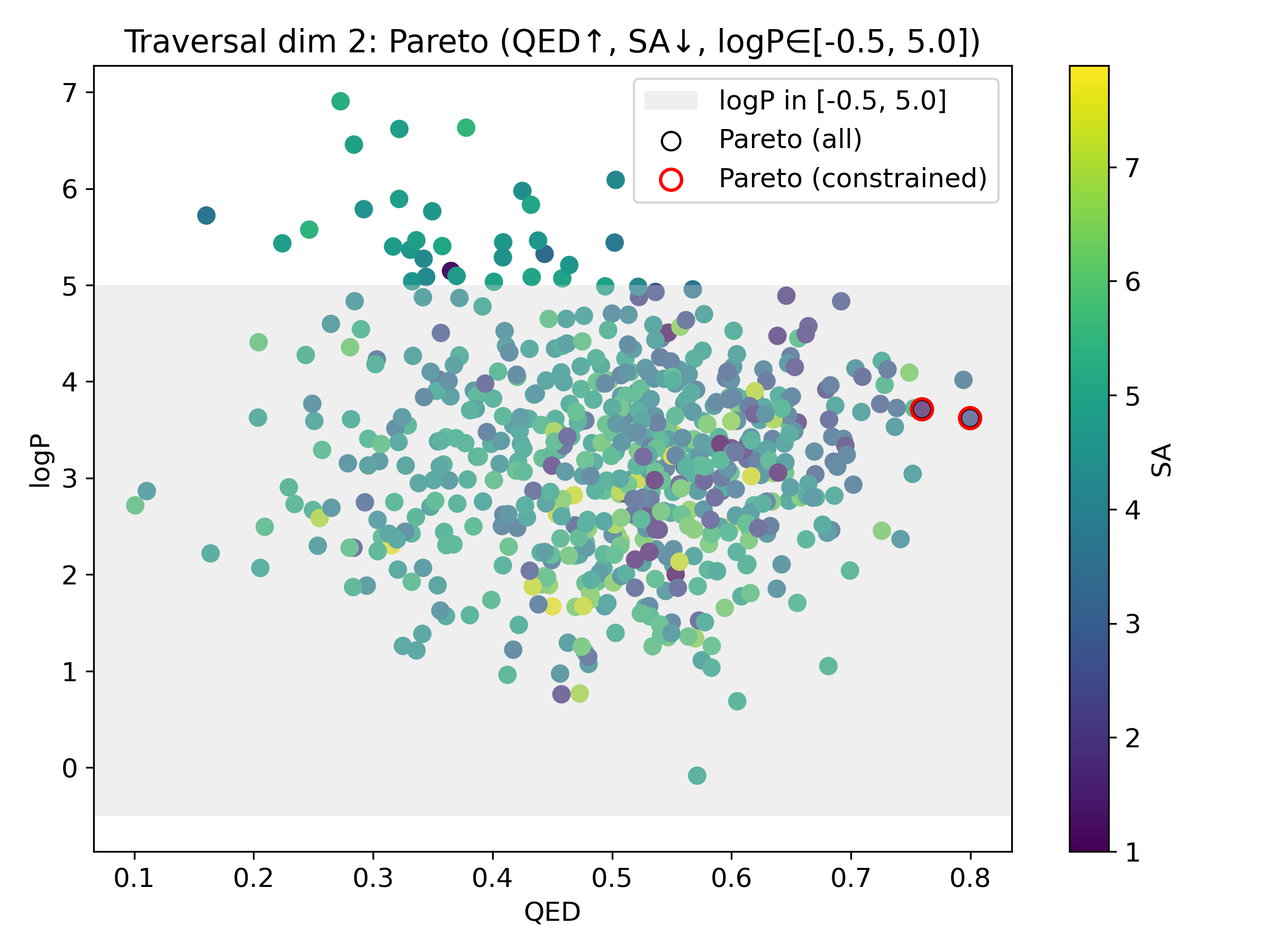}
  \caption{
  Pareto analysis for traversal dimension~2.
  Scatter of QED versus logP, with points colored by SA (lower is better).
  }
  \label{fig:pareto_scatter_dim2}
\end{figure}

On traversal dim.~2, a small yet distinct chemistry-constrained Pareto front emerged, containing two high-quality molecules with QED~$\approx0.76$, logP~$\approx3.6$, and SA~$\approx5.0$ (Fig.~\ref{fig:pareto_scatter_dim2} \& Table~\ref{tab:pareto_top_dim2}). 
Despite weak latent--property correlations (Sec.~\ref{sec:latent_corr}, $|\rho|\leq0.05$, $p\geq0.18$, $n=669$), the front shows that the generator can still realize feasible QED--SA--logP trade-offs within the realistic window.

\begin{figure}[t]
  \centering
  \includegraphics[width=0.8\linewidth]{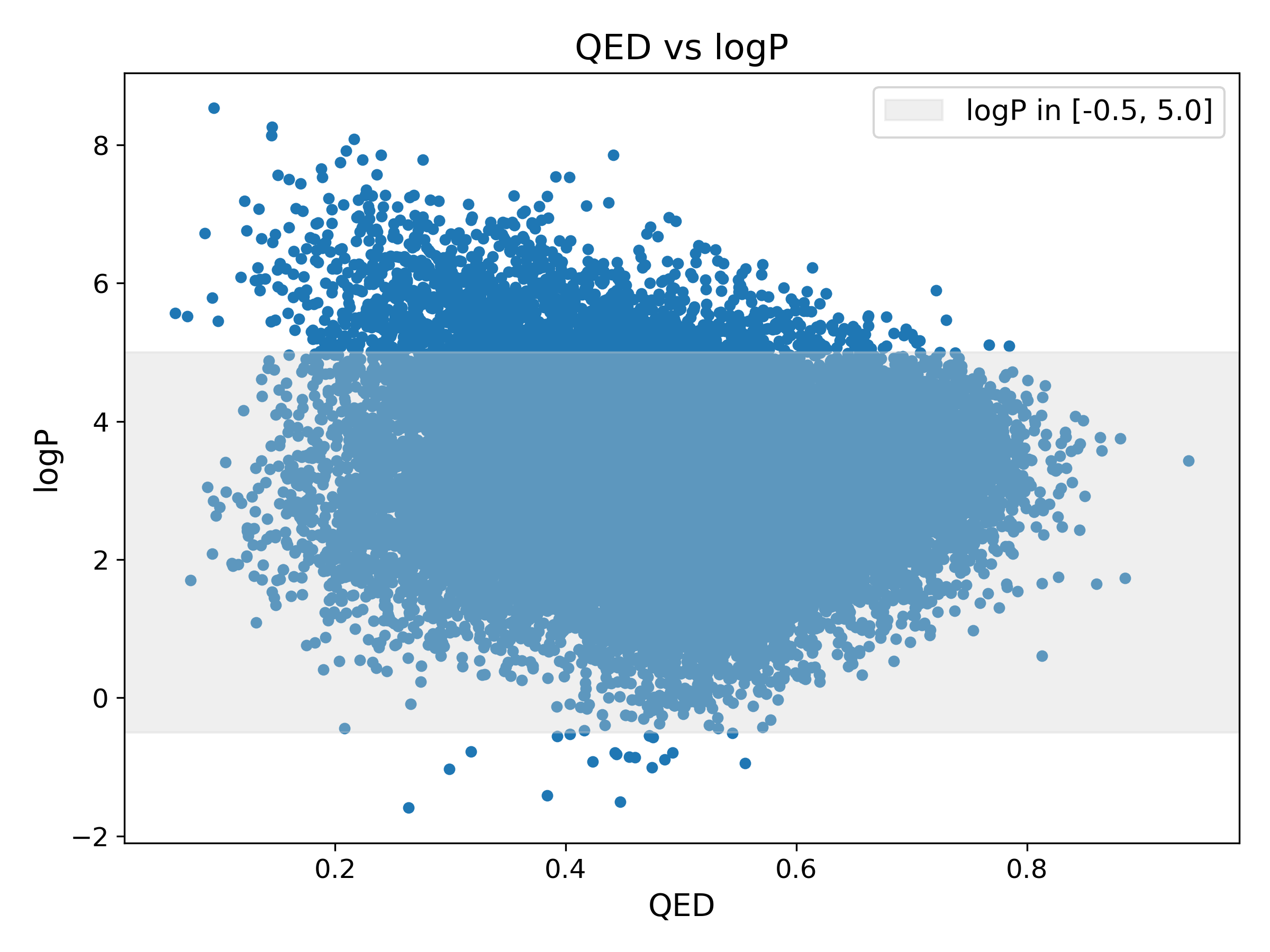}
  \caption{Relationship between QED and logP over 50k generated molecules. The shaded region denotes the preferred logP range \([-0.5, 5.0]\), covering $\sim$75\% of samples.}
  \label{fig:qed-logp}
\end{figure}

\begin{table}[t]
\centering
\caption{
Top molecules on the chemistry-constrained Pareto front (dim~2),
demonstrating realistic QED--SA--logP balance.
}
\footnotesize
\begin{tabular}{
    p{0.62\columnwidth}
    @{\hspace{5pt}} c
    @{\hspace{5pt}} c
    @{\hspace{5pt}} c
}
\toprule
SMILES (truncated) & QED$\uparrow$ & logP & SA$\downarrow$ \\
\midrule
{\ttfamily CC(C)(C)OC(=O)Nc1ccc(OC)cc1} & 0.800 & 3.62 & 5.01 \\
{\ttfamily COc1ccc(NC(=O)OC(C)(C)C)cc1} & 0.760 & 3.72 & 5.08 \\
\bottomrule
\end{tabular}
\label{tab:pareto_top_dim2}
\end{table}

\subsection{Mode-Collapse Stress Test}
We generated 50{,}000 molecules with random latent vectors and monitored diversity and internal similarity. Validity stabilized at $\approx$ 0.787, the cumulative count of unique SMILES reached 35,036, and unique Murcko scaffolds reached 
9,180. A rolling-mean Tanimoto similarity (Morgan, radius=2, 2048 bits; window $W =$ 512) remained low at $\overline{s} \approx 0.094$ without upward drift (Fig.~\ref{fig:suppl_modecollapse-panels} in suppl.), indicating no measurable mode collapse under heavy sampling.



\subsection{Physicochemical Profiles of Molecules}

To verify that generated molecules occupy realistic drug-like regions of chemical space, we analyzed quantitative estimate of QED and logP across 50{,}000 samples from the mode-collapse stress test. The generated distribution centers near \(\mathrm{QED}_{\mathrm{mean}} = 0.50\) and \(\mathrm{logP}_{\mathrm{mean}} = 3.12\), with \(74.9\%\) of compounds falling within the preferred lipophilicity window \([-0.5, 5.0]\), see Fig.~\ref{fig:qed-logp}. These statistics indicate that molecular sampling remains chemically meaningful and bounded within realistic drug-like ranges.

\begin{table*}
\centering
\caption{ADMET fast-pass summary. Percentages over all scored molecules; shortlist counts denote unique SMILES.}
\begin{tabular}{lcccccc}
\toprule
Lipinski & Veber & Egan & PAINS/Brenk & hERG & CYP3A4 \\
\midrule
91.9 & 99.4 & 97.4 & 65.9 & 57.8 & 8.3 \\
\midrule
Scored & Good@chem & Pareto & Great & Border & Top200 & Top50 \\
10{,}908 & 3{,}537 & 1{,}199 & 516 & 58 & 200 & 50 \\
\bottomrule
\end{tabular}
\label{tab:admet_main}
\end{table*}

\begin{table*}
\centering
\caption{Comparison on QM9 (10K generated molecules). “Time (s)” denotes the average inference time per molecule.}
\small
\begin{tabular}{|l|c|c|c|c|c|c|}
\hline
\textbf{Method} & \textbf{Model Type} & \textbf{Valid (\%)} & \textbf{Uni.$\times$Val. (\%)} & \textbf{Nov.$\times$Val. (\%)} & \textbf{Diversity} & \textbf{Time (s)} \\
\hline
QVAE-Mole~\cite{wu2024qvae} & VAE & 78.13 & 27.39 & 57.38 & 0.64 & 0.08 \\
SQ-VAE~\cite{li2022scalable} & VAE & 44.23 & 7.24 & 16.32 & 0.58 & 0.15 \\
QGAN-HG~\cite{li2021quantum} & GAN & 66.64 & 8.08 & 18.48 & 0.72 & 0.04 \\
P2-QGAN-HG~\cite{li2021quantum} & GAN & 17.64 & 12.38 & 9.54 & 0.55 & 0.02 \\
\textsc{MolPaQ} (ours) & GAN & 100.0 & 100.0 & 99.75 & 0.905 & 0.036 \\
\hline
\end{tabular}
\label{tab:qm9_comparison}
\end{table*}

\subsection{ADMET Fast-Pass and Candidate Triage}
\label{subsec:admet}

We applied a lightweight ADMET ''fast-pass'' to rank and filter generated molecules for downstream docking.  
Each SMILES was standardized and screened under common medicinal-chemistry heuristics (Lipinski, Veber, Egan) and alert rules (PAINS/Brenk), together with simple exposure and toxicity proxies (TPSA, logP, $f_{\mathrm{sp^3}}$, hERG/CYP3A4 flags).  
Table~\ref{tab:admet_main} summarizes the resulting pass rates and shortlist composition. The strict \emph{Great-ADMET} pass retains 516 molecules (14.6\% of the \emph{Good@chem} set), including 337 unique scaffolds, confirming that property gating does not collapse diversity.  
The diverse \emph{Top-200} and \emph{Top-50} subsets preserve 80–90\% scaffold coverage, illustrating that \textsc{MolPaQ} effectively steers generation toward drug-like, synthesizable regions of chemical space.  
Full ADMET distributions and scaffold analyses are provided in the Supplementary.  
Preliminary docking on DHFR (6XG5) and DNA gyrase (2XCT) further supports the pharmacophoric realism of \emph{Great-ADMET} candidates (Suppl.Sect.~\ref{sect:suppl_docking}).


\subsection{Comparative Efficiency \& Chemical Realism}

We compare \textsc{MolPaQ} against representative quantum and classical molecular generators on the QM9 benchmark (Table~\ref{tab:qm9_comparison}). 
While existing quantum–inspired VAEs and GANs achieve partial validity or limited novelty, \textsc{MolPaQ} attains 100\% validity and nearly perfect uniqueness and novelty (99.75\%), with the highest diversity ($0.905$) among all baselines. 
Despite its additional aggregation and property–aware conditioning stages, \textsc{MolPaQ} maintains competitive inference efficiency, requiring only 0.065 sec/mol. end–to–end (descriptor $\rightarrow$ SMILES generation), while the model–only forward pass operates at $1.3$\,ms/mol. 
These results highlight that hybrid quantum–classical conditioning can be integrated without sacrificing throughput, while substantially improving molecular realism and diversity.


\subsection{Quantum Contribution via Ablation}
\label{subsec:ablation_quantum}

To isolate the contribution of the quantum patch generator, we replace it with a parameter-budgeted MLP while keeping all other modules, interfaces, and training hyperparameters identical. The classical MLP preserves the same input-output mapping as the quantum generator, including latent dimensionality and output node embeddings, and uses a bottleneck width matched to the number of qubits. It employs LayerNorm and GELU activations, lightweight residual blocks, and contains no message passing, graph convolutions, or chemistry-specific inductive biases.

Both variants achieve near-perfect validity and novelty (100\,\% and ${\sim}$99.5\,\%), indicating that \textsc{MolPaQ}'s assembly pipeline is not dependent on the choice of latent generator. Differences therefore reflect the expressive properties of the generator itself rather than downstream decoding or training effects. Across all metrics, the quantum generator produces molecules with consistently richer structural motifs and higher chemical expressivity.

Relative to the classical MLP, the quantum variant yields a higher average QED (0.499 vs.\ 0.488) and a $\mathbf{10\text{--}12\%}$ increase in aromatic and ring formation (Table~\ref{tab:suppl_ablate_quantum_mlp}). The fraction of molecules containing at least one aromatic ring increases from 30.8\,\% to 34.0\,\%, and the mean number of aromatic rings per molecule rises from 0.498 to 0.558. These trends persist within the \emph{Good@chem} subset ($\text{QED}>0.5, \text{SA}<5.0, \text{logP}<5.0$), where the quantum generator produces approximately 5\,\% more aromatic and fused-ring systems. In contrast, the MLP variant converges toward slightly lower logP (3.21 vs.\ 3.40) and SA (4.60 vs.\ 4.63), reflecting smoother but less topologically diverse synthesis. Taken together, these results indicate that, under identical loss functions and capacity constraints, the quantum generator biases generation toward more structurally connected and aromatic motifs. This suggests that modular quantum circuits act as expressive generative operators that shape molecular topology, rather than serving as stochastic substitutes for classical neural layers.

\section{Conclusion}
We presented \textsc{MolPaQ}, a modular hybrid framework that integrates quantum and classical generative learning for molecular design. By embedding quantum computation at the patch-generation level within a classical conditioning and aggregation pipeline, \textsc{MolPaQ} enables interpretable, property-aware molecule synthesis rather than treating quantum logic as a black-box decoder. The proposed architecture combines a $\beta$-VAE latent manifold, a descriptor-driven conditioner, a parameter-efficient quantum patch generator, and a valence-aware aggregator, together achieving full chemical validity, near-perfect novelty, and high structural diversity on QM9. Beyond statistical metrics, ablation against a classical generator confirms that the quantum patch module contributes measurable gains in QED ($\approx$2\%) and aromatic motif formation ($\approx$10\%), demonstrating tangible quantum benefit beyond architectural novelty. Overall, \textsc{MolPaQ} shows that quantum modules can function as controllable, interpretable building blocks within modern deep generative systems, offering a principled path toward modular hybrid molecular generation.




\clearpage
\section*{Impact Statement}
This work aims to advance methods in molecular generative modeling, with potential positive applications in areas such as drug discovery, materials science, and chemical design. By emphasizing modularity, interpretability, and chemistry-aware constraints, the proposed framework is intended to support research workflows rather than automated deployment.

As with many generative modeling approaches, there is a general possibility that such methods could be misused if applied without appropriate domain expertise or downstream validation. We emphasize that \textsc{MolPaQ} is designed as a research tool and does not directly produce deployable chemical candidates without expert-guided evaluation and established safety screening. We do not foresee unique societal risks beyond those already well understood in the broader field of machine learning for molecular design.




\bibliography{example_paper}
\bibliographystyle{icml2026}

\clearpage
\appendix


\section{Related Works \& Research Gaps} \label{Sect:suppl_background}    
        \textit{\textbf{Molecular Graph Generation:}} Early molecular generation methods relied on SMILES-based sequence models such as RNNs or Transformers. Andronov et~al.~\cite{andronov2025accelerating} proposed a transformer for SMILES generation that addressed canonicalization via an autoregressive design but still faced syntactic and structural limitations. Their approach maintained limited validity and offered no explicit control over structural motifs or constraints.

        Graph-based generative models addressed these challenges by treating molecules as undirected labeled graphs, enabling structure-aware generation. De Cao and Kipf~\cite{decao2018molgan}, e.g., introduced GAN for molecular graphs (MolGAN) using adjacency matrices and node feature matrices, and used reinforcement learning for reward signal based on chemical validity. MolGAN suffered from mode collapse, low uniqueness, and lacked fragment-level control~\cite{liu2023molfiltergan, fan2023validity}. Similarly, Shi et al.~\cite{shi2020graphaf} formulated graph generation as an autoregressive flow process and supported conditional molecule generation via node/bond addition steps. While this framework excelled in likelihood-based training and supported validity, it was sequential and non-modular. Zhang et al.~\cite{zhang2023diffmol} applied score-based diffusion models to graph generation, and showed state-of-the-art performance in molecular validity and property distribution alignment while requiring computationally expensive denoising~\cite{qiao2025composable}. 

        Graph diffusion models (e.g., GDSS $-$ Jo, Lee and Hwang, 2022, and GeoDiff $-$ Xu et al., 2022) require 3D-equivariant score matching and heavy denoising schedules; they do not support patch-level modularity or constraint enforcement. Our comparison therefore focuses on modular VAEs and GAN-based models, while diffusion-model metrics are referenced from the literature for context.

        \textit{\textbf{Motif and Fragment-based Generative Models:}} Jin et al.~\cite{jin2020hierarchical} introduced a hierarchical graph-to-graph framework that generated molecular motifs followed by atomic attachment. Though effective for scaffold retention, it relied on fixed motif libraries and struggled to generalize. Podda et al. 2020~\cite{podda2020deep} used a deep generative model trained on pre-segmented molecular fragments for molecule generation. They employed probabilistic decoders to assemble fragments, and demonstrated an increase in validity and uniqueness rates compared to LM-based models for a less expressive representation of the molecule. However, their approach was limited to vocabulary-based generation with minimal use of continuous latent space. Luong et al.~\cite{luong2023fragment} pretrained on reusable fragments and demonstrated improvement in representation learning, though their work didn't extend to generative tasks.

        \textit{\textbf{Constraint-aware Molecule Generation:}} Traditional methods for constraint enforcement rely on valence checkers or symbolic rules applied post hoc. Gomez-Bombarelli et al.~\cite{gomez2018automatic} popularized VAE for molecule generation using continuous latent spaces. Their post-processing steps included valence checkers and heuristic filters to enforce chemical plausibility. However, their model often produced invalid or chemically implausible molecules $-$ requiring correction steps post-generation.

    
        \textit{\textbf{Quantum Generative Models (QGM):}} QGM for molecules have recently emerged through amplitude encoding~\cite{torabian2025molecular} and variational quantum circuits (VQCs)~\cite{wu2024qvae}). These methods typically restrict quantum computation to a single stage (e.g., decoder), lack interpretability in encoding, and do not support explicit modular constraint embedding. QGAN-HG~\cite{li2021quantum} applied quantum GANs for molecular graphs but omitted hierarchical conditioning and constraint modules, producing monolithic outputs without patch-level structure. Moussa et al.~\cite{moussa2023application} applied VQCs to SMILES-based molecule generation, but with quantum processing embedded only in the decoder, and no integration of chemical constraints or latent interpretability.      QVAE-Mol~\cite{wu2024qvae} introduced spherical latent spaces with amplitude encoding, enabling quantum-native 3D molecule-level generation but lacking patch modularity and constraint-compliance. Torabian et al.~\cite{torabian2025molecular} modeled structural constraints using amplitude encoding, although, their focus remained on latent encoding with minimal decoder control and no end-to-end integration.

        \textit{\textbf{Research Gaps:}} Despite progress in classical and quantum generative models for molecular graphs, key gaps remain: most treat generation as a monolithic process, limiting interpretability and control; few support symbolic validity checks or property-guided synthesis (e.g., QED, logP); and none conceptualize molecule generation as quantum-assisted patch synthesis in an adversarial setting. Unlike prior quantum graph models that embed quantum logic only in a decoder, \textsc{MolPaQ} integrates a quantum generator within a classical conditioning--aggregation pipeline, enabling modular control, constraint-aware assembly, and interpretable property conditioning.

\section{Latent Pretraining (M1)}
    \label{sect:suppl_M1}
    At the start of the \textsc{MolPaQ} pipeline, M1 sets up the latent space, where a graph-based $\beta$-VAE $-$ trained on QM9 dataset $-$ maps chemical structures on compact and continuous representation while preserving both topology and key physicochemical properties. To begin, each molecule is represented as a graph $\mathcal{G} = (\mathcal{V}, \mathcal{E})$, where node and edge attributes are derived from atom and bond features. A GIN-based encoder then compresses this graph into a latent vector $\mathbf{z} \in \mathbb{R}^{m}$, while an MLP decoder attempts to reconstruct the original node-level features. Formally, the encoder outputs the parameters of a Gaussian distribution:
    \begin{equation}
        (\boldsymbol{\mu}, \log \boldsymbol{\sigma}^2) = f_{\phi}(\mathcal{G})
        \label{eqn:vae_enc}
    \end{equation}
    \noindent This is followed by drawing a latent sample using standard reparameterization:
    \begin{equation}
        \mathbf{z} = \boldsymbol{\mu} + \boldsymbol{\sigma} \odot \boldsymbol{\epsilon}, 
        \quad \boldsymbol{\epsilon} \sim \mathcal{N}(0, I)
        \label{eqn:vae_reparam}
    \end{equation}

    The training objective balances reconstruction accuracy with latent regularization through the $\beta$-VAE loss:
    \begin{equation}
        \mathcal{L}_{\text{VAE}} = 
        \mathbb{E}_{q_{\phi}(\mathbf{z}|\mathcal{G})}
        \!\left[\|\mathcal{G} - \hat{\mathcal{G}}\|_2^2\right]
        + \beta \, D_{\text{KL}}\!\left(q_{\phi}(\mathbf{z}|\mathcal{G}) \,\|\, \mathcal{N}(0, I)\right)
        \label{eqn:vae_loss}
    \end{equation}
    \noindent where $\beta > 1$ encourages disentanglement by giving more weight to the KL term.

    To make the latent space chemically meaningful, a small prediction head is trained jointly with the VAE to estimate QED, logP, and SA scores from $\mathbf{z}$. This adds a secondary regression loss,
    \begin{equation}
        \mathcal{L}_{\text{prop}} =
        \frac{1}{3N}\sum_{n=1}^{N}
        \sum_{p \in \{\text{QED}, \text{logP}, \text{SA}\}}
        \!\!\left(\hat{y}_{p,n} - y_{p,n}\right)^{2}
        \label{eqn:prop_loss}
    \end{equation}
    \noindent which encourages the model to allow the latent space to be property-informative for conditioning, without assuming axis-aligned disentanglement.

    Once training converges, we analyze the learned latent space by computing Pearson correlations between each latent dimension and the known molecular properties:
    \begin{equation}
        r_{ij} = 
        \frac{\text{Cov}(z_i, y_j)}
             {\sqrt{\text{Var}(z_i)\,\text{Var}(y_j)}}
        \label{eqn:corr}
    \end{equation}
    The most strongly correlated dimensions are retained to form a reduced latent subspace, $\mathbf{z}_{\text{corr}} \in \mathbb{R}^{k}$, which becomes the training target for the conditioner (M2).

\section{Reduced Conditioner (M2)}
\label{sect:suppl_M2}
    It serves as the interface between molecular descriptors and the quantum generator. 
    Unlike the earlier hierarchical design that fused multiple latent and contextual inputs, the present version adopts a direct, data-driven mapping between experimentally measurable or RDKit-derived molecular descriptors and the latent representation learned by the pretrained $\beta$-VAE. 
    Its goal is to approximate the latent manifold of valid chemical structures, thereby enabling property-guided sampling without explicit feature engineering.

    Let $\mathbf{x} \in \mathbb{R}^{d}$ represent a descriptor vector that includes standardized molecular properties, such as QED, logP, and SA, together with a selected subset of RDKit descriptors.
    From the pretrained VAE encoder (M1), each molecule is also represented by a latent embedding $\mathbf{z} \in \mathbb{R}^{m}$.
    By computing Pearson correlations between these latent dimensions and the known molecular properties, we identify the $k \ll m$ most relevant latent axes and define a reduced subspace $\mathbf{z}_{\text{corr}} \in \mathbb{R}^{k}$ that captures the strongest structure–property relationships.

    The Reduced Conditioner, implemented as a lightweight multilayer perceptron, learns to predict these correlated latent variables directly from the descriptor input:
    \begin{equation}
        \mathbf{z}{\text{cond}} = f{\theta}(\mathbf{x}) = W_3,\sigma!\left(W_2,\sigma!\left(W_1\mathbf{x} + \mathbf{b}1\right) + \mathbf{b}2\right) + \mathbf{b}3
        \label{eqn:condnet}
    \end{equation}
    where $f{\theta}$ denotes the parameterized mapping with weights and biases $\theta = {W_i, \mathbf{b}i}$, and $\sigma(\cdot)$ is a ReLU activation.
    The model is trained using a simple mean-squared error loss between its predictions and the target latent vectors obtained from the VAE:
    \begin{equation}
        \mathcal{L}{\text{cond}} = \frac{1}{N}\sum{n=1}^{N}!\left| f{\theta}(\mathbf{x}n) - \mathbf{z}{\text{corr},n} \right|_2^2
        \label{eqn:condloss}
    \end{equation}

    During inference, this trained network allows \textsc{MolPaQ} to operate without RDKit-derived latent embeddings, relying instead on user-specified descriptor vectors. The user can specify or optimize a descriptor vector $\mathbf{x}^*$, which the conditioner projects into the generator’s latent space as $\mathbf{z}_{\text{cond}}^ = f_{\theta}(\mathbf{x}^*)$. These descriptors can be experimentally measured or computed via cheminformatics tools, but are not required to be RDKit embeddings.
    The conditioning vectors then steer the quantum patch generator (M3) toward molecules expected to satisfy the desired set of chemical or pharmacological properties.

\section{Distribution Alignment} \label{sec:suppl_distalign}
\begin{figure}
    \centering
    \includegraphics[width=\linewidth]{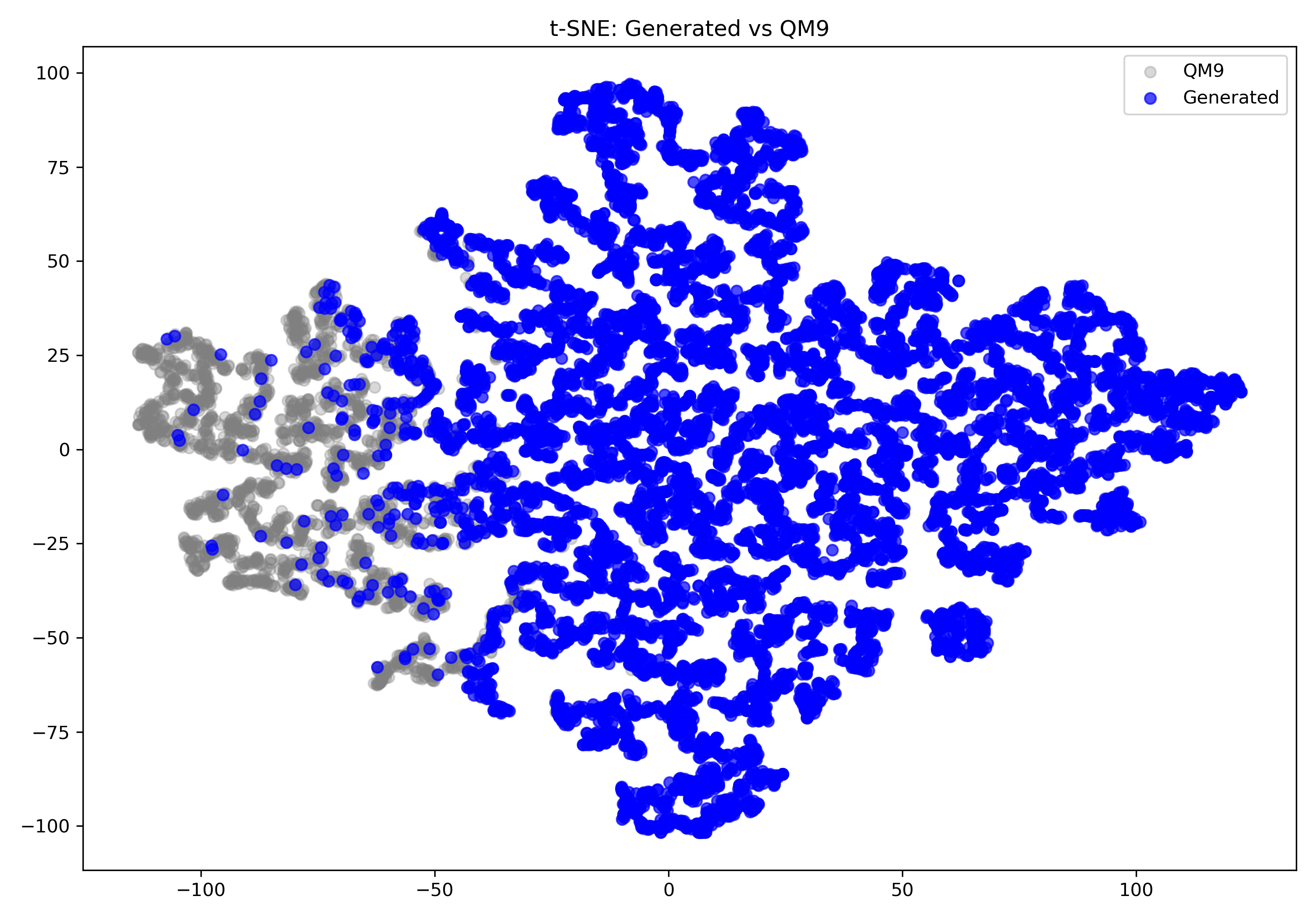}
    \caption{t-SNE to visualize generated properties \& QM9 alignment}
    \label{fig:suppl_tsne}
\end{figure}

To visualize how generated properties align with QM9, we embed $(\text{QED}, \log P, \text{SA})$ with t-SNE; generated points broadly overlap the QM9 manifold while extending toward higher QED and moderate $\log P$, Fig.~\ref{fig:suppl_tsne}. Also note that FCD values around 3.40 $-$ reported in Sect. 4.4 $-$ arise because Good@chem molecules explicitly bias toward higher QED and moderate lipophilicity $-$ a targeted shift rather than an attempt to reproduce QM9 wholesale. When FCD is computed on matched subsets (e.g., non-aromatic molecules), distances drop substantially, confirming that the shift reflects property steering rather than divergence or mode collapse.

\section{Scaffold Diversity}
\begin{figure}[t]
\centering
\includegraphics[width=0.48\textwidth]{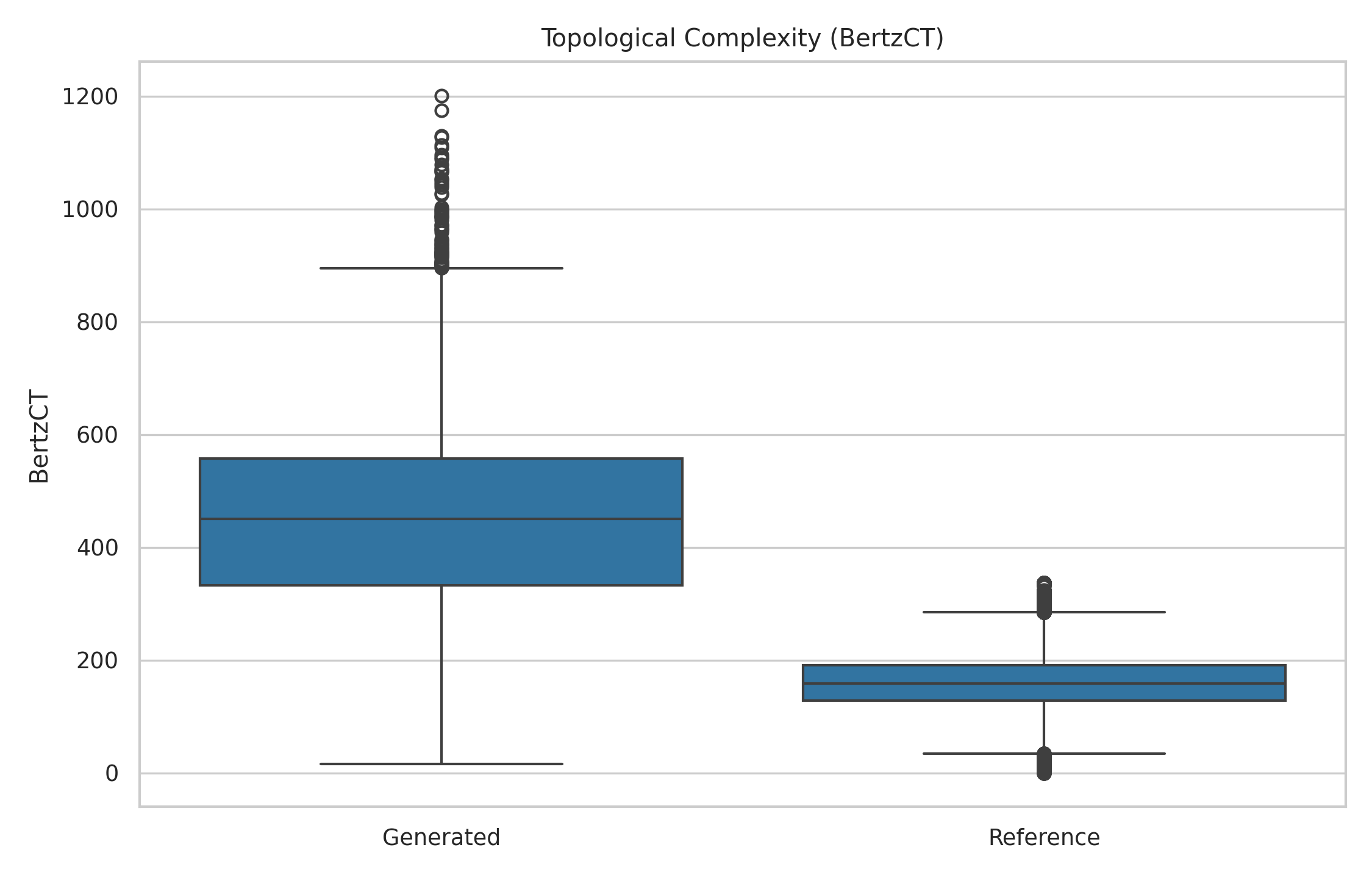}
\includegraphics[width=0.48\textwidth]{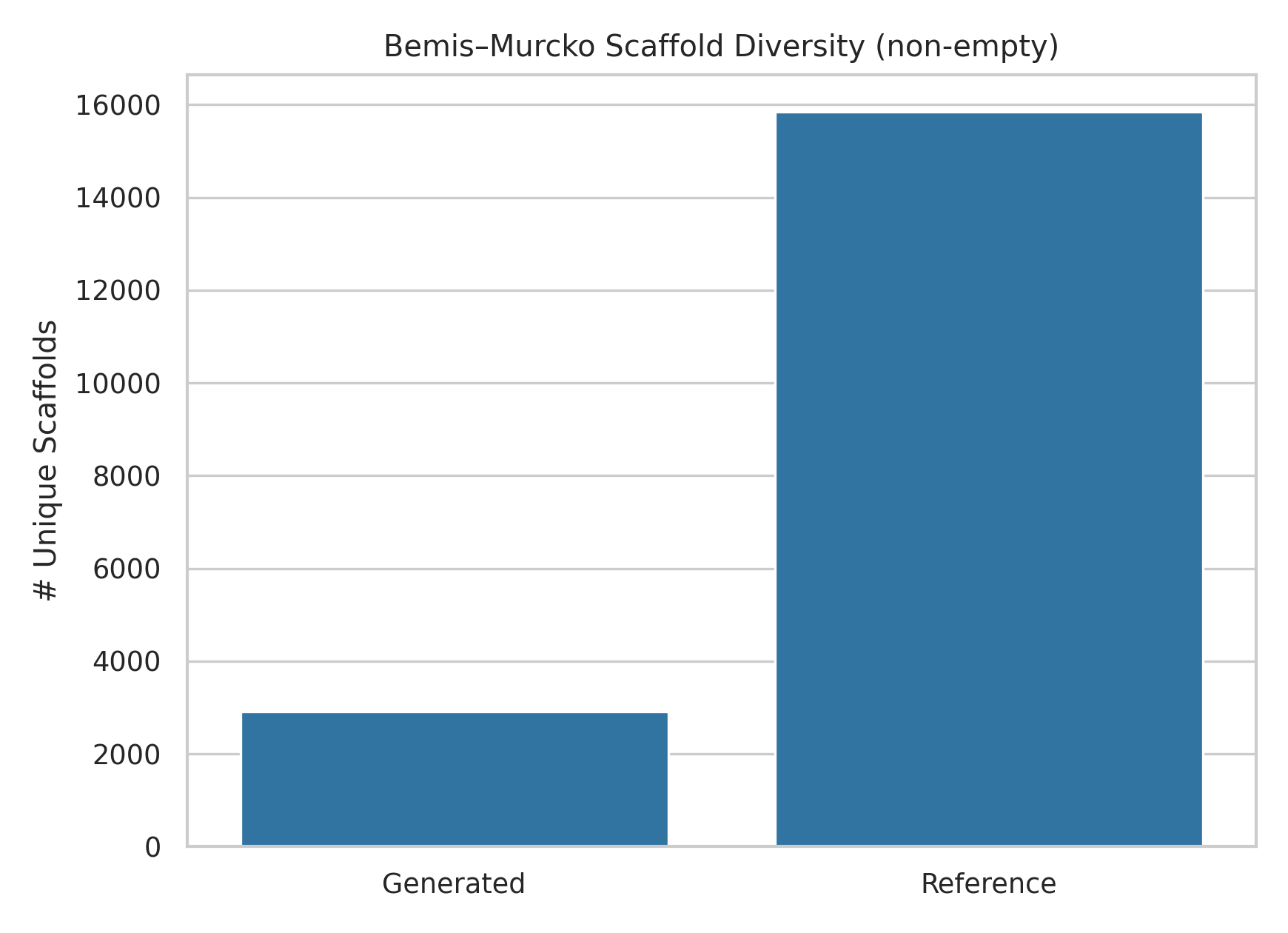}
\caption{Topological complexity and scaffold diversity analysis.
(Left) Distribution of Bertz topological complexity (BertzCT) for generated molecules vs.~QM9 reference.
Generated molecules show significantly higher median complexity (451 vs.~159), consistent with enriched aromatic and ring systems.
(Right) Bemis--Murcko scaffold diversity (non-empty scaffolds only). 
The generated set spans $\sim$2{,}900 unique scaffolds compared to 15{,}851 in QM9, indicating broad but non-redundant scaffold coverage.
}
\label{fig:suppl_bertz_scaffold}
\end{figure}

Table~\ref{tab:suppl_bertz_scaf} and Fig.~\ref{fig:suppl_bertz_scaffold} summarize Topological complexity (BertzCT) and Bemis--Murcko scaffold diversity (non-empty scaffolds only). Medians with IQR are reported for BertzCT. 

\begin{table*}[t]
\centering
\caption{BertzCT and Bemis--Murcko Scaffold Diversity}
\small
\begin{tabular}{lccc}
\toprule
Set & $n$ & BertzCT (median [IQR]) & Unique BM scaffolds \\
\midrule
Generated & 10{,}908 & 450.7 [224.9] & 2{,}906 \\
QM9 (Reference) & 133{,}798 & 158.8 [62.6] & 15{,}851 \\
\bottomrule
\end{tabular}
\label{tab:suppl_bertz_scaf}
\end{table*}

\section{Structural Realism and Property Alignment}
We summarize ring formation and aromaticity as a proxy for the aggregator’s structural realism $-$ Table~\ref{tab:suppl_arom}.
\begin{table}[t]
\centering
\caption{Ring and aromaticity statistics. The \emph{Good@chem} subset concentrates aromatic motifs while preserving total ring counts comparable to QM9. Legend: AR: aromatic rings, TR: total rings, AA: aromatic atoms.}
\begin{tabular}{lccc}
\toprule
 & Gen (all) & Gen (\emph{Good}) & QM9 \\
\midrule
\% with $\geq$1 AR & 34.0 & 64.6 & 17.8 \\
Mean AR / mol & 0.750 & 1.258 & 0.207 \\
Mean TR / mol    & 1.282 & 1.620 & 1.741 \\
Mean AA / mol & 2.7   & 5.4   & 1.0 \\
\bottomrule
\end{tabular}
\label{tab:suppl_arom}
\end{table}

\begin{figure*}[]
    \centering
    \begin{tabular}{cc}
         \includegraphics[width=0.4\linewidth]{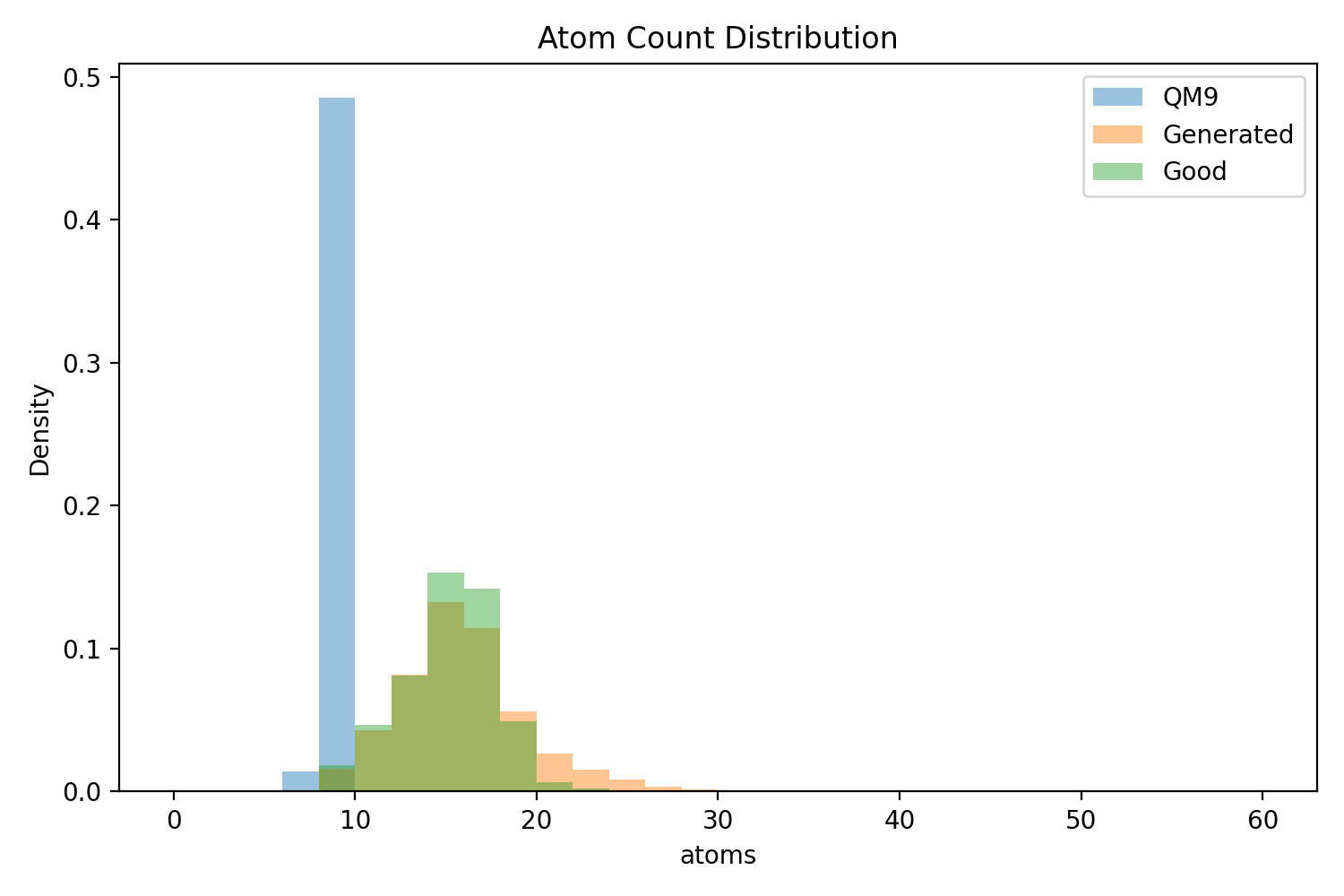} & \includegraphics[width=0.4\linewidth]{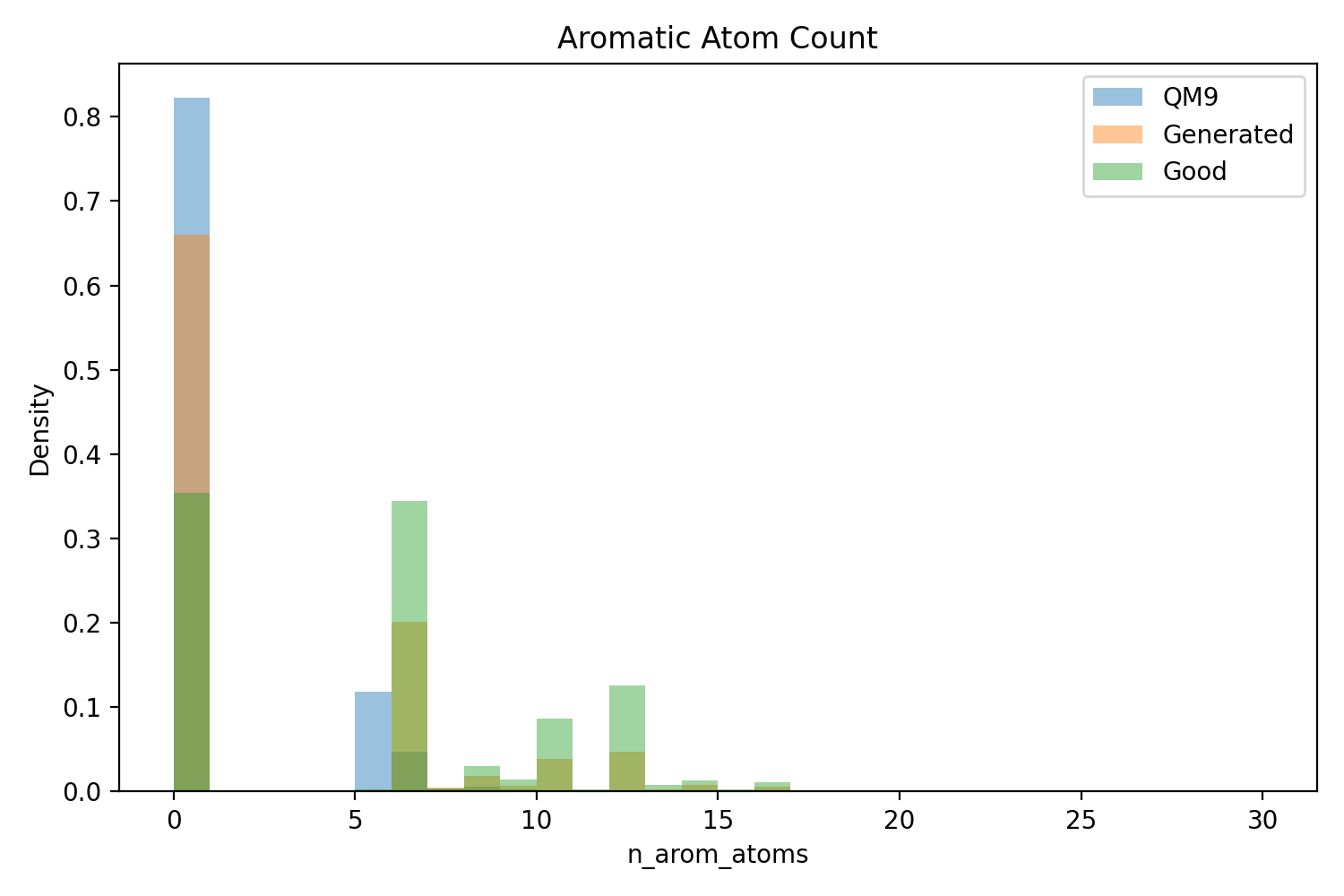} \\         
    \end{tabular}
    \caption{Distributions of (left) total atom count and (right) aromatic atom count for QM9, all generated molecules, and the \emph{Good@chem} subset. Generated molecules exhibit broader atom-count variation (10–25 atoms) relative to QM9 (centered near 9–10), while the \emph{Good@chem} subset increases aromatic atom frequency, reflecting enrichment of conjugated motifs.}
    \label{fig:suppl_histatoms}
\end{figure*}

\begin{figure*}[]
    \centering
    \begin{tabular}{cc}
         \includegraphics[width=0.4\linewidth]{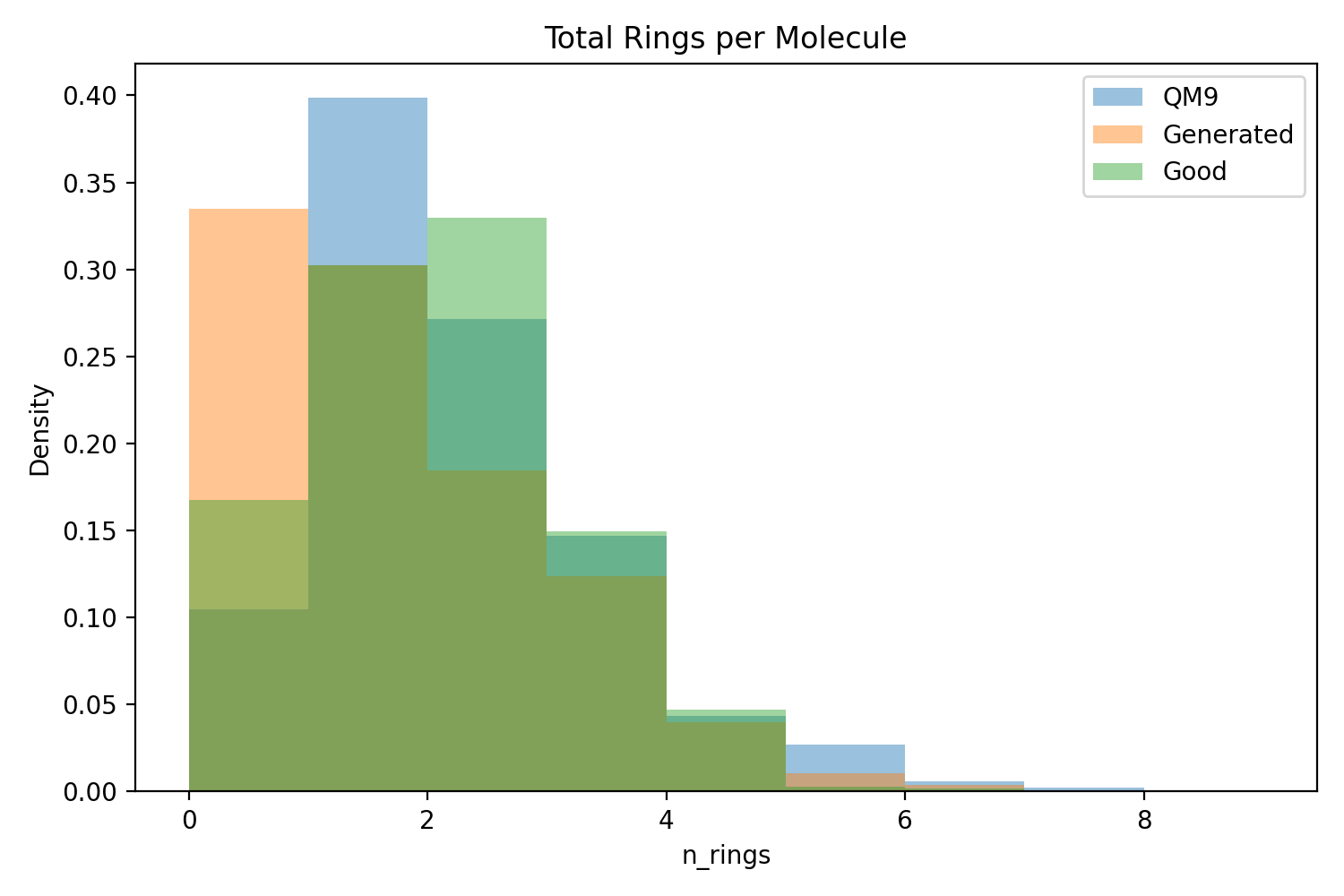} & \includegraphics[width=0.4\linewidth]{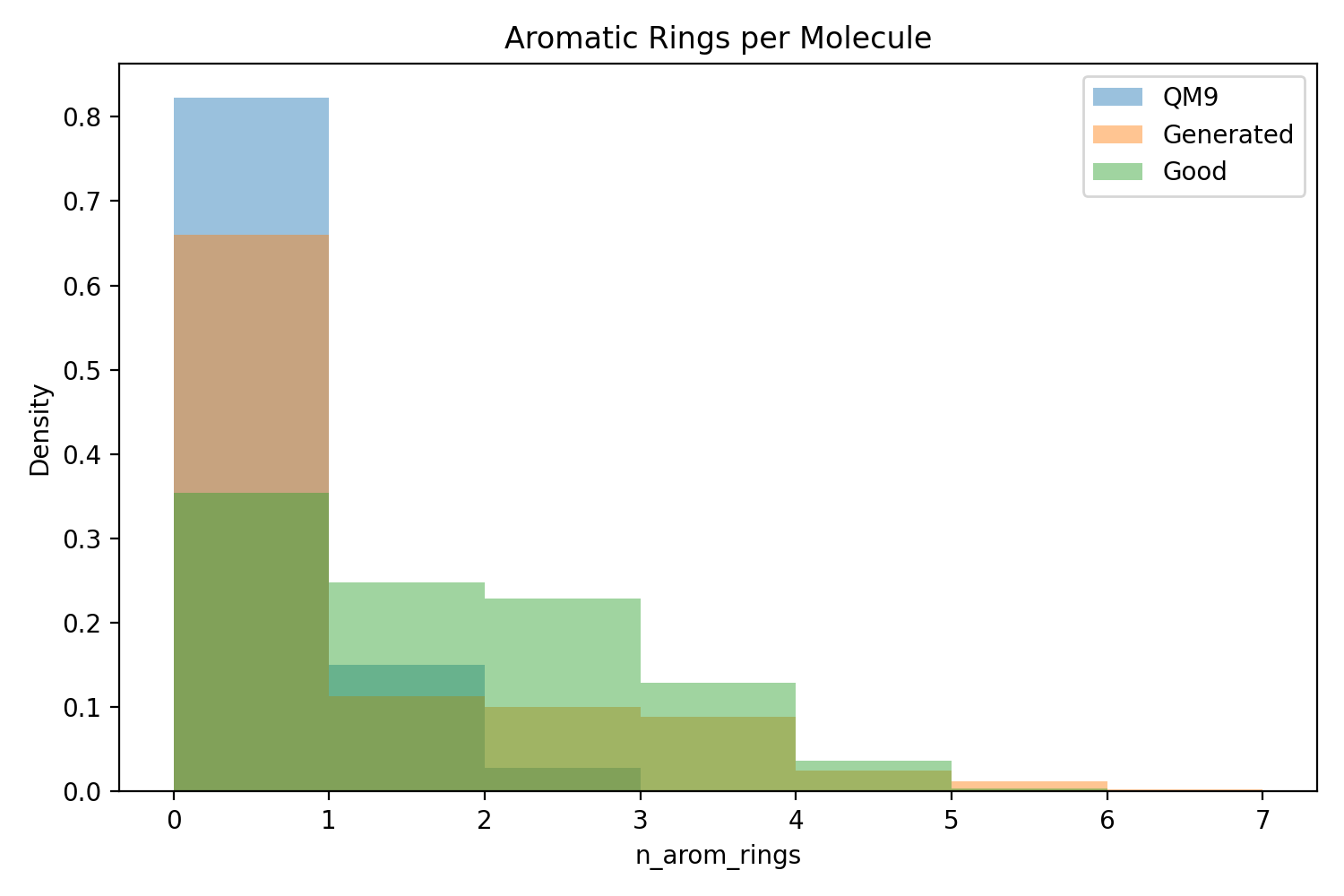} \\         
    \end{tabular}
    \caption{Distributions of (left) total rings and (right) aromatic rings per molecule. While total ring counts remain similar to QM9, the \emph{Good@chem} subset shows a higher fraction of aromatic rings (peaking at 1–2 rings per molecule), indicating that the aggregator captures ring-forming tendencies typical of drug-like compounds.}
    \label{fig:suppl_histrings}
\end{figure*}

\begin{figure*}[]
    \centering
    \begin{tabular}{ccc}
         \includegraphics[width=0.3\linewidth]{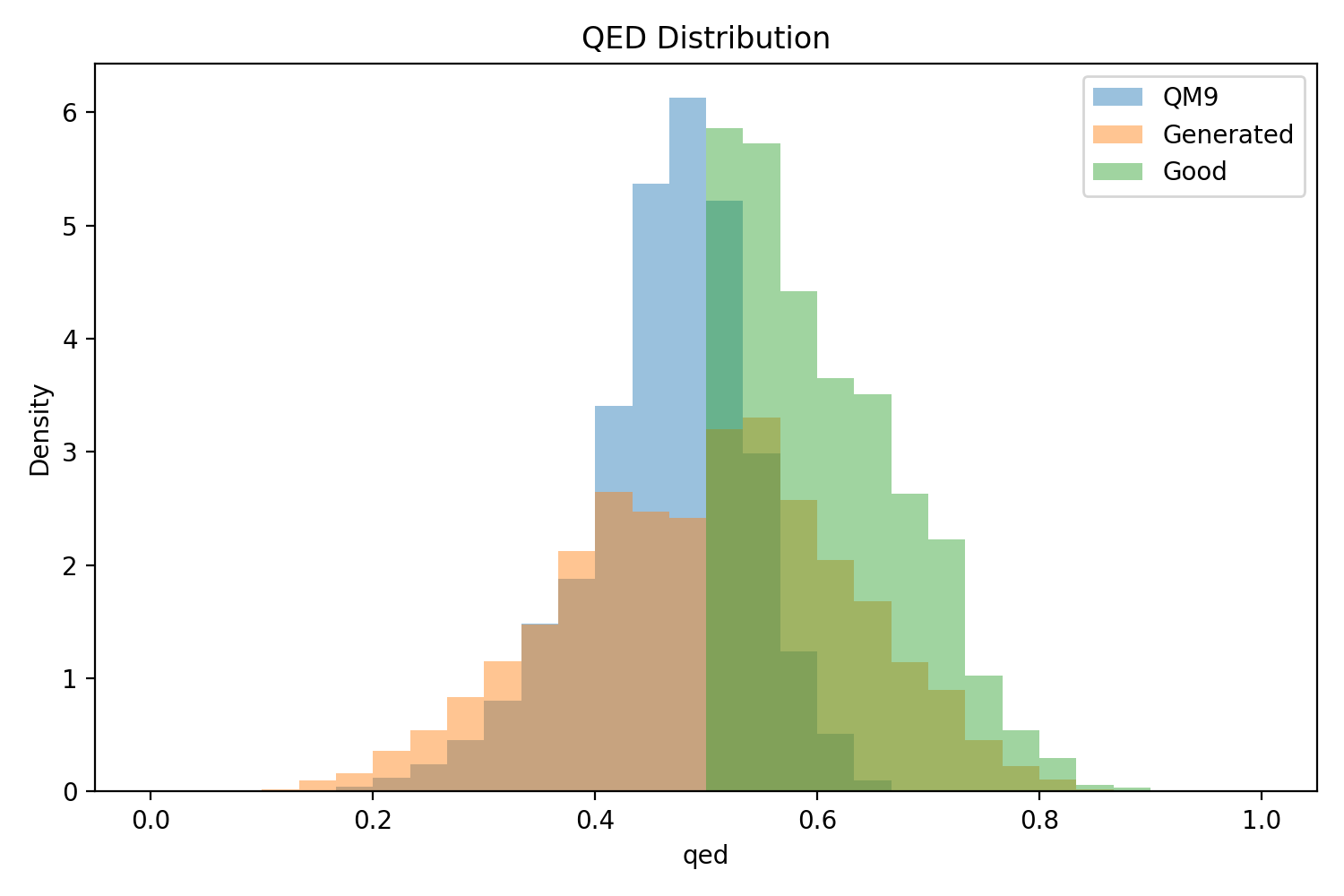} & \includegraphics[width=0.3\linewidth]{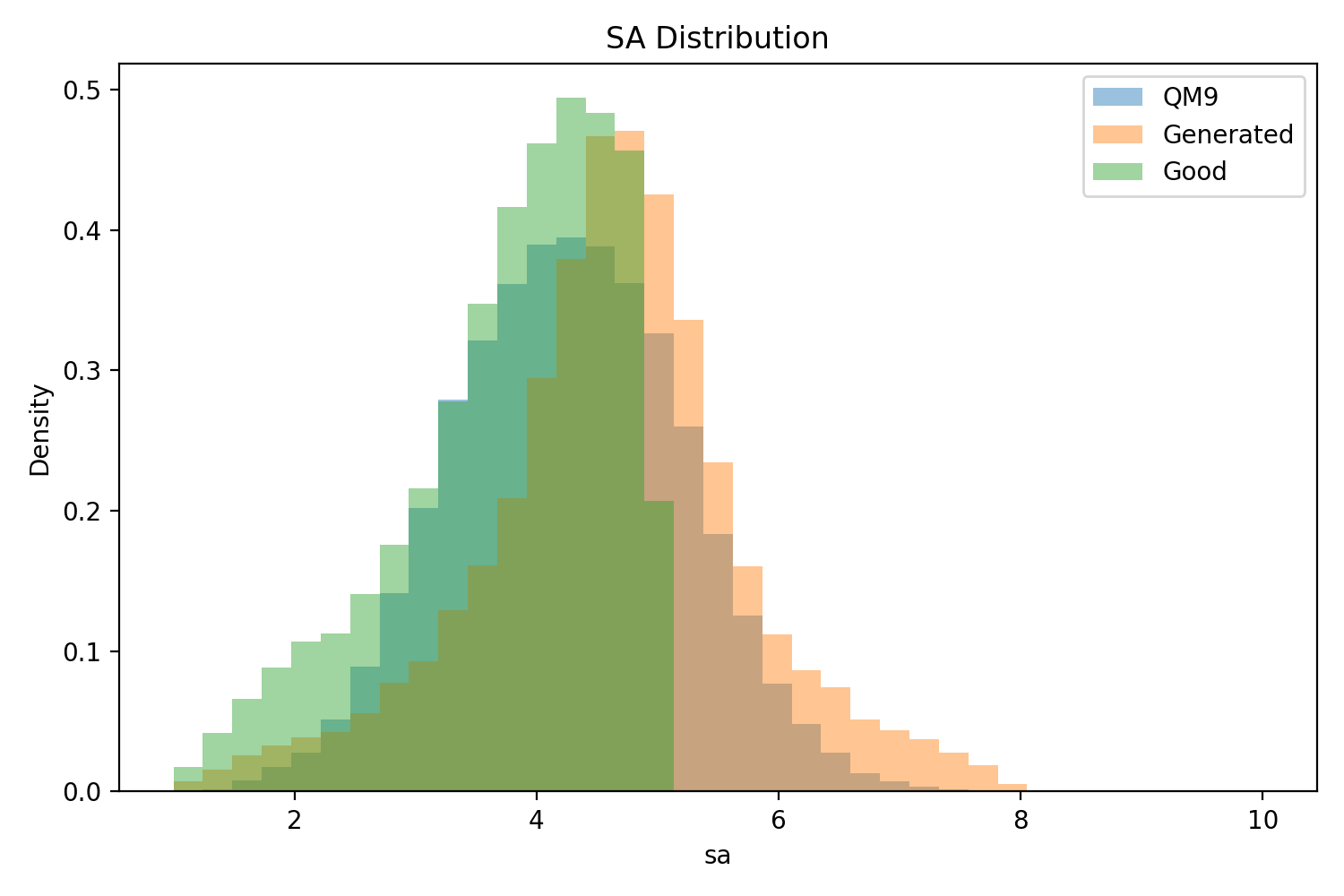} &
         \includegraphics[width=0.3\linewidth]{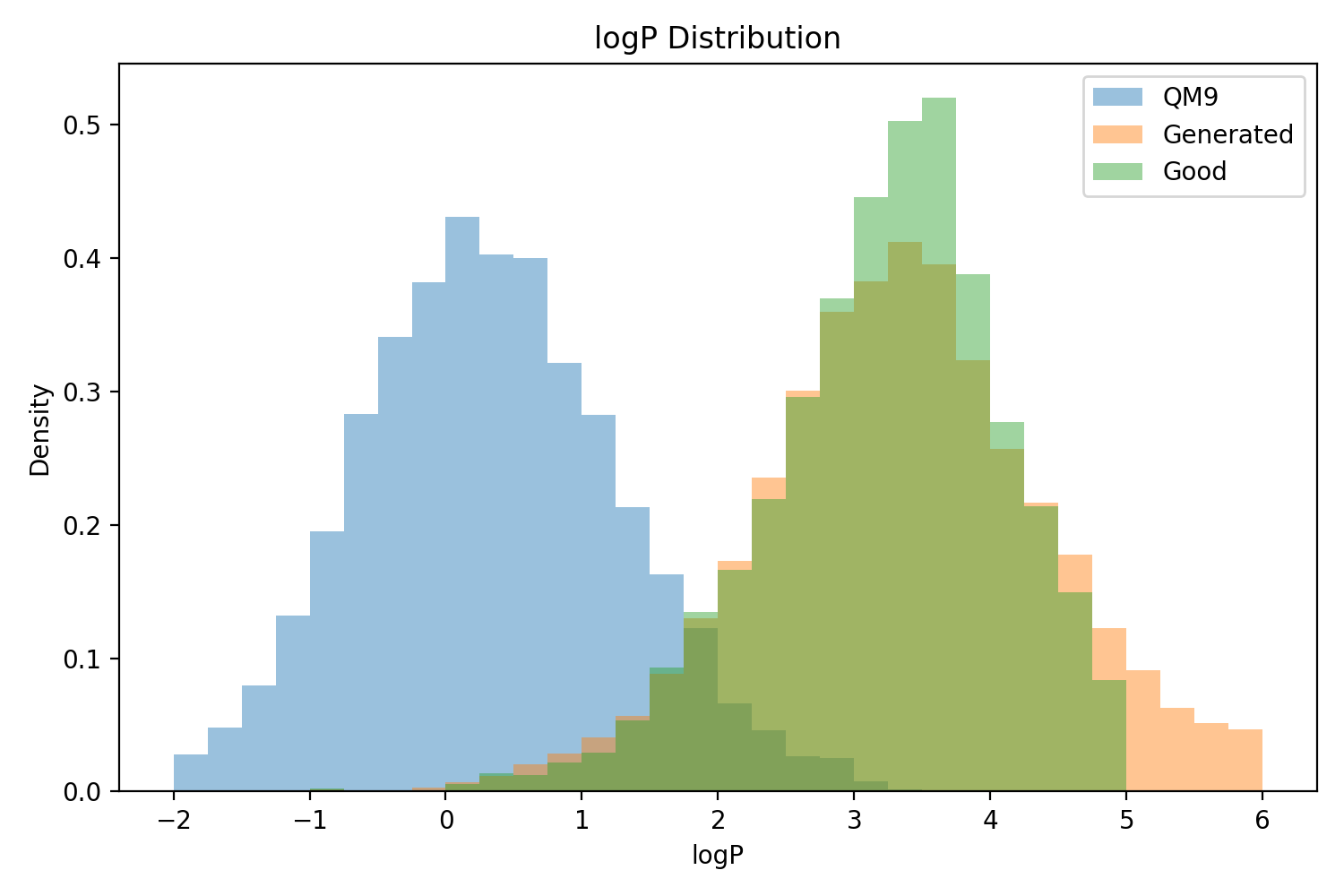} \\
    \end{tabular}
    \caption{Property alignment between QM9, generated, and \emph{Good@chem} molecules for (left) QED, (center) SA, and (right) $\log P$. The \emph{Good@chem} subset shifts toward higher QED and moderate $\log P$, with lower SA scores than the full generated distribution, demonstrating that \textsc{MolPaQ}’s conditioning and aggregation modules effectively steer generation toward chemically desirable regions of property space.}
    \label{fig:suppl_histprops}
\end{figure*}

\noindent
\textit{Aromaticity and Property Alignment:} To assess structural realism and property alignment, we compare distributions of atomic and ring-level statistics between QM9, all generated molecules, and the \emph{Good@chem} subset.
Figure~\ref{fig:suppl_histatoms}–\ref{fig:suppl_histprops} show that the \emph{Good@chem} molecules concentrate around higher QED and moderate $\log P$ values, while maintaining synthetic accessibility (SA) and atom-count distributions consistent with real molecules.
The aromatic ring and atom histograms, Figure~\ref{fig:suppl_histrings}, further confirm that the quantum–classical aggregator generates chemically plausible ring systems and enriches aromatic motifs without collapsing diversity.

\section{Descriptor–Only Inference and Property Steering Validation}
\label{sect:suppl_descriptor_steer}

\paragraph{Objective.}
To verify that \textsc{MolPaQ} can operate entirely from descriptor inputs $-$ without relying on encoder latents $-$ we evaluated a \emph{descriptor-only} inference mode using the trained Reduced Conditioner~(M2).  
The experiment tests whether the system can steer molecular generation toward desired properties by specifying only target descriptors $\mathbf{x}^*$ consisting of RDKit-derived features (\texttt{qed}, \texttt{MolLogP}), while the generator~(M3) and aggregator~(M4) remain fixed.

\paragraph{Setup.}
Descriptor vectors were drawn from the interquartile range of the QM9 descriptor distribution.  
We varied two steering dimensions, \texttt{qed} and \texttt{MolLogP}, over a $3{\times}3$ quantile grid (25\,\%, 50\,\%, 75\,\%).  
For each target $(q_{\text{QED}}, q_{\text{logP}})$, we sampled 512 molecules without encoder latents and measured achieved QED, logP, and SA.  
To reduce stochastic variance, each target was jittered by Gaussian noise ($\sigma{=}0.05$\,IQR per descriptor, 16 replicates), and the reported statistics summarize the pooled set of valid molecules ($n\!\approx\!200$ per cell on average).  
A lightweight monotone calibration of logP was applied using a 5-point quantile map learned from calibration targets (\texttt{run\_meta.json}).

\paragraph{Results.}
Table~\ref{tab:suppl_descriptor_steer} reports the median$\pm$IQR of achieved properties for diagonal cells (25–25, 50–50, 75–75).  
Across the full grid (nine targets), QED tracks its targets tightly (median MAE~$\approx$~0.07) whereas logP exhibits a narrow achievable range (median~$\approx$~3.0–3.4) with weak monotonicity (Spearman~$\rho\!<\!0.1$).  
The conditioner thus provides consistent QED steering but only coarse control of lipophilicity.  
SA is not steered as an input and serves as an outcome indicator of synthetic accessibility.

\begin{table*}[t]
\centering
\caption{Descriptor-only steering on interquartile QED$\times$logP targets (median~$\pm$~IQR over generated valid molecules, $n\!\approx\!200$ per cell).}
\label{tab:suppl_descriptor_steer}
\begin{tabular}{lccc}
\toprule
Target quantiles (QED, logP) & QED & logP & SA \\
\midrule
(25\%, 25\%) & 0.53 $\pm$ 0.10 & 2.99 $\pm$ 0.83 & 4.49 $\pm$ 1.09 \\
(50\%, 50\%) & 0.53 $\pm$ 0.11 & 3.23 $\pm$ 0.93 & 4.54 $\pm$ 1.13 \\
(75\%, 75\%) & 0.56 $\pm$ 0.09 & 2.98 $\pm$ 0.87 & 4.62 $\pm$ 1.17 \\
\bottomrule
\end{tabular}
\end{table*}

\begin{figure}[t]
    \centering
    \includegraphics[width=0.9\linewidth]{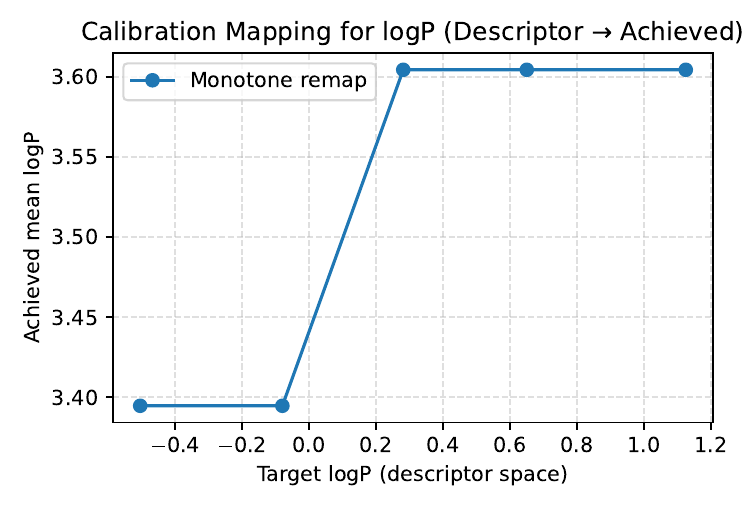}
    \vspace{-1ex}
    \caption{LogP calibration curve (achieved~vs.~target).  
    The near-flat response highlights the model’s bias toward moderate lipophilicity.}
    \label{fig:suppl_calibration_logp}
\end{figure}

\paragraph{Calibration behavior.} Figure~\ref{fig:suppl_calibration_logp} plots the calibration mapping. The achieved logP values plateau between $\approx$3.4–3.6 across target quantiles, explaining the flat response observed in the main grid. LogP steering behaves more coarsely in descriptor-only mode due to the narrow logP range of QM9 and the fact that the Reduced Conditioner (M2) is trained primarily on QED-correlated axes. Extending M2 with additional curated descriptors (e.g., extended logP/SA features) is a straightforward modification and left for future extensions.

\paragraph{Summary.}
Descriptor-only inference validates that \textsc{MolPaQ} can generate chemically valid molecules directly from numeric property inputs.  
Steering precision currently extends to QED, while logP and SA require expanded training descriptors and conditioner calibration.  
These findings reinforce the framework’s modular interpretability and guide future improvements in property-to-latent conditioning.

\section{Latent-Property Interpretability Audit}
\label{sec:suppl_latent_audit}

\begin{figure}[h]
\centering
\includegraphics[width=1.0\linewidth]{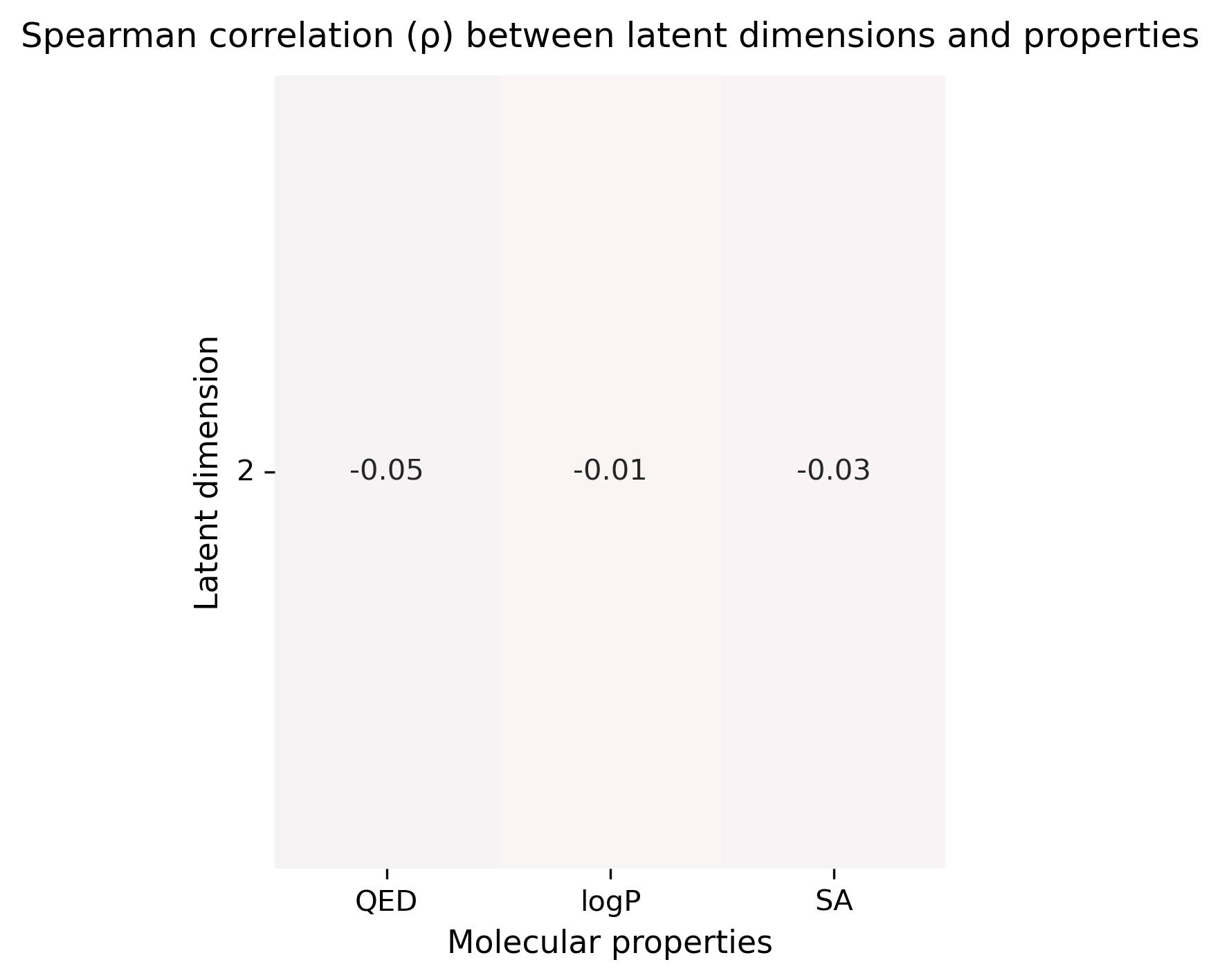}
\caption{Spearman $\rho$ between the traversed latent dimension and molecular properties. Values near zero indicate no monotonic association.}
\label{fig:suppl_latent_corr}
\end{figure}

\begin{figure*}[h]
\centering
\includegraphics[width=0.32\linewidth]{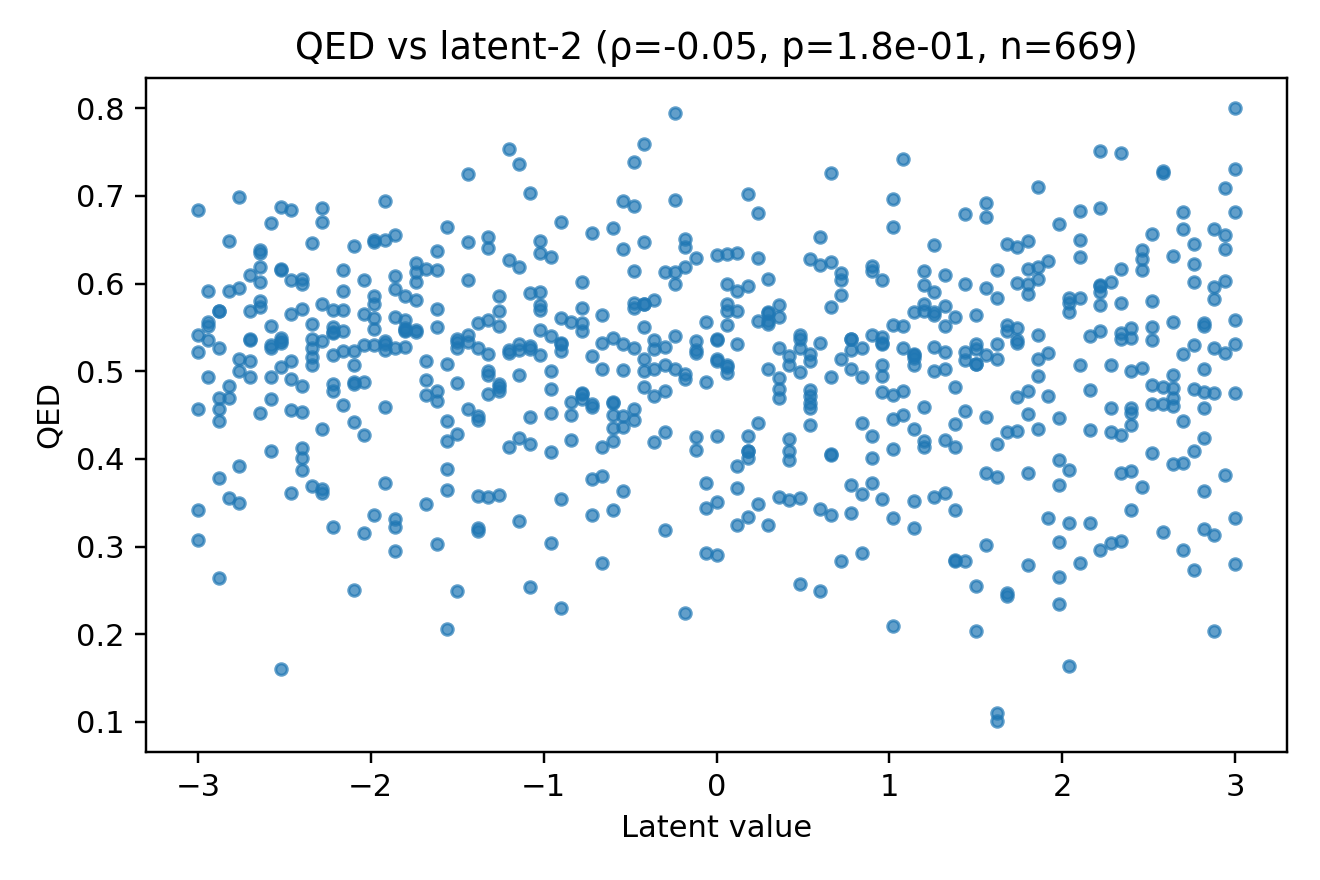}
\includegraphics[width=0.32\linewidth]{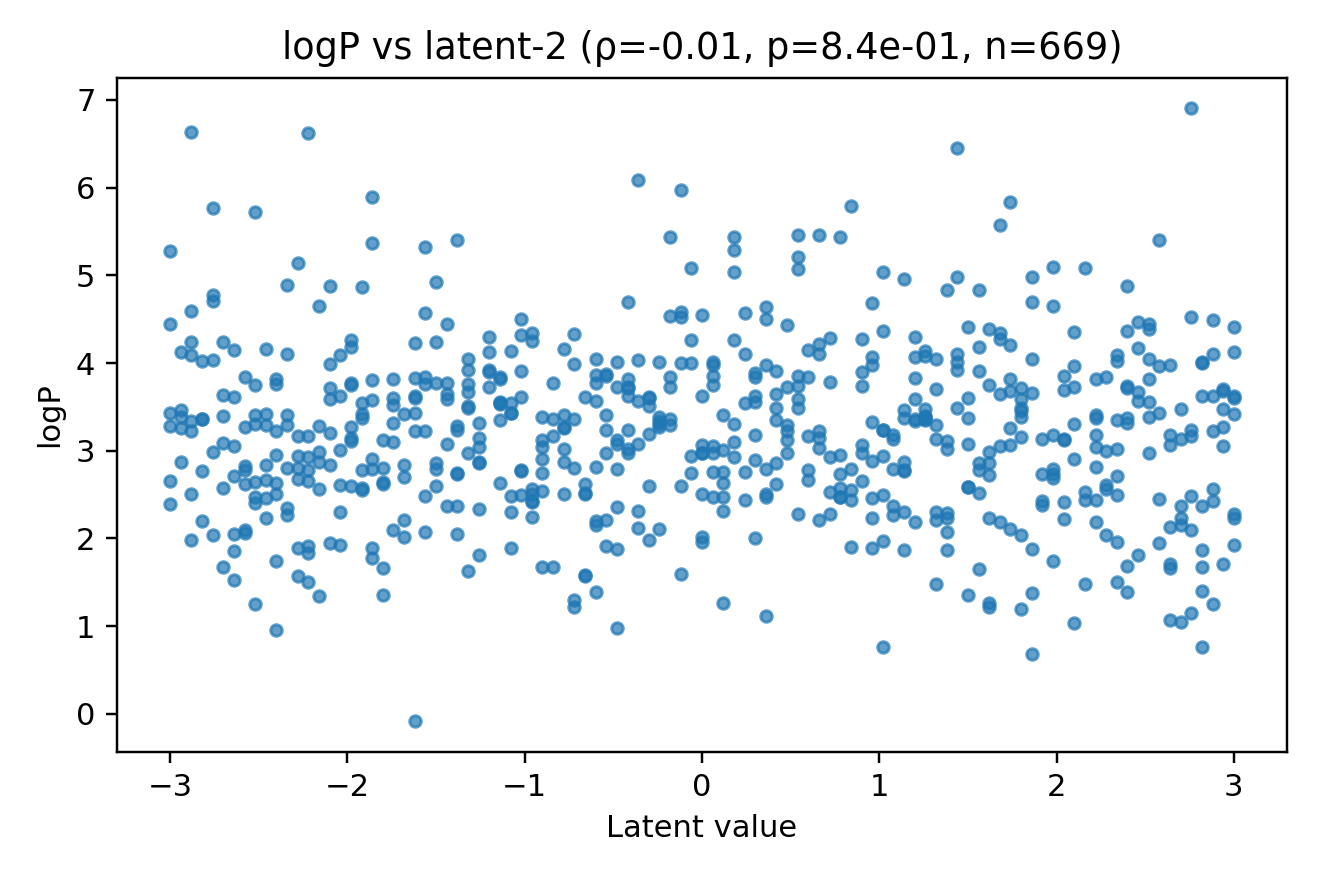}
\includegraphics[width=0.32\linewidth]{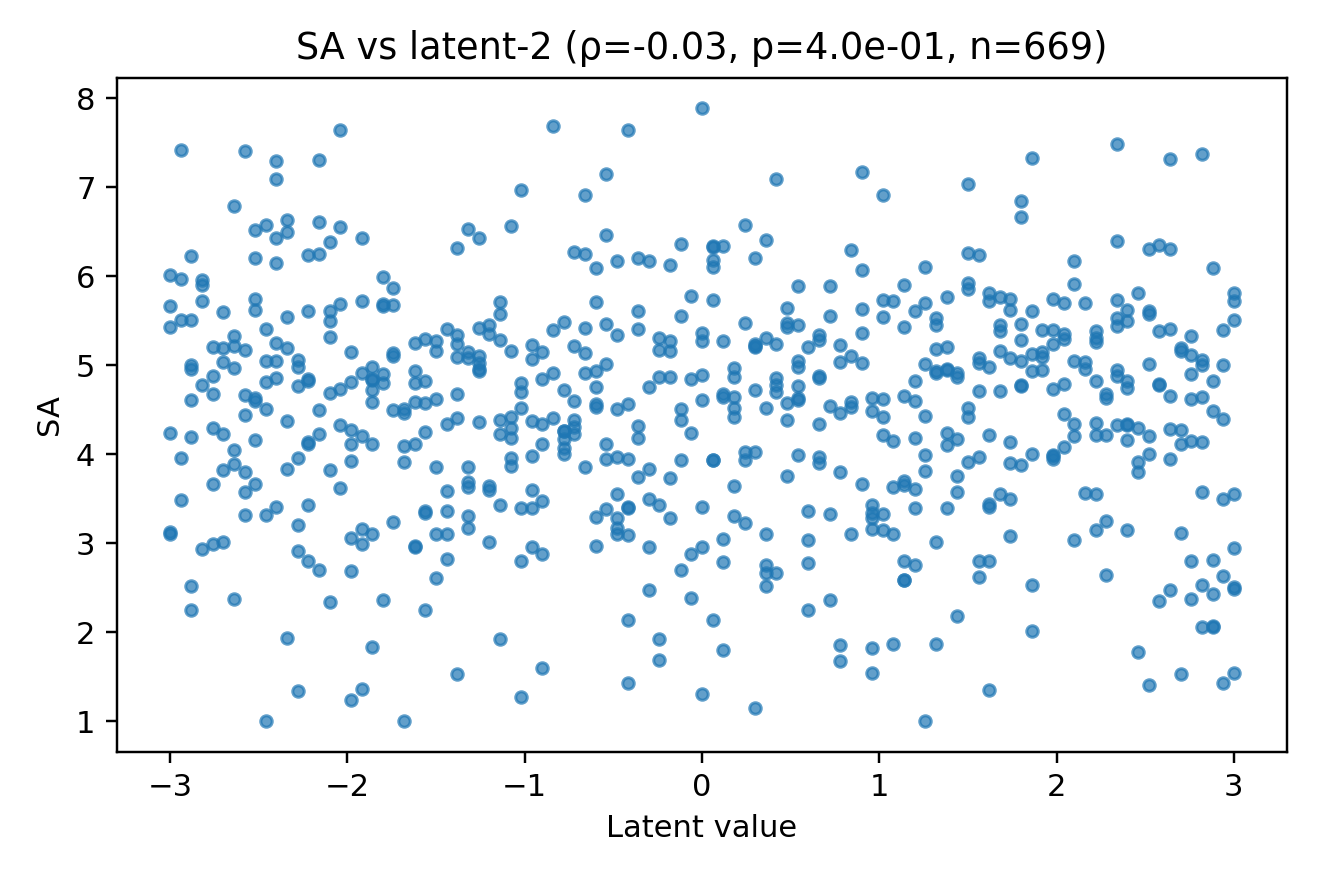}
\caption{Property values vs latent value for dim~2 with least-squares guide line. No visible trend is observed.}
\label{fig:suppl_latent_scatter}
\end{figure*}

\paragraph{Overview.}
To examine whether individual latent coordinates encode chemically meaningful trends, we performed a latent–property audit on the traversed dimension (dim~2). As shown in Table~\ref{tab:suppl_latent_audit_tbl} and Fig.~\ref{fig:suppl_latent_corr}, Spearman correlations with QED, logP, and SA are extremely small ($|\rho|\le 0.05$ with $p>0.18$), indicating no monotonic association. The scatter plots in Fig.~\ref{fig:suppl_latent_scatter} confirm this behavior: property values form uniform, noise-like clouds across the entire latent range, with no directional gradient or visible trend. The heatmap likewise shows values clustered tightly around zero.

This pattern is expected for \textsc{MolPaQ}. Because molecular structure is synthesized through the interaction of the conditioner, quantum patch generator, and aggregator, property information is distributed across multiple latent dimensions rather than aligned with a single axis. The quantum generator in particular mixes latent amplitudes nonlocally, producing entangled representations where properties emerge from joint multi-dimensional interactions. Consequently, single-dimension traversals do not capture global structure–property relationships, even though the model exhibits smooth property shifts under coordinated manipulation of multiple latent variables (main paper, Sec.~4.12).

Overall, the audit confirms that \textsc{MolPaQ} employs a multi-factorial latent geometry rather than relying on axis-aligned directions, consistent with its modular quantum–classical design and its ability to generate coherent, property-compliant structures.

\begin{table}[h]
\centering
\caption{Latent–property correlations for the traversed dimension (dim 2). Spearman $\rho$ and two-sided $p$-values; $n{=}669$.}
\begin{tabular}{lccc}
\toprule
Property & $\rho$ & $p$-value & $n$ \\
\midrule
QED  & $-0.052$ & $0.18$ & 669 \\
SA   & $-0.032$ & $0.40$ & 669 \\
logP & $-0.008$ & $0.84$ & 669 \\
\bottomrule
\end{tabular}
\label{tab:suppl_latent_audit_tbl}
\end{table}

\section{Top-36 Generated Molecules}
\begin{figure*}
    \centering
    \includegraphics[width=\linewidth]{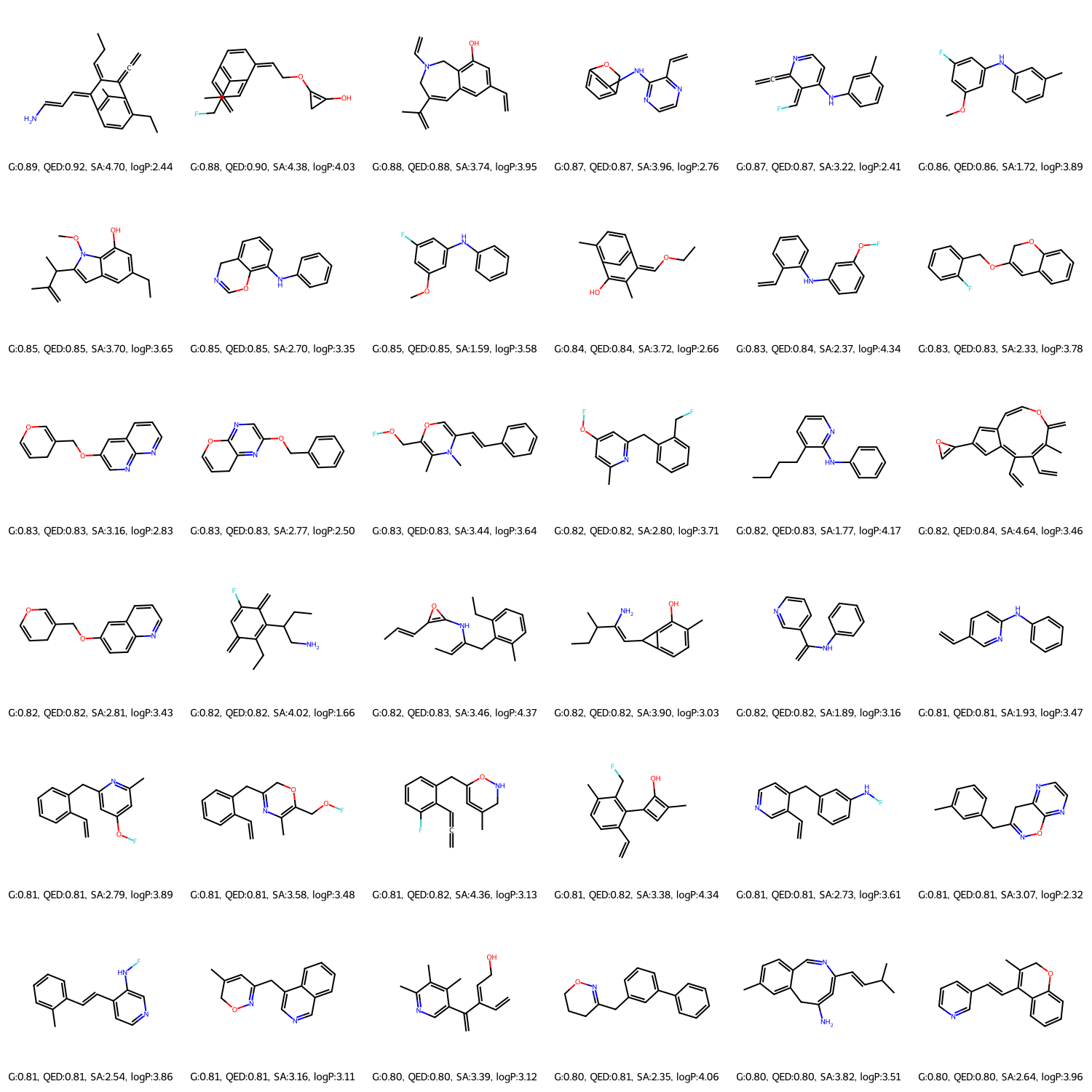}
    \caption{Expanded view of the top-36 generated molecule for improved readability}
    \label{fig:suppl_molgrid}
\end{figure*}

Figure ~\ref{fig:suppl_molgrid} presents expanded view of the the top-36 generated molecules for improved visibility and readability.

\section{Extended Pareto analysis (dim~2)}

\paragraph{Overview.}
The full Pareto analysis includes both unconstrained and constrained fronts derived from the latent traversal along dimension~2.
Each point represents a unique molecule decoded from the generator, annotated with its computed QED, logP, and SA values.
The constrained front enforces SA~$\leq6$ and logP~$\in[-0.5,5.0]$.

\begin{figure}[h]
  \centering
  \includegraphics[width=\linewidth]{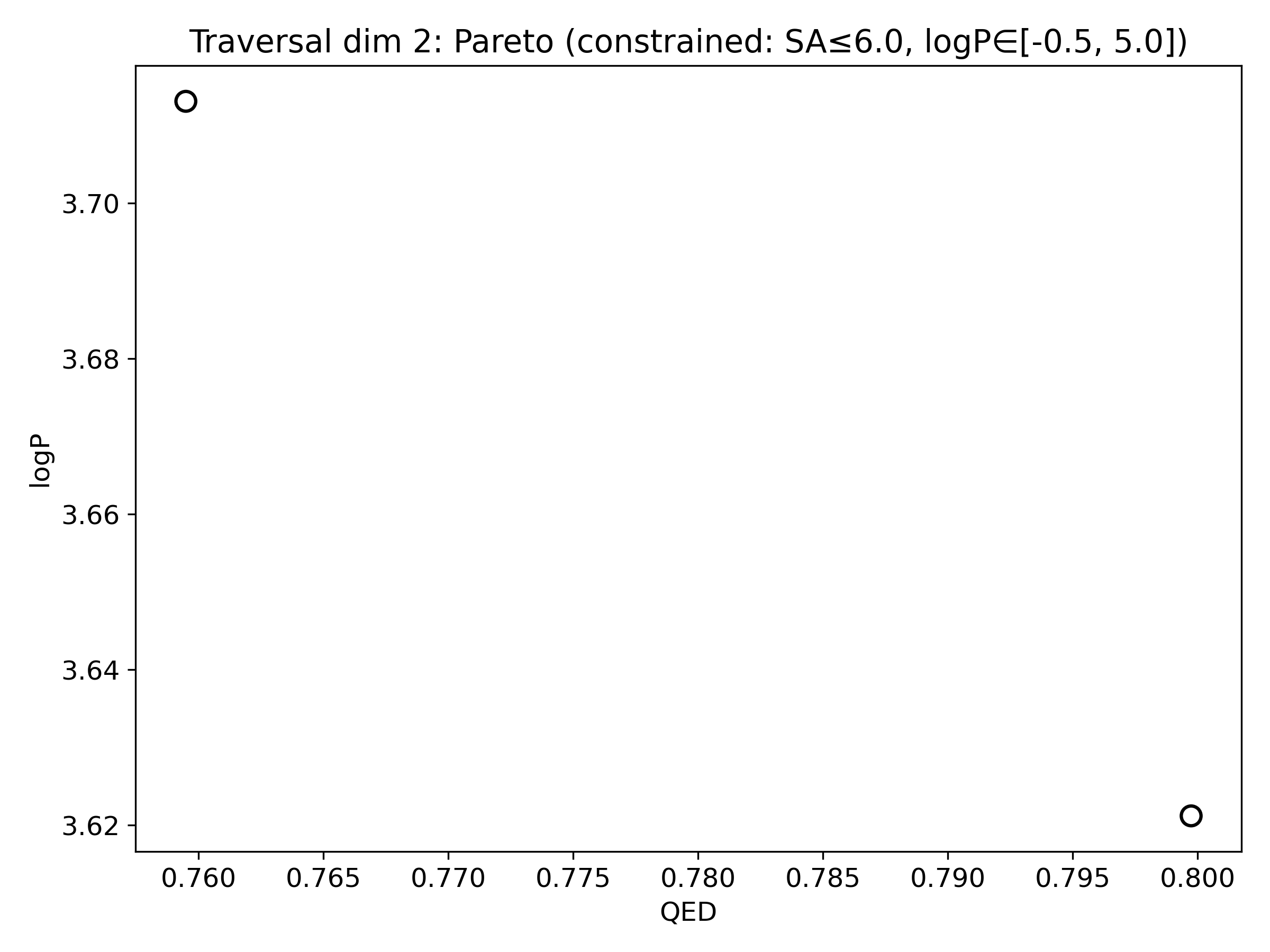}
  \caption{
  Front-only view highlighting the efficient set for traversal dimension~2.
  Only two molecules satisfy all chemistry constraints, corresponding to the red circled points in Fig.~\ref{fig:pareto_scatter_dim2}.
  }
  \label{fig:pareto_front_only_dim2}
\end{figure}


\paragraph{Summary statistics.}
Among 669 valid decoded molecules, two (0.30\%) met all constraints and lie on the Pareto frontier, Table~\ref{tab:pareto_full_dim2} and Fig.~\ref{fig:pareto_front_only_dim2}.

\begin{table}[h]
\centering
\caption{
Full chemistry-constrained Pareto front for traversal dimension~2.
Both candidates show desirable QED and logP within bounds, and moderate SA, confirming chemically realistic outputs.
}
\footnotesize
\begin{tabular}{r l c c c}
\toprule
\# & SMILES & QED & logP & SA \\
\midrule
1 & \texttt{CC(C)(C)OC(=O)Nc1ccc(OC)cc1} & 0.800 & 3.62 & 5.01 \\
2 & \texttt{COc1ccc(NC(=O)OC(C)(C)C)cc1} & 0.760 & 3.72 & 5.08 \\
\bottomrule
\end{tabular}
\label{tab:pareto_full_dim2}
\end{table}

\section{Extended results for the mode-collapse stress test}
\begin{figure*}[t]
\centering
\begin{subfigure}[t]{0.48\linewidth}
\includegraphics[width=\linewidth]{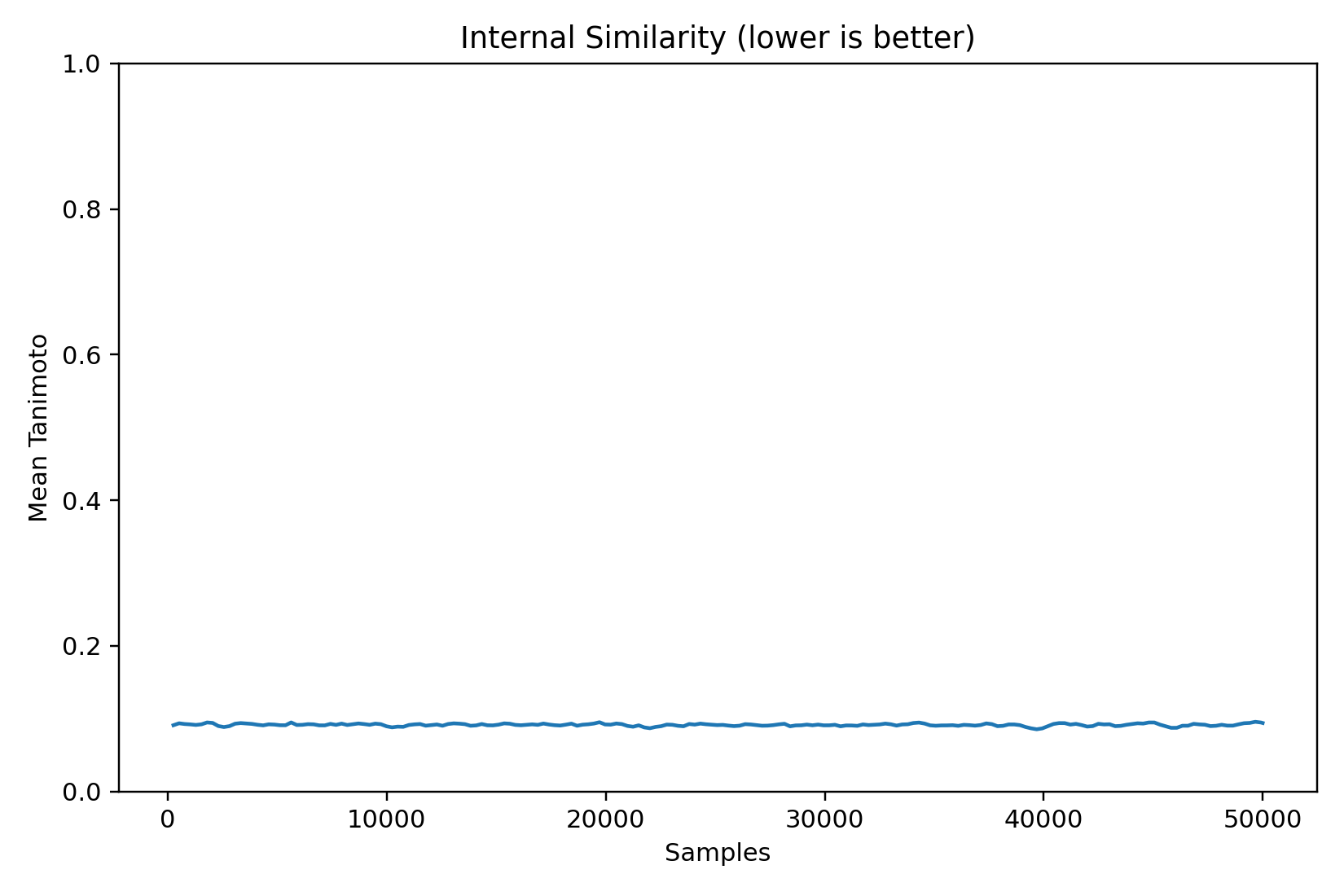}
\caption{Rolling mean Tanimoto (W=512).}
\end{subfigure}\hfill
\begin{subfigure}[t]{0.48\linewidth}
\includegraphics[width=\linewidth]{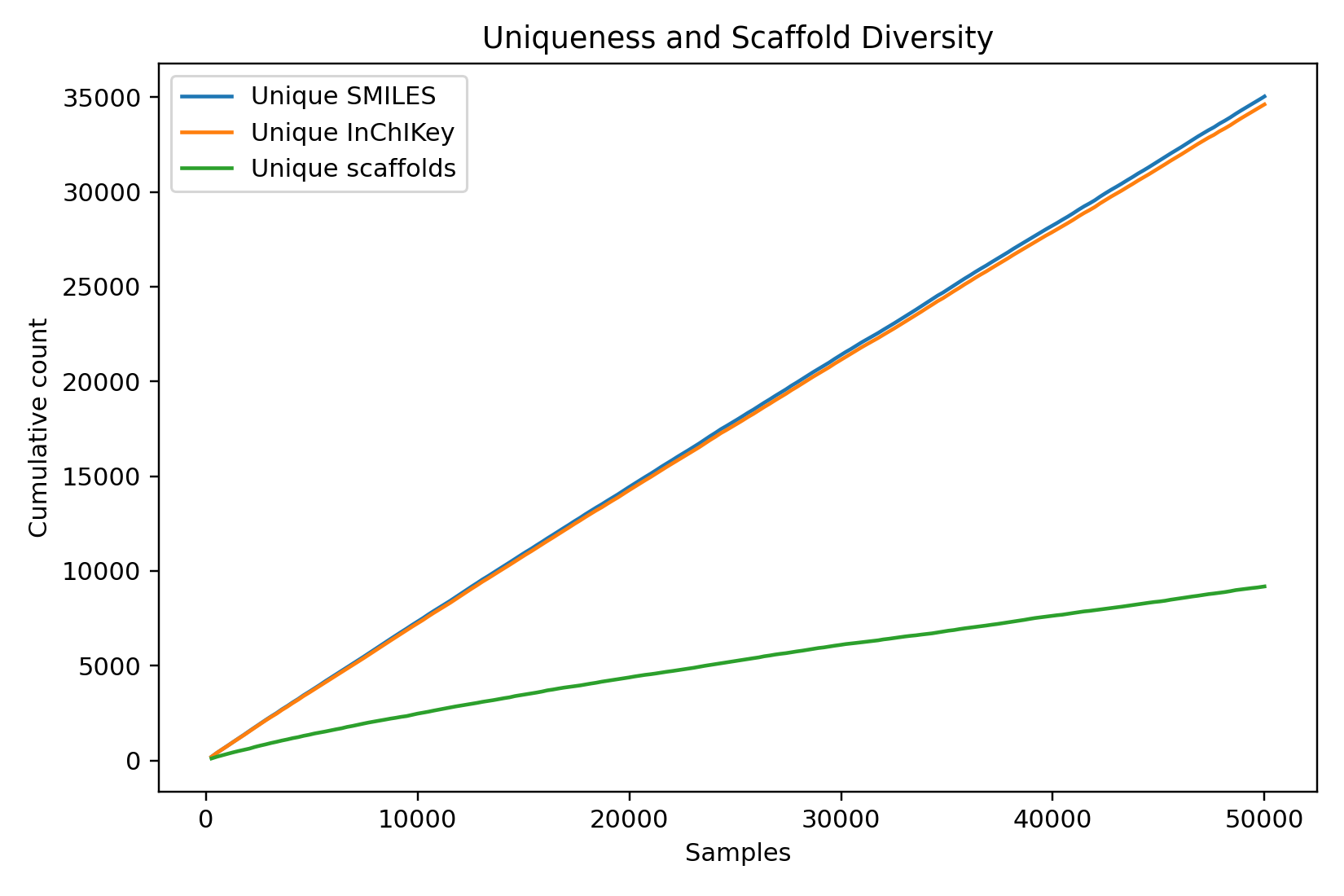}
\caption{Cumulative unique SMILES / scaffolds.}
\end{subfigure}
\caption{Mode-collapse stress test over 50k samples.}
\label{fig:suppl_modecollapse-panels}
\end{figure*}

\begin{figure}[h]
\centering
\includegraphics[width=0.9\linewidth]{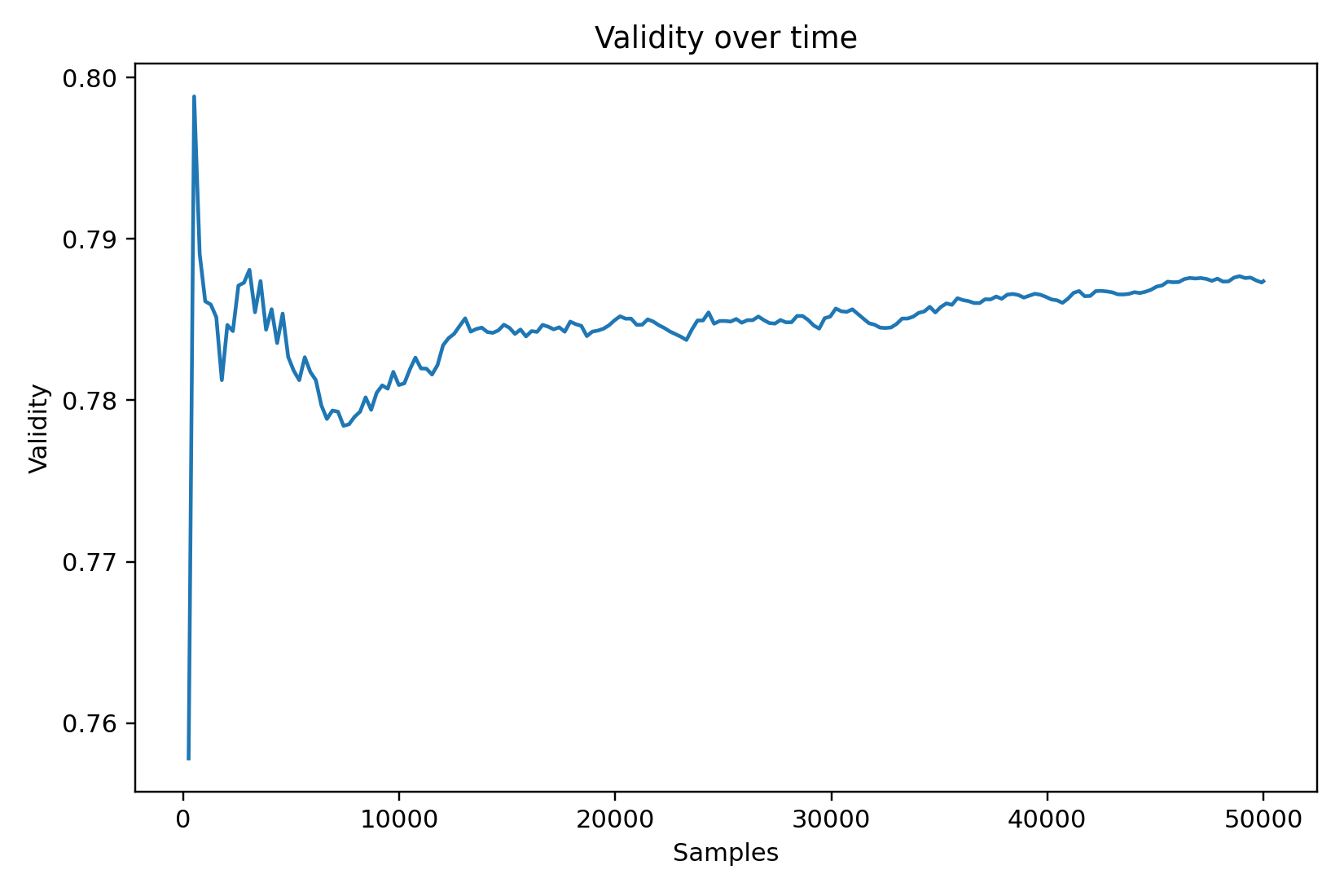}
\caption{validity vs.\ samples. Both diversity tracks grow steadily; validity stabilizes near 0.787.}
\label{fig:suppl_modecollapse-suppl-curves}
\end{figure}

We report full curves and a compact summary of end-of-run metrics for the 50k-sample stress test (Morgan fingerprints, 
$r =$ 2, 2048 bits; rolling window $W =$ 512). See Figures~\ref{fig:suppl_modecollapse-panels} and \ref{fig:suppl_modecollapse-suppl-curves} and Table~\ref{tab:suppl_modecollapse-summary} for details. 

Note that the 100\% validity reported for descriptor-guided sampling refers to conditioned generation. The 50k-sample stress test uses unconditional random latents to probe mode collapse $-$ a far harder setting $-$ and is not intended to match the conditioned validity rate. The difference reflects conditioning difficulty, not instability.

\begin{table*}
\centering
\small
\begin{tabular}{lcc}
\toprule
Metric & Value & Notes \\
\midrule
Samples generated & 50{,}000 & random latent draws \\
Validity (final) & 0.787 & RDKit sanitization pass fraction \\
Unique SMILES & 35{,}036 & cumulative at 50k \\
Unique scaffolds & 9{,}180 & Murcko scaffolds at 50k \\
Rolling mean Tanimoto & 0.094 & Morgan ($r = 2$, 2048), window $W = 512$ \\
\bottomrule
\end{tabular}
\caption{Summary metrics at the end of the 50k-sample mode-collapse stress test.}
\label{tab:suppl_modecollapse-summary}
\end{table*}

\section{Extended Physicochemical Property Analysis}

\begin{figure}[h]
  \centering
  \includegraphics[width=0.95\linewidth]{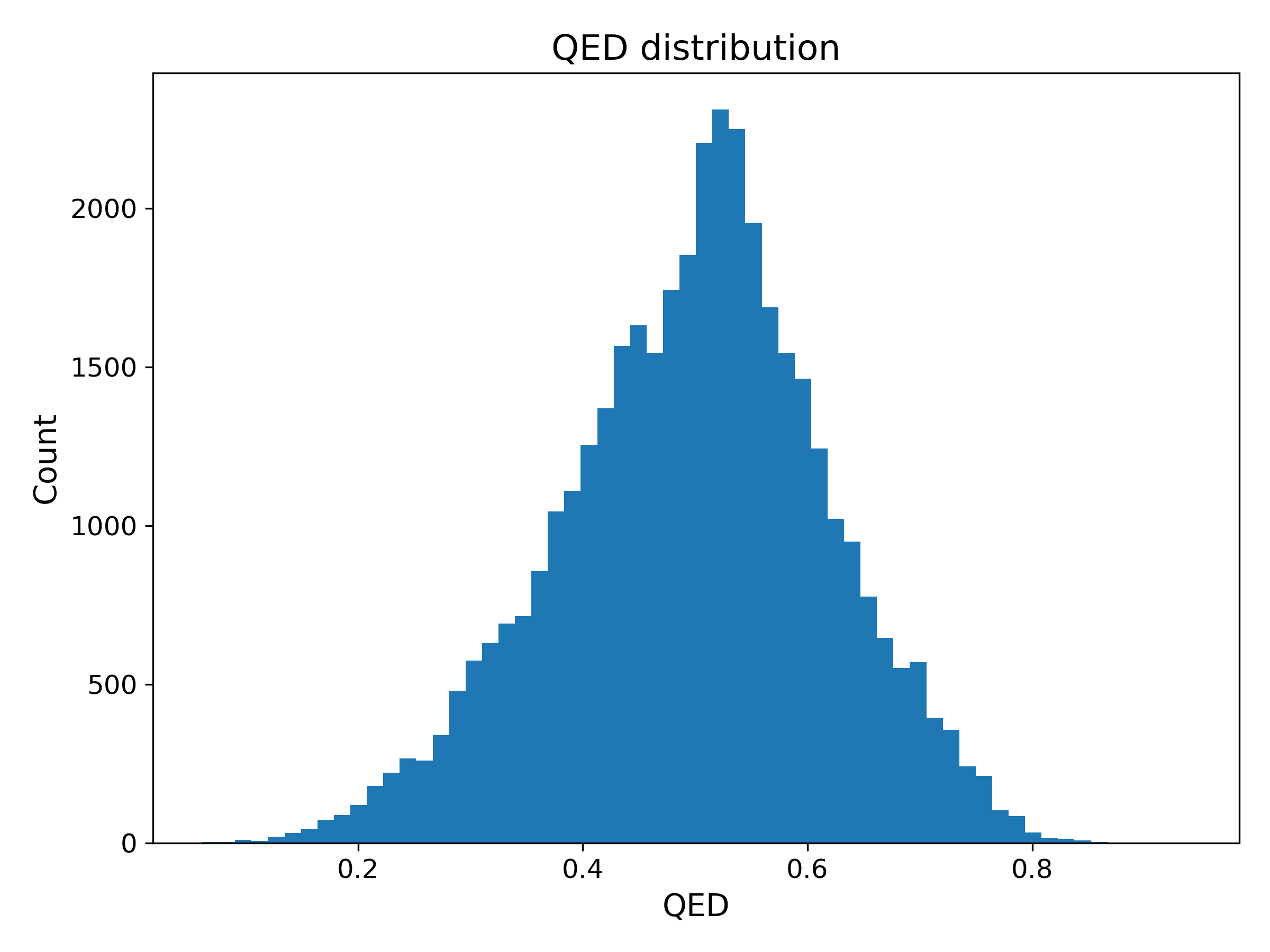}\\
  \includegraphics[width=0.95\linewidth]{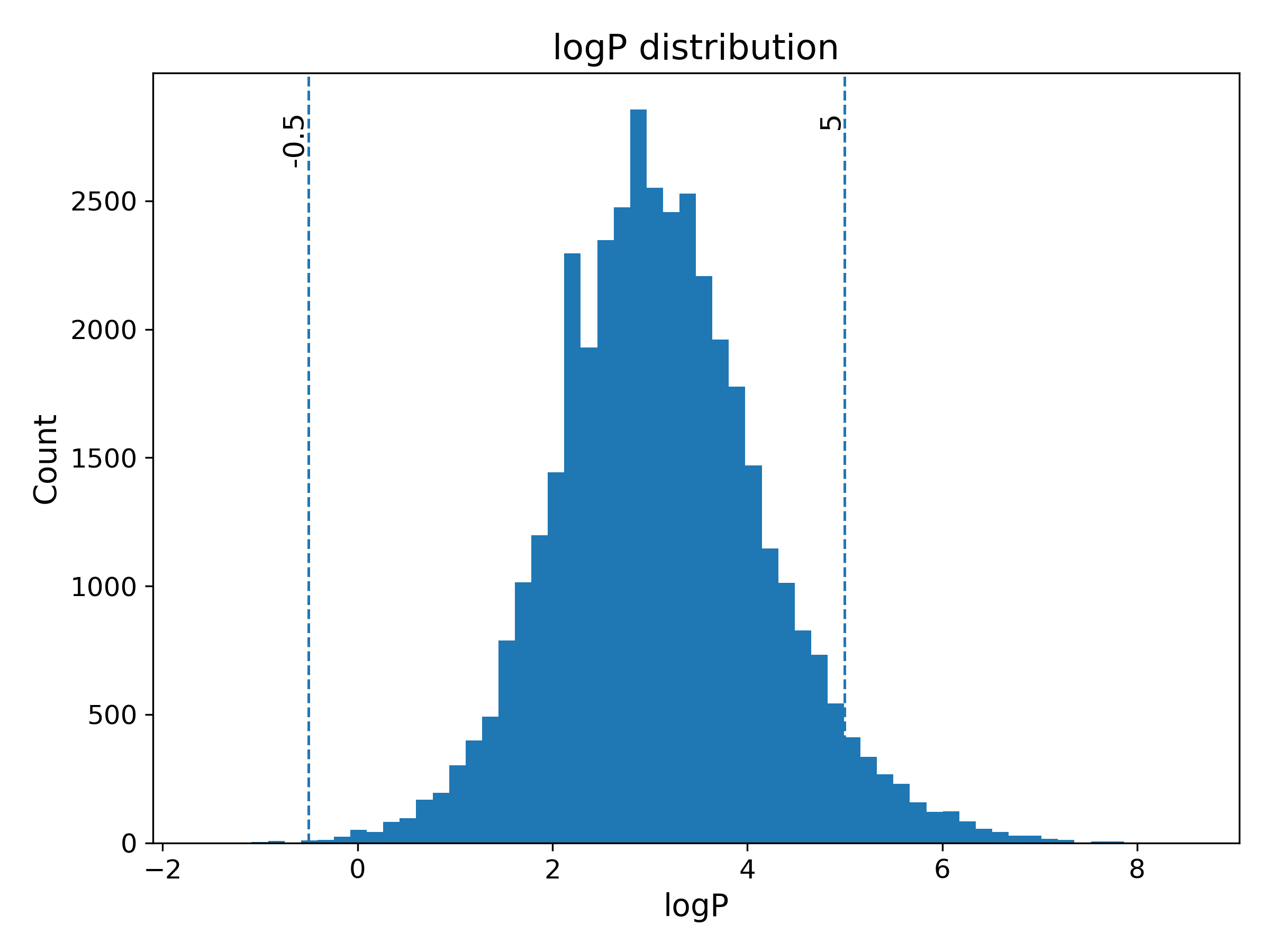}\\
  \caption{Histograms of QED and logP for 50k generated molecules. The logP histogram shows most samples within the recommended range of $[-0.5, 5.0]$.}
  \label{fig:hist-props}
\end{figure}

\begin{figure}[h]
  \centering
  \includegraphics[width=0.95\linewidth]{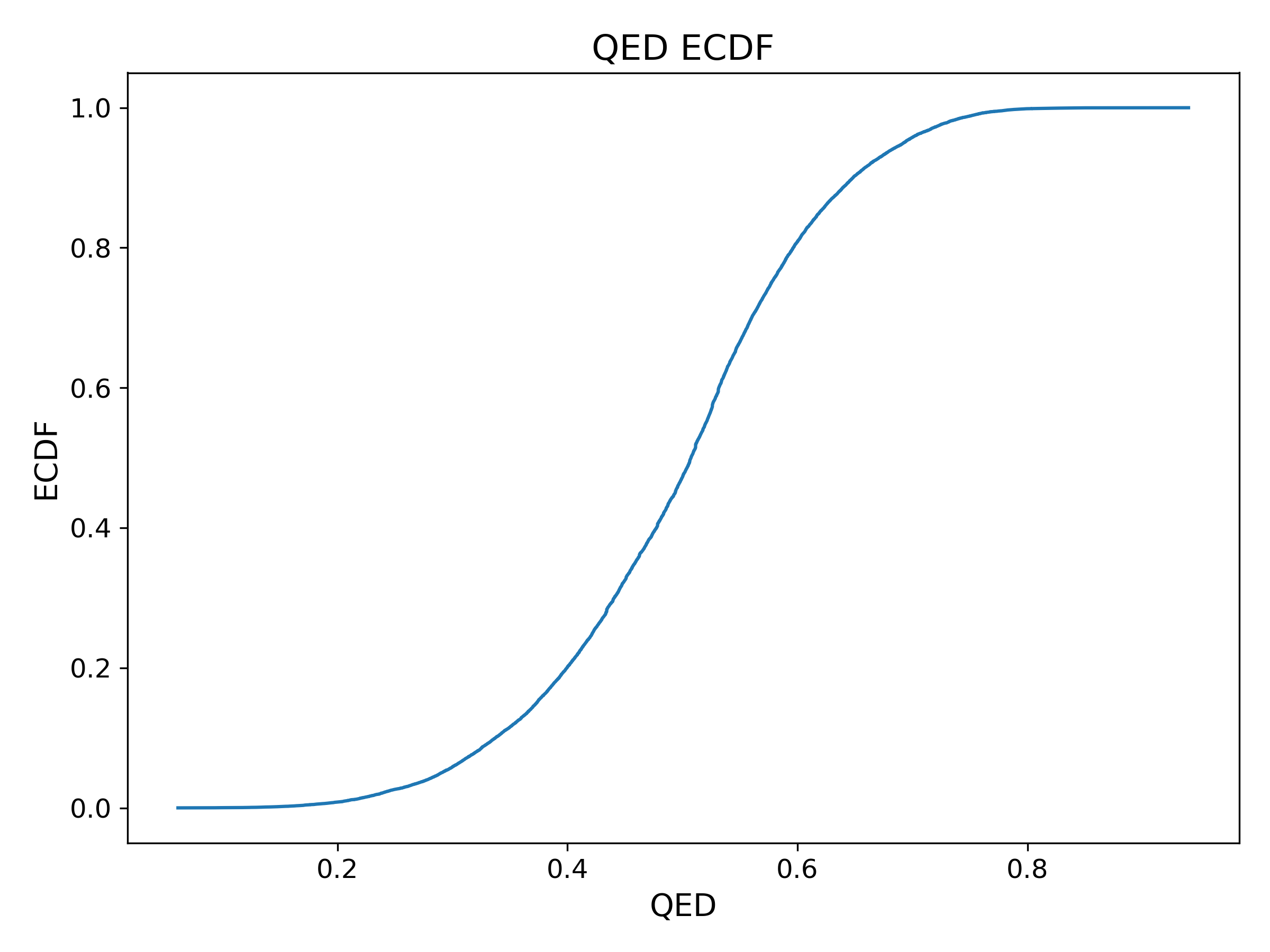}\\
  \includegraphics[width=0.95\linewidth]{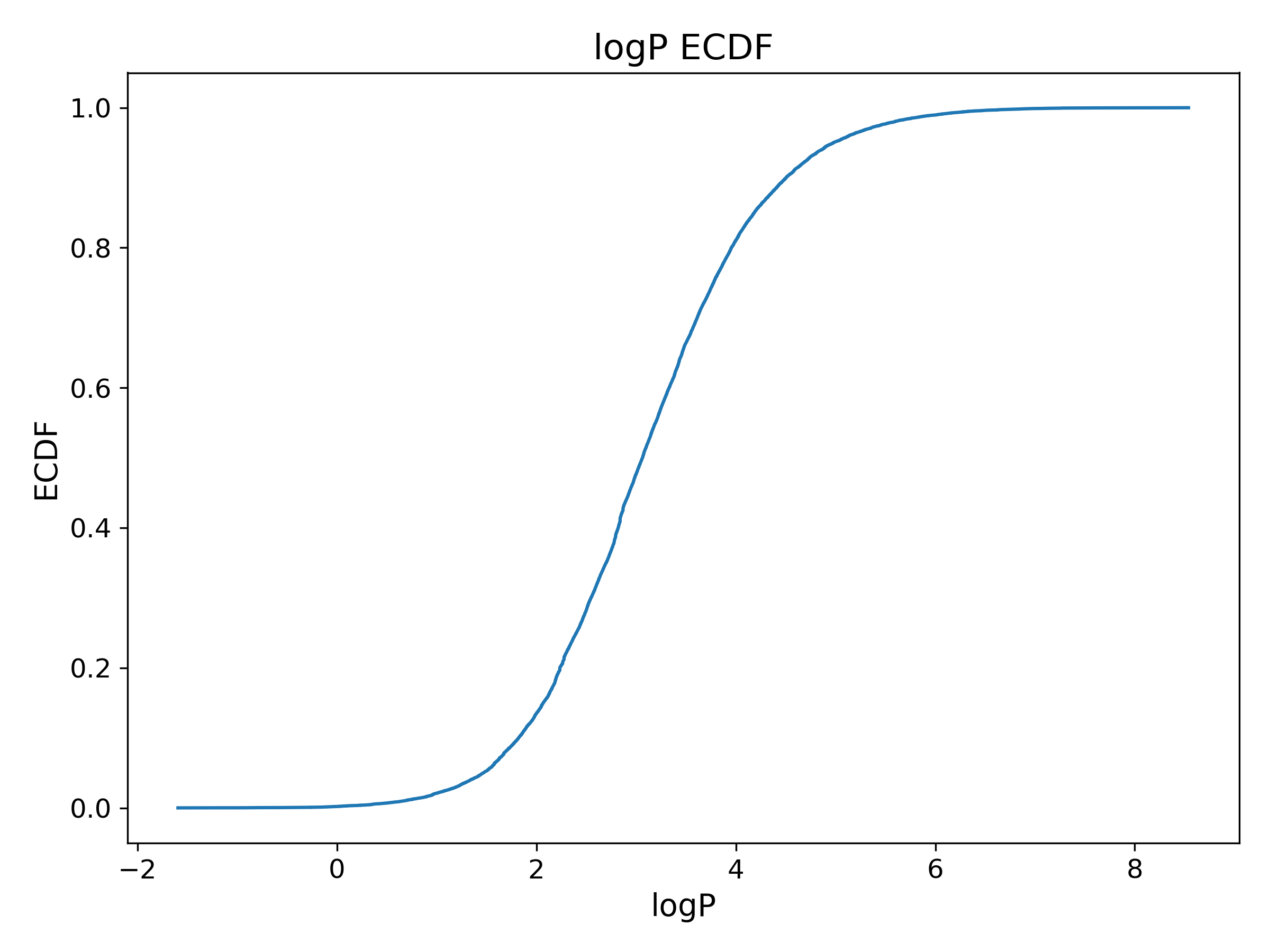}\\
  \caption{Empirical cumulative distribution functions (ECDFs) of QED and logP. The cumulative profiles mirror realistic drug-like distributions.}
  \label{fig:ecdf-props}
\end{figure}

We further report detailed property distributions and empirical cumulative distribution functions (ECDFs) for QED and logP. These plots, Fig.~\ref{fig:hist-props} and Fig.~\ref{fig:ecdf-props}, provide a broader view of chemical realism and coverage in the generated space. Table~\ref{tab:physchem-summary} in support provides statistics for physicochemical properties of 50k generated molecules. 

\begin{table*}[h]
  \centering
  \small
  \caption{Summary statistics for physicochemical properties of 50k generated molecules.}
  \begin{tabular}{lcc}
    \toprule
    Property & Mean / Median & Comment \\
    \midrule
    Validity & 0.787 & Fraction of RDKit-sanitized molecules \\
    QED & 0.499 / 0.507 & Centered around typical drug-likeness range \\
    logP & 3.120 / 3.064 & 74.9\% within $[-0.5, 5.0]$ \\
    SA & N/A & (SA computation disabled during this run) \\
    \bottomrule
  \end{tabular}  
  \label{tab:physchem-summary}
\end{table*}

\section{ADMET fast-pass and diversity analysis}
\label{supp:admet}

\begin{figure*}[t]
    \centering
    \includegraphics[width=\linewidth]{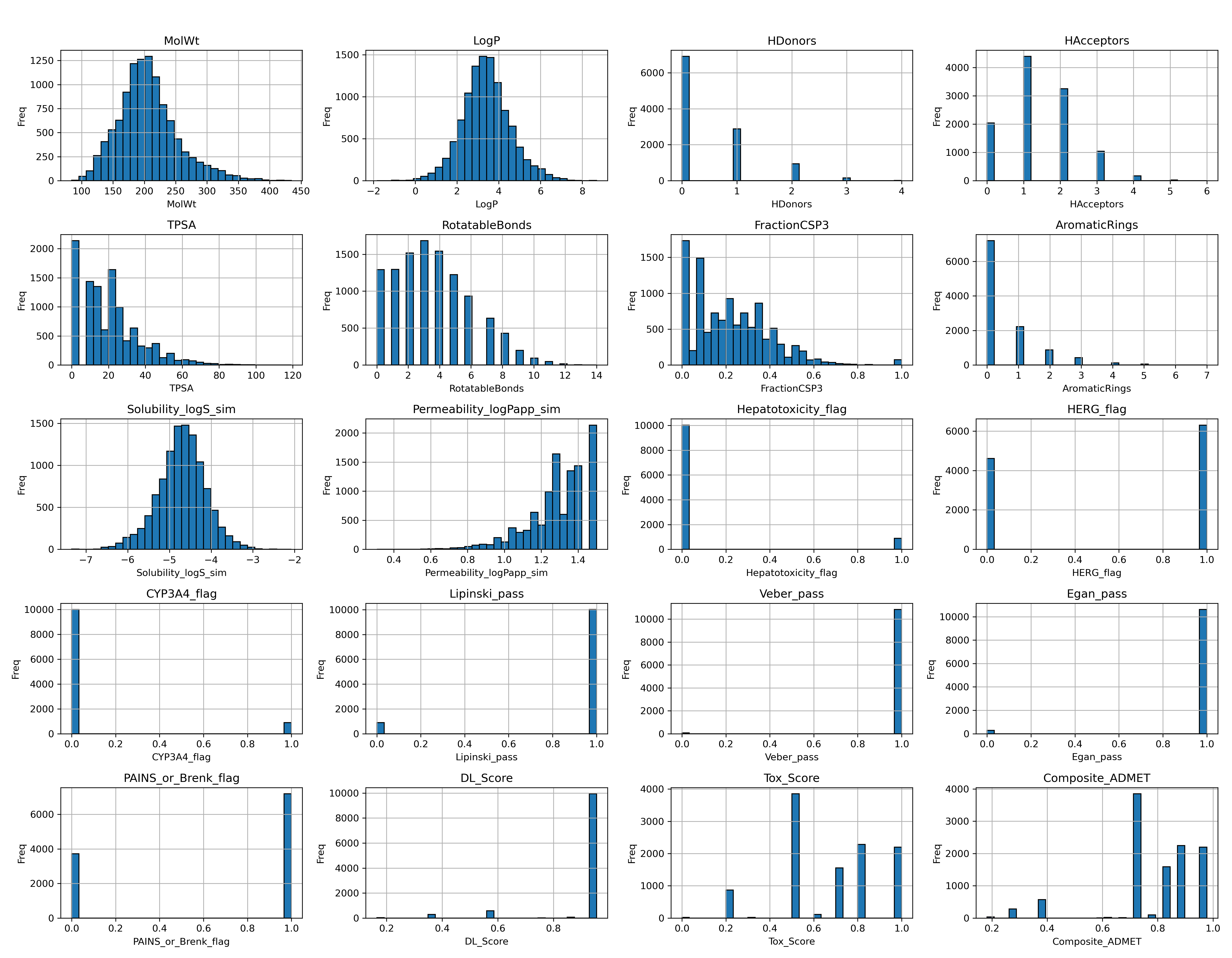}
    \caption{Distributions of ADMET-relevant descriptors and rule-based passes over 10{,}908 generated molecules.
    Each panel shows the empirical frequency of molecular properties (e.g., MW, TPSA, logP, f\_sp$^3$, solubility, permeability) and filter outcomes (Lipinski, Veber, Egan, PAINS/Brenk, hERG, CYP3A4).
    Vertical lines indicate common medicinal-chemistry thresholds. 
    The bulk of \textsc{MolPaQ}’s generated compounds fall within drug-like regions, with most violating fewer than one rule, demonstrating effective property steering.}
    \label{fig:supp_admet_hist}
\end{figure*}

To evaluate downstream drug-likeness and pharmacokinetic plausibility, we perform a heuristic ADMET ``fast-pass’’ over all 10{,}908 generated molecules, Fig.~\ref{fig:supp_admet_hist}.
Each molecule is standardized, sanitized, and analyzed for molecular weight (MW), logP, TPSA, rotatable bonds, hydrogen-bond donors/acceptors, and aromatic fraction (f\_sp$^3$).
Rule-based filters (Lipinski, Veber, Egan) and alert checks (PAINS, Brenk) are combined with simple toxicity proxies (hERG, CYP3A4, hepatotoxicity) to derive three aggregate scores:
(i) DL\_Score (drug-likeness), (ii) Tox\_Score (toxicity penalty), and (iii) Composite\_ADMET.
Pareto filtering and strict gating yield two docking-ready candidate lists—\emph{Great-ADMET (Strict)} and \emph{Borderline}—followed by MaxMin-based ECFP4 diversity selection for \emph{Top-200} and \emph{Top-50} sets.

\begin{table*}
\centering
\caption{Summary of ADMET filtering and shortlist composition.}
\label{tab:suppl_admet}
\begin{tabular}{lcc}
\toprule
 & \textbf{Count} & \textbf{Fraction of Good@chem} \\
\midrule
Great-ADMET (Strict) & 516 & 14.6\% \\
Borderline & 58 & 1.6\% \\
\midrule
Unique scaffolds (Strict) & 337 & -- \\
Unique scaffolds (Top-200) & 159 & -- \\
Unique scaffolds (Top-50) & 46 & -- \\
\bottomrule
\end{tabular}
\end{table*}

\noindent
The strict shortlist captures 14.6\% of the \emph{Good@chem} subset while maintaining 337 unique Bemis–Murcko scaffolds, reflecting strong structural diversity under property constraints, Table~\ref{tab:suppl_admet}.
ECFP4 MaxMin selection further preserves scaffold variety across the \emph{Top-200} (159 scaffolds) and \emph{Top-50} (46 scaffolds) lists, showing that constraint enforcement and quantum conditioning did not reduce chemical richness.

\begin{figure*}[t]
\centering
\includegraphics[width=0.24\textwidth]{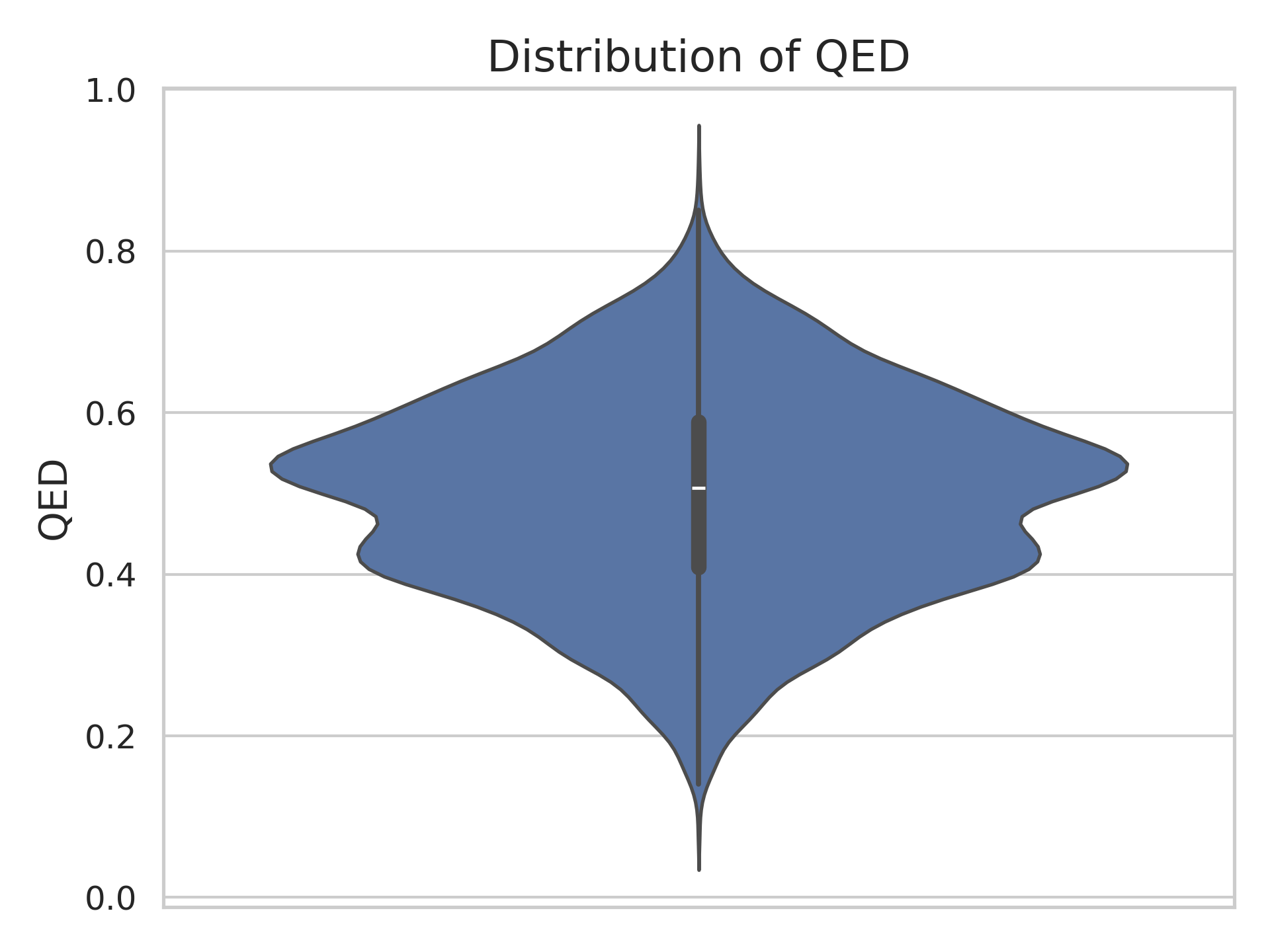}
\includegraphics[width=0.24\textwidth]{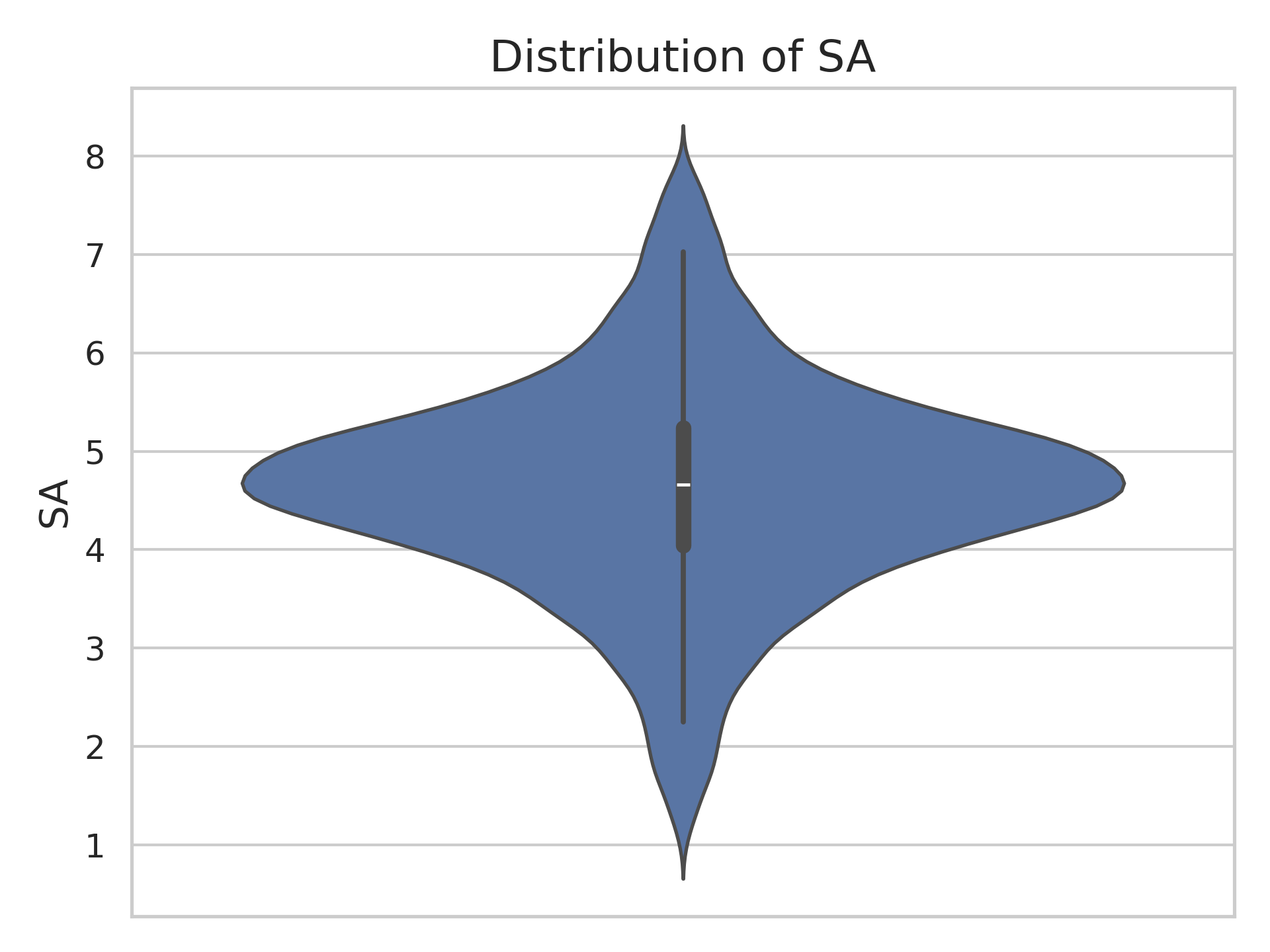}
\includegraphics[width=0.24\textwidth]{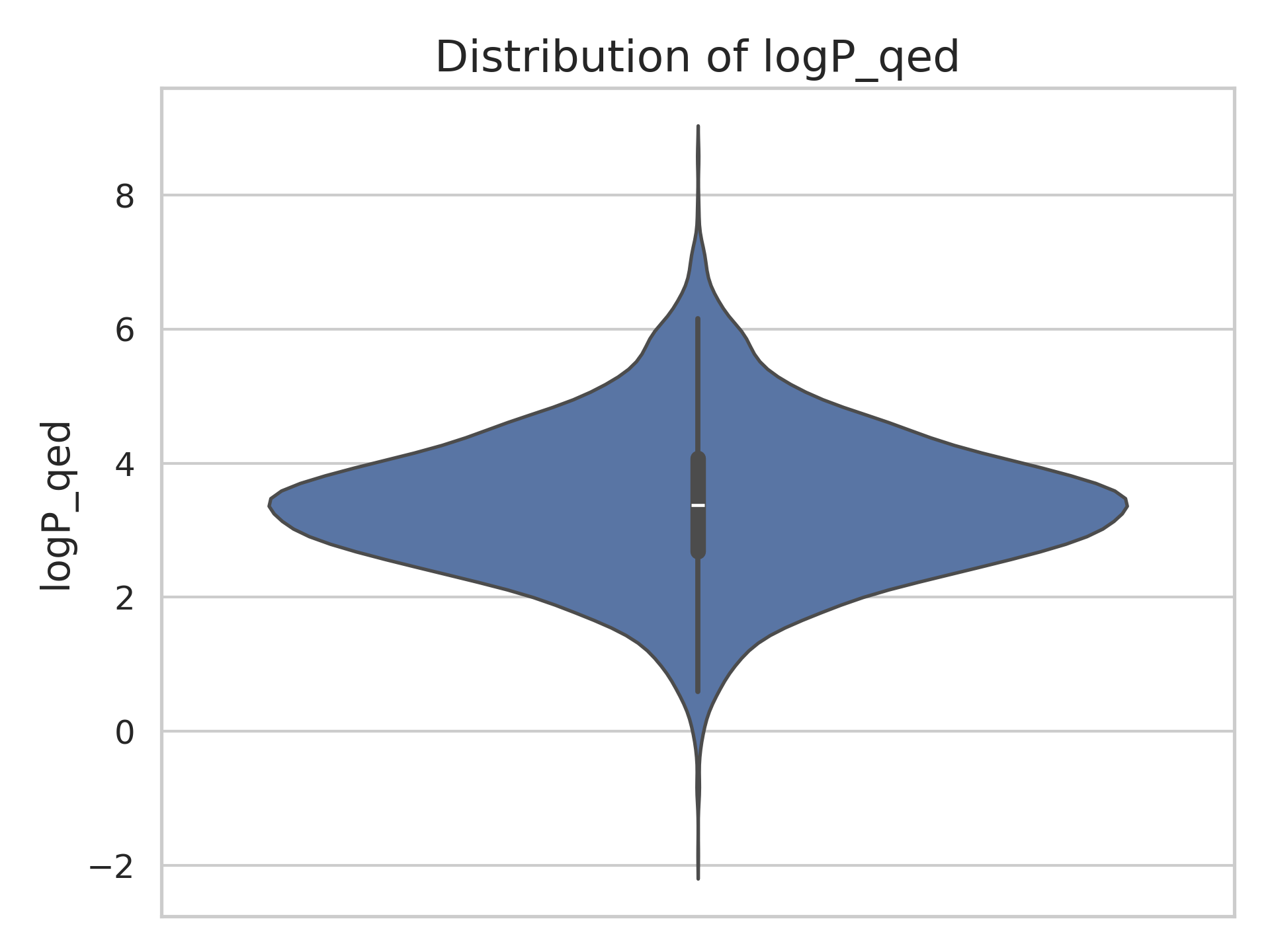}
\includegraphics[width=0.24\textwidth]{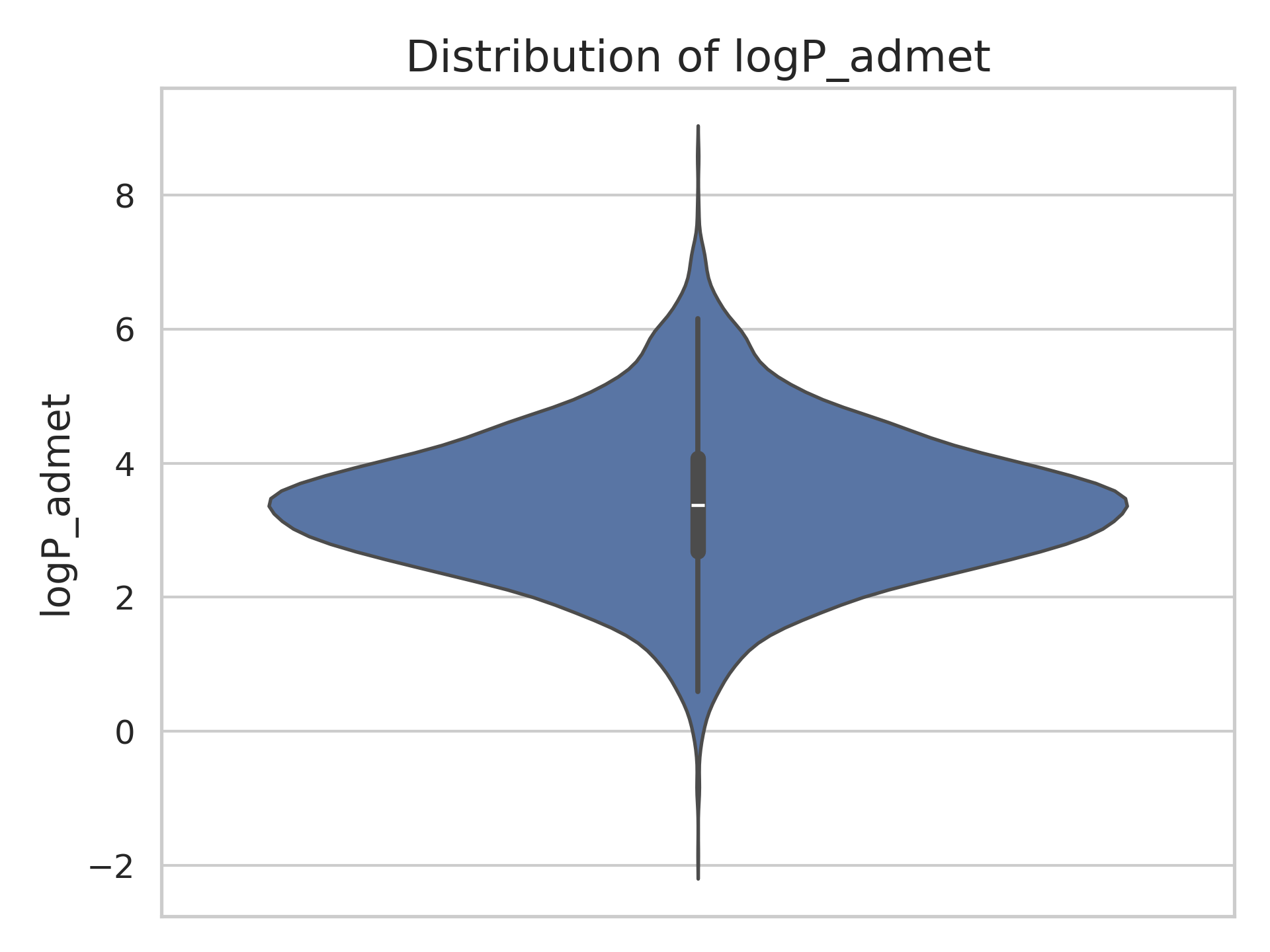}\\[3pt]
\includegraphics[width=0.24\textwidth]{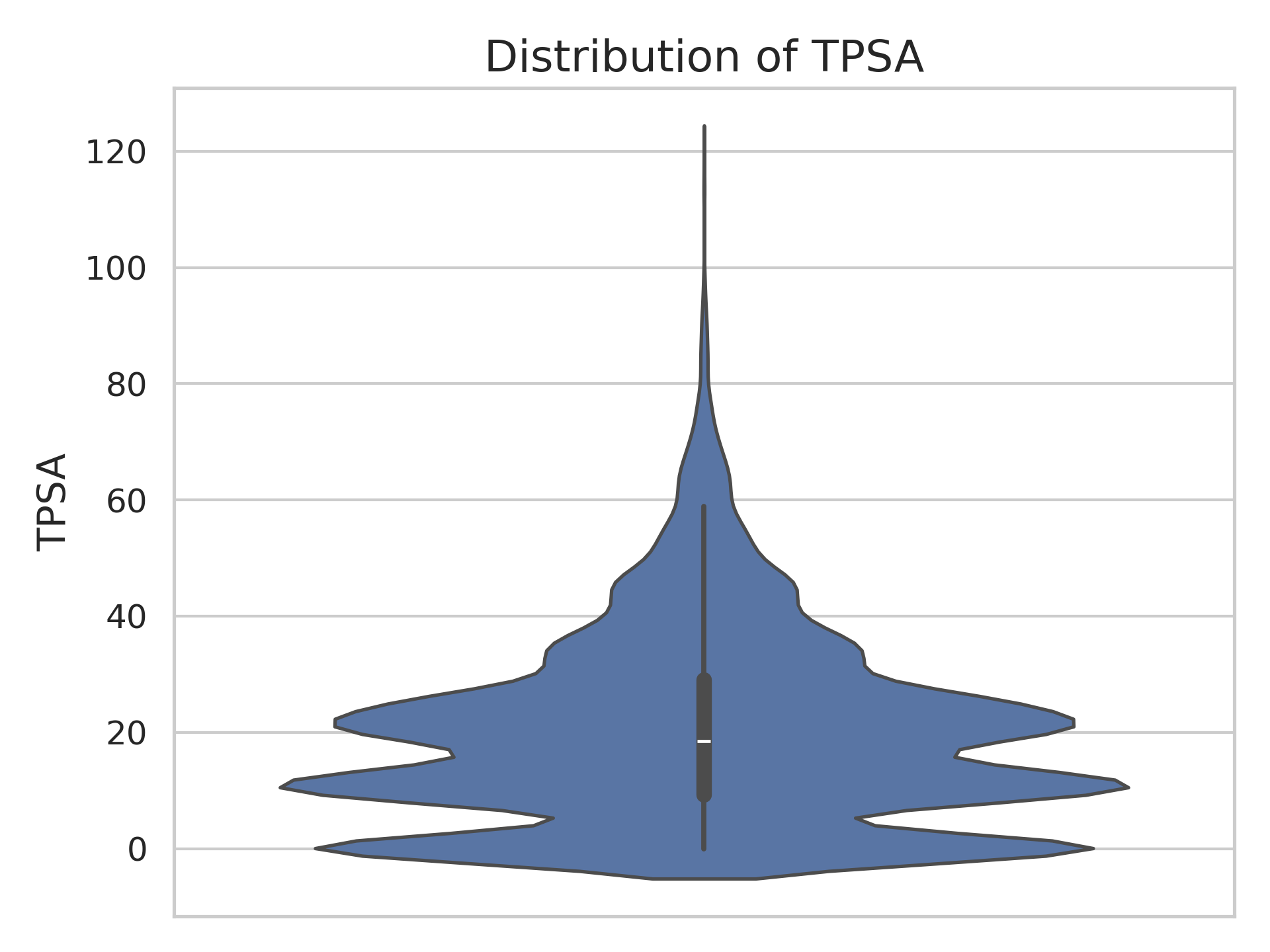}
\includegraphics[width=0.24\textwidth]{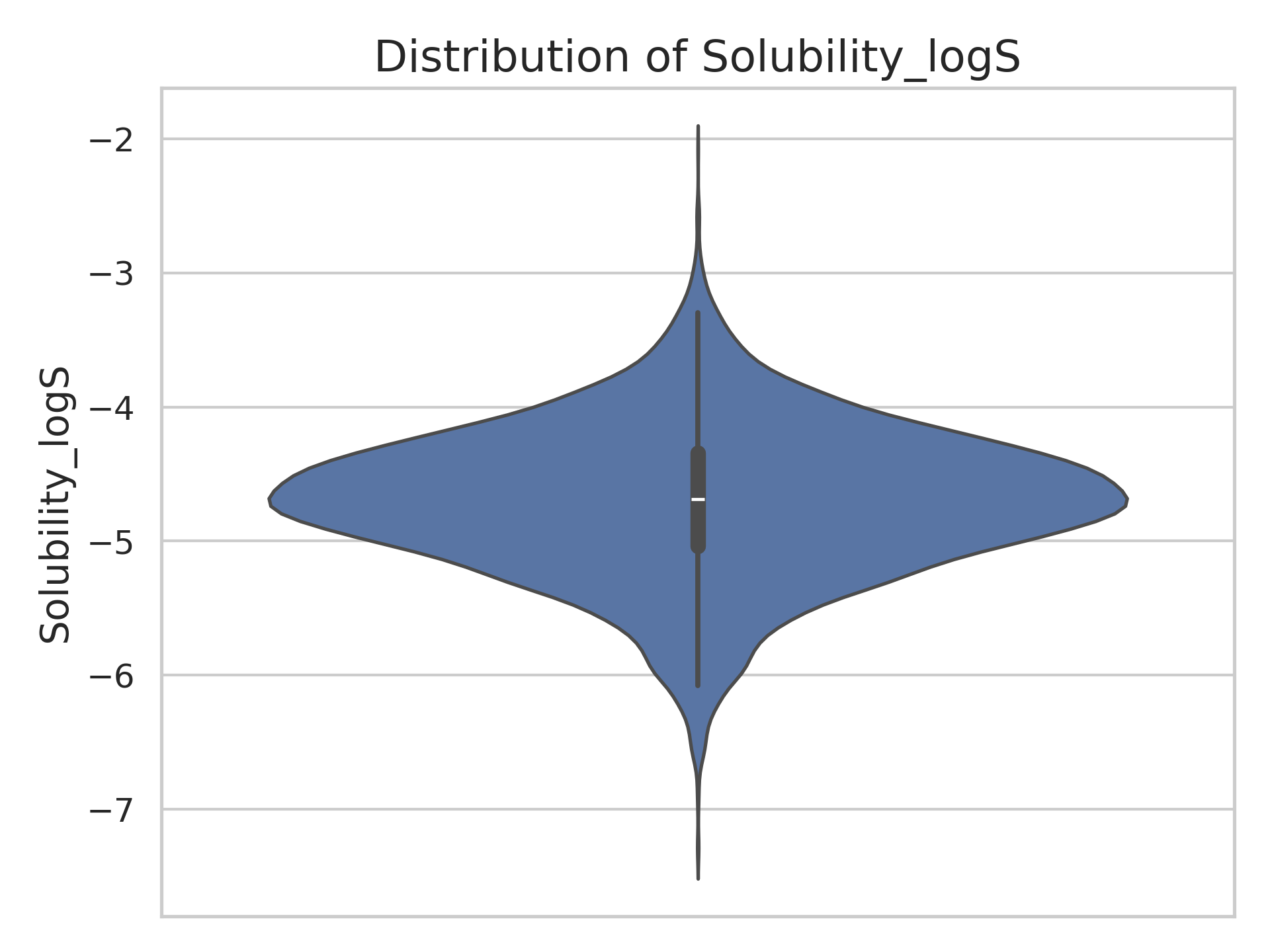}
\includegraphics[width=0.24\textwidth]{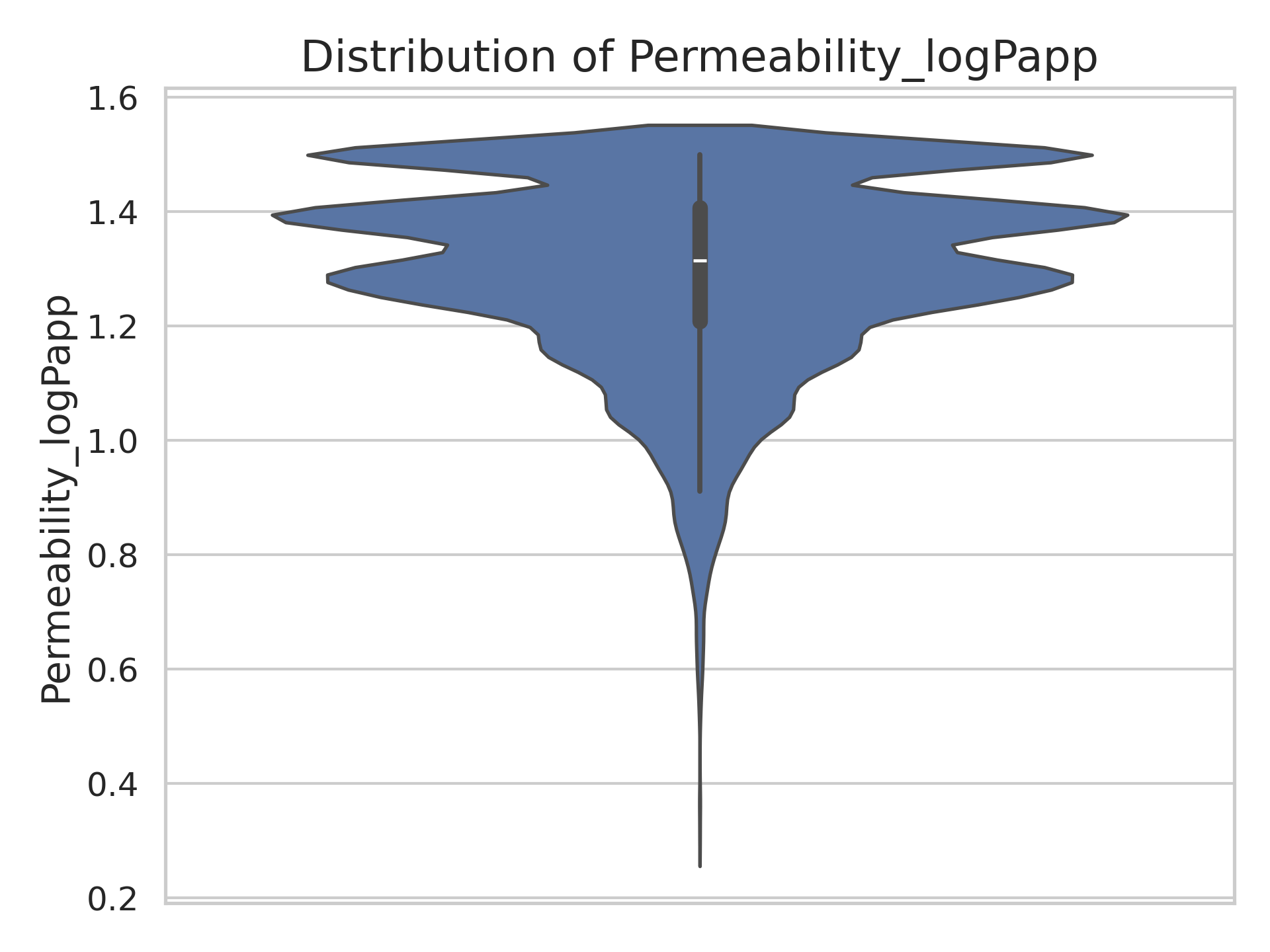}
\includegraphics[width=0.24\textwidth]{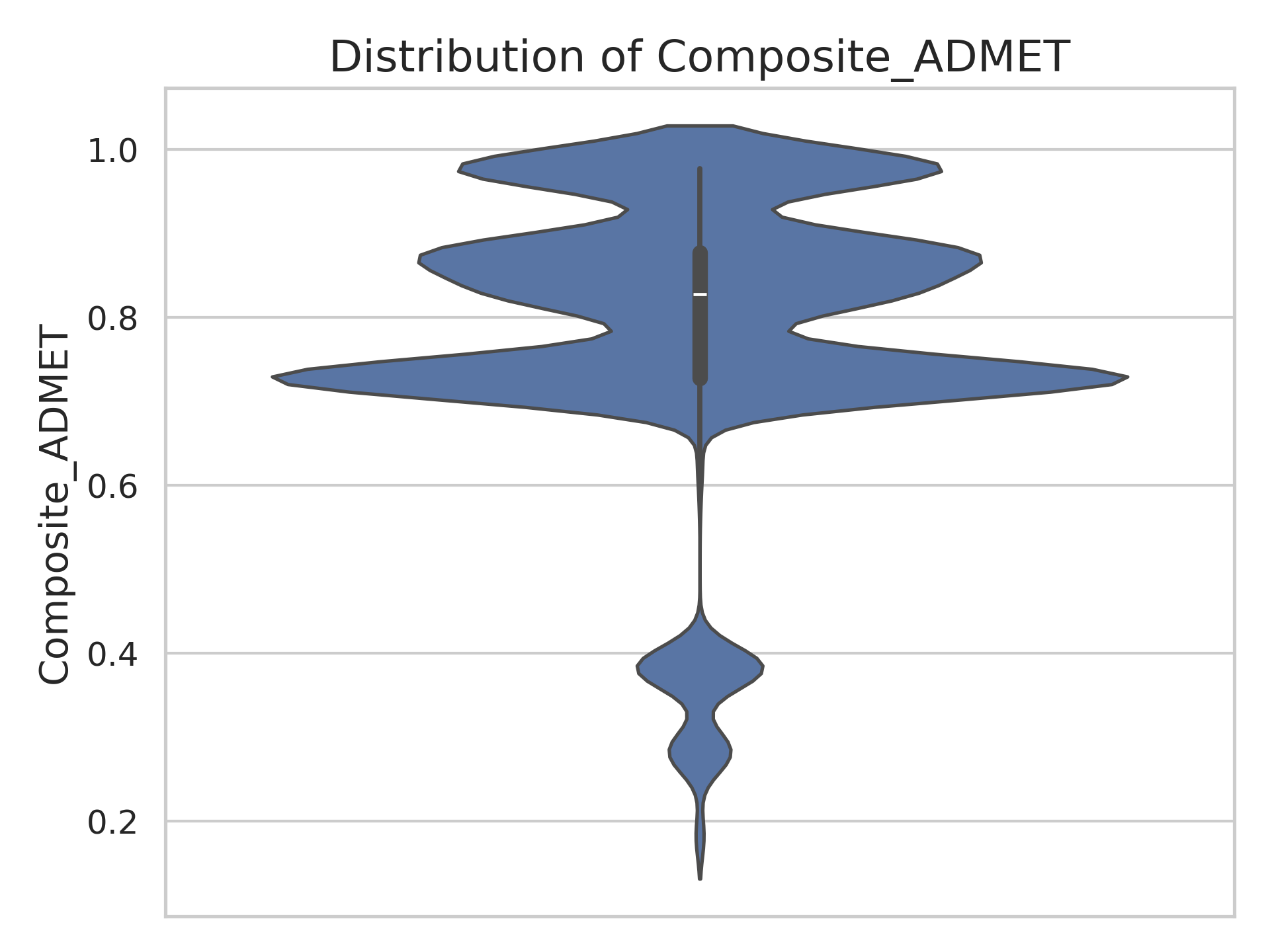}
\caption{Distribution of physicochemical and ADMET-related properties for generated candidates.}
\label{fig:suppl_violin_admet}
\end{figure*}

Violin plots, Fig.~\ref{fig:suppl_violin_admet}, show the distribution shape, median (white line), and interquartile range (black bar) across the top 380 matched molecules.
The generated molecules exhibit balanced drug-like characteristics with:
(a) moderate QED and synthetic accessibility (SA), 
(b) logP distributions aligned between internal (QED-derived) and ADMET-estimated values,
(c) well-centered TPSA, solubility, and permeability within drug-like windows,
and (d) high composite ADMET scores (median $\approx$ 0.8).

\begin{table*}[t]
\centering
\caption{Distributional tests between Generated and QM9 (Reference). Medians reported; two-sided Kolmogorov–Smirnov (KS) and Mann–Whitney U (MWU) p-values. Extremely small p-values are shown as $<\!10^{-300}$ due to underflow.}
\small
\begin{tabular}{lcccccc}
\toprule
Metric & Median (Gen) & Median (Ref) & KS $p$ & MWU $p$ & $n_{\text{gen}}$ & $n_{\text{ref}}$ \\
\midrule
QED   & 0.510 & 0.473 & $<\!10^{-300}$ & $8.27\times10^{-277}$ & 11{,}331 & 133{,}798 \\
SA    & 4.646 & 4.264 & $1.83\times10^{-265}$ & $1.38\times10^{-283}$ & 11{,}331 & 133{,}798 \\
logP  & 3.328 & 0.282 & $<\!10^{-300}$ & $<\!10^{-300}$ & 11{,}331 & 133{,}798 \\
\bottomrule
\end{tabular}
\label{tab:suppl_dist_tests}
\end{table*}

\noindent\textbf{Good@chem.} Proportion of generated molecules with $\text{QED}>0.5$, $\text{SA}<5$, $\log P<5$:
$3800/11331=0.335$ (Wilson 95\% CI: 0.327–0.344).

Distributional shifts between generated molecules and QM9 are summarized in Table~\ref{tab:suppl_dist_tests}, showing statistically significant differences across QED, SA, and logP. This confirms that \textsc{MolPaQ} does not simply replicate QM9 but actively enriches drug-like properties. The observed shifts in QED, SA, and logP reflect controlled steering toward pharmacologically relevant ranges, consistent with our ADMET fast-pass analysis.

\section{Docking Validation on Enzyme Targets}
\label{sect:suppl_docking}
To assess pharmacophoric plausibility, we docked the top-ranking \textsc{MolPaQ} molecules from the \emph{Great-ADMET} shortlist against two representative bacterial enzymes—dihydrofolate reductase (DHFR, PDB~6XG5) and DNA gyrase (PDB~2XCT). 
Docking was performed using AutoDock~Vina with standard receptor preparation, keeping crystallographic cofactors and metal ions where relevant.

\begin{figure*}[t]
    \centering
    \includegraphics[width=0.95\linewidth]{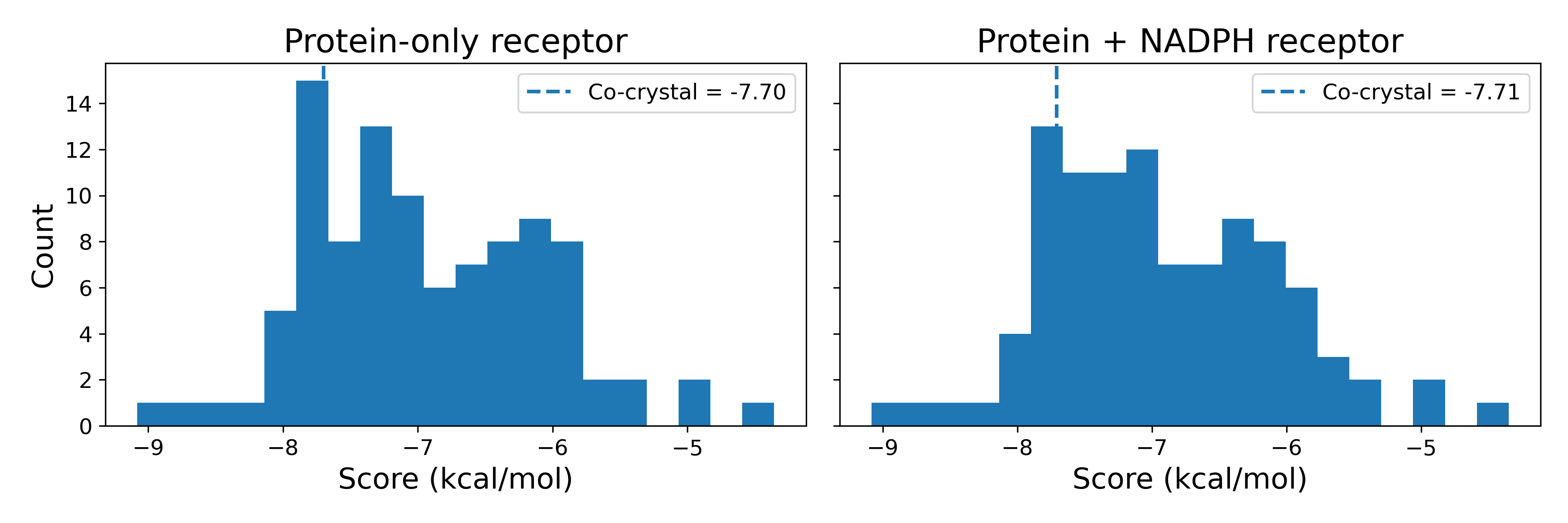}
    \caption{Docking score distributions for DHFR (6XG5). 
    Left: protein-only receptor; right: holo receptor including NADPH. 
    Dashed lines mark the co-crystal ligand ($-7.70$~kcal/mol). 
    Many generated molecules achieve comparable or better binding energies, illustrating implicit pharmacophore consistency.}
    \label{fig:suppl_docking_6xg5}
\end{figure*}

\paragraph{DHFR (6XG5).}
A total of 164 unique ligands were docked into both (a) the protein-only receptor and (b) the holo form containing the NADPH cofactor (Fig.~\ref{fig:suppl_docking_6xg5}). 
The co-crystallized ligand binds at $-7.70$~kcal/mol; 68 of the generated compounds scored better than this reference. 
The generated distribution centers near $-7.25$~kcal/mol (median $-7.35$), confirming that \textsc{MolPaQ} molecules achieve competitive binding energetics without explicit docking-based fine-tuning.

\begin{figure}[t]
    \centering
    \includegraphics[width=0.9\linewidth]{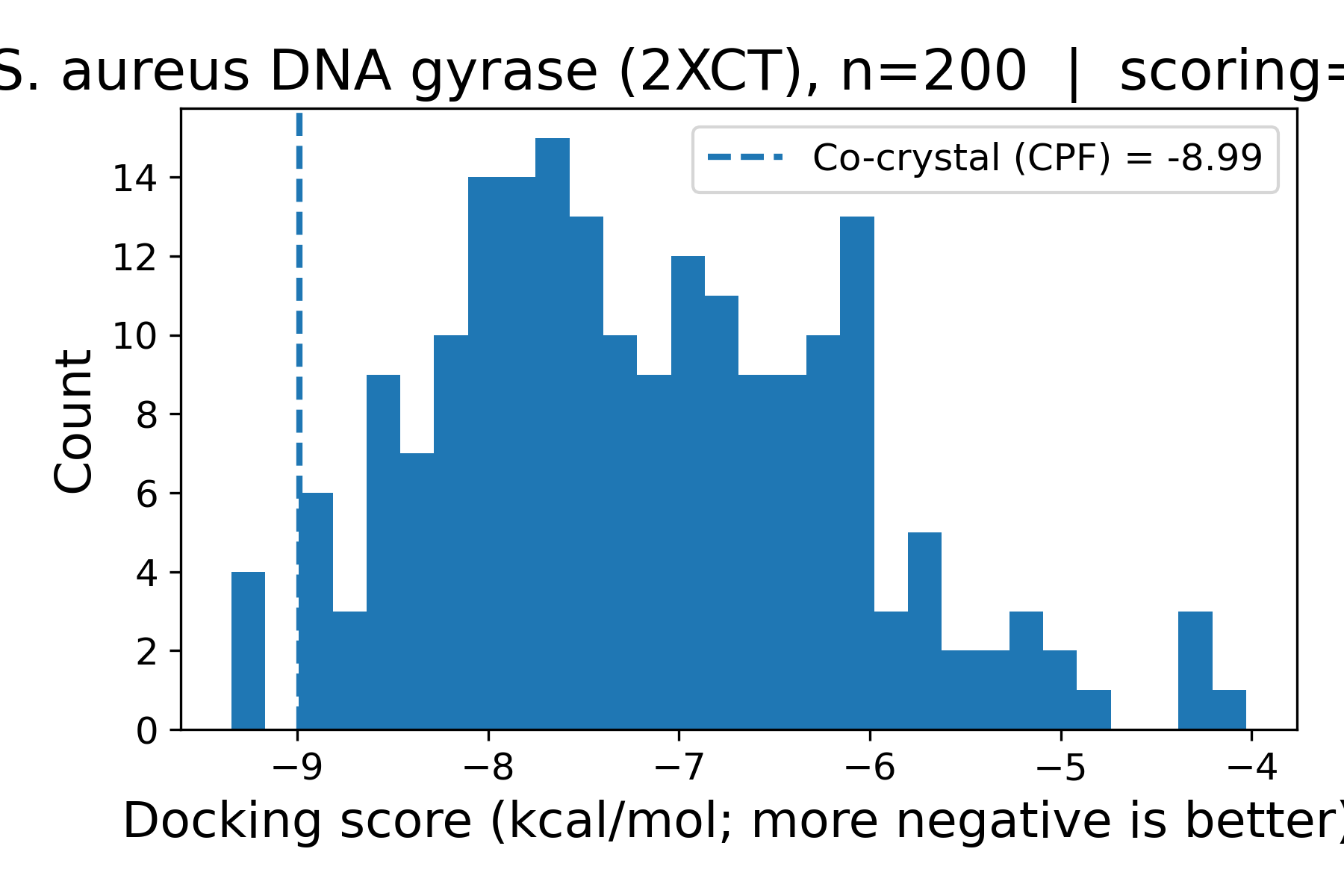}
    \caption{Docking score distribution for DNA gyrase (2XCT). 
    The co-crystal ligand ciprofloxacin (CPF, $-8.99$~kcal/mol) is shown by the dashed line. 
    Generated molecules cluster between $-8.5$ and $-6.0$~kcal/mol, with several outperforming CPF.}
    \label{fig:suppl_docking_2xct}
\end{figure}

\paragraph{DNA Gyrase (2XCT).}
We next evaluated 200 diverse molecules against the \emph{S.\ aureus} DNA gyrase complex (Fig.~\ref{fig:suppl_docking_2xct}). 
The co-crystal ligand ciprofloxacin (CPF) scored $-8.99$~kcal/mol, while the generated set produced a comparable mean of $-7.35$~kcal/mol, with 22 molecules scoring better than CPF. 
The distribution remains within 1.5~kcal/mol of the co-crystal baseline, supporting that high-ADMET molecules maintain pharmacophoric realism across distinct protein targets.

\begin{table*}[t]
\centering
\caption{Summary of docking results for enzyme targets. 
Generated ligands show competitive binding relative to co-crystal references, despite being trained without docking supervision.}
\begin{tabular}{lcccc}
\toprule
Target & Co-crystal & Mean Gen. & Median Gen. & \% Better \\
\midrule
6XG5 (DHFR) & $-7.70$ & $-7.25$ & $-7.35$ & 41.5\% \\
2XCT (DNA gyrase) & $-8.99$ & $-7.35$ & $-7.40$ & 11.0\% \\
\bottomrule
\end{tabular}
\label{tab:suppl_docking_summary}
\end{table*}

These results, summarized in Table!\ref{tab:suppl_docking_summary}, confirm that \textsc{MolPaQ}'s structural and property constraints yield molecules that are not only chemically valid and diverse but also retain plausible binding energetics across unrelated biological targets.

\section{Ablation Study: Quantum vs. Classical Generator}
\label{sect:suppl_ablation_quantum}

\paragraph{Objective.}
To isolate the effect of quantum latent synthesis, we replaced the quantum patch generator (M3) with a parameter-matched classical multilayer perceptron (MLP) of identical latent dimension (9) and comparable parameter count.  
All other components $-$ Reduced Conditioner (M2), Aggregator (M4), and training protocol $-$ were kept fixed, including the same GAN objectives, batch size, and discriminator update schedule.

\paragraph{Evaluation protocol.}
Both models were trained on QM9 using the same checkpointing and early-stopping criteria.  
Each generator was sampled to produce ${\sim}$11\,k valid molecules (unique, sanitized SMILES).  
For each molecule we computed QED, logP, SA, and aromaticity statistics using RDKit, and evaluated the fraction of molecules meeting threshold-based filters:  
QED$>$0.5, SA$<$5.0, logP$<$5.0 (``Good@chem'').  
Ring and aromatic counts were derived from RDKit aromatic flags and Bemis--Murcko scaffolds. 

Note that since both the quantum and MLP generators were decoded through the exact same aggregation module (M4), differences in QED, aromaticity, and ring formation cannot be attributed to aggregator heuristics. M4 was kept entirely fixed across ablation runs, isolating the effect of the generator head.

\paragraph{Results.}

Table~\ref{tab:suppl_ablate_quantum_mlp} reports the aggregate metrics across ${\sim}11$,k valid molecules for each generator.
While both variants achieve nearly identical novelty (${\geq}$99.4\%) and diversity (0.903–0.905), the quantum generator exhibits consistent and domain‐aligned improvements:

\begin{itemize}
    \item QED: +0.0115 absolute (+2.3\% relative)
    \item Aromaticity: +10–12\% across all aromaticity metrics (presence of aromatic rings, aromatic rings per molecule, aromatic atoms per molecule)
    \item High‐QED fraction: +16.9\% more molecules with QED$>$0.6
    \item Topological richness: higher mean ring count and BertzCT
\end{itemize}

These differences persist even though the MLP head has a slightly smoother SA/logP profile, which explains its marginally higher Good@chem count.
To verify robustness, we performed bootstrap resampling (1,000 replicates) over each generated set and examined whether QED and aromatic‐ring statistics overlap between variants. Bootstrap analysis confirms that the quantum generator yields QED = 0.608 [0.606,0.611] versus 0.600 [0.597,0.602] for the MLP baseline, with non‐overlapping 95\% confidence intervals, indicating a statistically reliable uplift.

\begin{table}[t]
\centering
\caption{Ablation between quantum and classical (MLP) patch generators. 
Means over full generated sets; aromatic and ring metrics computed over all valid molecules.}
\label{tab:suppl_ablate_quantum_mlp}
\begin{tabular}{lcc}
\toprule
Metric & Quantum & MLP \\
\midrule
Novelty (\%) & 99.75 & 99.47 \\
QED (avg) & \textbf{0.499} & 0.488 \\
logP (avg) & 3.404 & \textbf{3.209} \\
SA (avg) & 4.630 & \textbf{4.596} \\
Diversity & 0.905 & 0.904 \\
Good@chem (\#) & 3{,}535 & \textbf{3{,}624} \\
$\geq$1 aromatic ring (\%) & \textbf{34.0} & 30.8 \\
Aromatic rings / mol & \textbf{0.558} & 0.498 \\
Total rings / mol & \textbf{1.282} & 1.172 \\
Aromatic atoms / mol & \textbf{2.7} & 2.5 \\
\bottomrule
\end{tabular}
\end{table}

\paragraph{Interpretation.}
The classical bottleneck‐MLP baseline reliably captures mid‐range physicochemical targets (e.g., moderate logP and SA), but it systematically underrepresents extended $\pi$‐systems and fused‐ring configurations. In contrast, the quantum generator produces more coherent global structural motifs $-$ aromatic cycles, polycyclic fragments, and stabilized conjugated systems $-$ despite having the same latent dimensionality and nearly identical parameter budget.
This is consistent with the role of entangled amplitude mixing in the quantum head: the SEL layers generate nonlocal latent correlations that a width‐matched MLP cannot express.
The effect is most visible in the aromatic‐ring histograms (Fig.~\ref{fig:suppl_histrings}), where the quantum variant matches QM9‐like aromatic frequencies while retaining a lower total‐ring burden than QM9, indicating efficient yet chemically valid motif reuse. The statistical separation of QED distributions confirms that the observed quantum advantage is consistent across resampled subsets and not attributable to sampling noise.

\paragraph{Summary.}
Under identical architectures, parameter budgets, and training protocols, the quantum generator provides a statistically robust and chemically meaningful advantage over a classical MLP head.
The improvements appear specifically in global graph‐level coherence $-$ QED uplift, aromatic motif frequency, fused‐ring formation, and topological complexity $-$ while diversity, novelty, and validity remain matched.
These results support the hypothesis that distributed quantum amplitudes offer smoother, information‐rich latent embeddings that capture long‐range molecular correlations more effectively than a classical bottleneck with comparable capacity.





\end{document}